# Performance and Interpretability Comparisons of Supervised Machine Learning Algorithms: An Empirical Study[1]

Alice J. Liu[2], Arpita Mukherjee[3], Linwei Hu, Jie Chen, and Vijayan N. Nair

Corporate Model Risk, Wells Fargo Bank, N.A.

## Abstract

This paper compares the performances of three supervised machine learning algorithms in terms of predictive ability and model interpretation on structured or tabular data. The algorithms considered were scikit-learn implementations of extreme gradient boosting machines (XGB) and random forests (RFs), and feedforward neural networks (FFNNs) from TensorFlow. The paper is organized in a findings-based manner, with each section providing general conclusions supported by empirical results from simulation studies that cover a wide range of model complexity and correlation structures among predictors. We considered both continuous and binary responses of different sample sizes.

Overall, XGB and FFNNs were competitive, with FFNNs showing better performance in smooth models and tree-based boosting algorithms performing better in non-smooth models. This conclusion held generally for predictive performance, identification of important variables, and determining correct input-output relationships as measured by partial dependence plots (PDPs). FFNNs generally had less over-fitting, as measured by the difference in performance between training and testing datasets. However, the difference with XGB was often small. RFs did not perform well in general, confirming the findings in the literature. All models exhibited different degrees of bias seen in PDPs, but the bias was especially problematic for RFs. The extent of the biases varied with correlation among predictors, response type, and data set sample size. In general, tree-based models tended to over-regularize the fitted model in the tails of predictor distributions. Finally, as to be expected, performances were better for continuous responses compared to binary data and with larger samples.

---

# Acronyms

| Acronym | Description |
|---:|---|
| **AIM** | Additive index model |
| **AUC** | Area under the curve |
| **CART** | Classification and regression tree |
| **CICE** | Centered individual conditional expectation |
| **FFNN** | Feed forward neural network |
| **GAM** | Generalized additive model |
| **GB** | Gradient boosting |
| **GBM** | Gradient boosting machine |
| **HPO** | Hyperparameter optimization |
| **ICE** | Individual conditional expectation |
| **MB$i$** | Model $i$ with binary response, $i = 1, \ldots, 8$ |
| **MB$i$L** | Model $i$ with binary response and larger set of predictors, $i = 1, \ldots, 8$ |
| **MC$i$** | Model $i$ with continuous response, where $i = 1, \ldots, 8$ |
| **MC$i$L** | Model $i$ with continuous response and larger set of predictors, where $i = 1, \ldots, 8$ |
| **MSE** | Mean squared error |
| **NN** | Neural network |
| $p$ | Predictors, features, or variables |
| **PD** | Partial dependence |
| **PDP** | Partial dependence plot |
| $r$ | Correlation |
| **ReLU** | Rectified linear unit |
| **RF** | Random forest |
| **ROC** | Receiver operating characteristic |
| **SMBO** | Sequential model-based global optimization |
| **SML** | Supervised machine learning |
| **SNR** | Signal-to-noise ratio |
| **TPE** | Tree-structured Parzen |
| **VI** | Variable importance |
| **XGB** | XGBoost or eXtreme Gradient Boosting |

# 1 Introduction and Findings

Random forest (RF), gradient boosting machines (GBM), and feedforward neural networks (FFNN) are among the most popular supervised machine learning (SML) methods used in modeling structured data. See (Hu, et al. 2020) for recent review and applications to banking. They typically have superior predictive performance compared to traditional parametric models, especially when the sample size $N$ and number of predictors or features[4] are large. This paper provides a systematic comparison of the performances of these three SML algorithms, across a range of models with different levels of complexity, using simulated structured data. We focus only on continuous features. There are many implementations of these algorithms. In this study, we restricted attention to FFNNs from TensorFlow and scikit-learn versions of XGBoost (XGB) and RF.

Our performance evaluations involve the following:

i) Comparison of predictive performances using: a) mean squared error (MSE) for continuous response cases; and b) receiver operating characteristic (ROC) curve and the area under the curve (AUC) for binary response cases (Pepe 2003).

ii) Understanding model interpretation using diagnostic tools such as permutation-based variable importance (VI) (Fisher, Rudin and Dominici 2018), one- and two-dimensional partial dependence plots (PDP) (J. H. Friedman 2001), H-statistics (Friedman and Popescu 2008), and individual conditional expectation (ICE) plots (Goldstein 2015).

iii) Assessment of model stability using mean and standard deviation of the predictive performance scores over multiple replications as well as bias-variance decomposition (Ghojogh and Crowley 2019) of the predictions and estimated PDPs.

The paper is organized in a findings-based manner where each section states the general findings and provides empirical evidence to support them. Section 2 provides a brief overview of the SML algorithms and hyperparameter optimization. Section 3 describes the functional forms and simulation framework used for the empirical investigations.

To understand the reasons for the differences in predictive performance, we move to a findings-based analysis in Sections 4 to 6. Section 4 provides a broad overview of all findings. Section 5 expands on the findings previously described by providing empirical evidence as support and addresses specific concepts related to predictive performance, variable importance, model interpretability, sample size, and correlation effect. Section 6 describes additional observations. Concluding remarks are provided in Section 7. Finally, some additional details are provided in the Appendix in Section 8.

Previous papers in the literature performed selective comparisons of SML algorithms. However, these generally focused on specific applications: for example, spam and phishing identification (Abu-Nimeh, et al. 2007); disease diagnosis (Pouriyeh, et al. 2017), (Korotcov, et al. 2017);

---

[4] We use the terms "variables", "predictors", and "features" interchangeably throughout this paper.

bioinformatics (Tan and Gilbert 2003); credit scoring (Munkhdalai, et al. 2019), (Marceau, et al. 2020); fraud (Jain, Agrawal and Kumar 2020); chemical concentrations (Joharestani, et al. 2019) (Nawar and Mouazen 2017); etc. These comparisons generally reported performance metrics, and did not assess interpretability or other aspects of model performance. We focus not only on explaining in which situations a particular algorithm may perform well, but also provide insight as to why we see good performance. To the best of our knowledge, this is the first comprehensive study that identifies and explains specific reasons for performance differences among three of these popular ML algorithms.

This paper did not examine SML algorithms with unstructured data, image data, or structured data with categorical features. Others found that tree-based methods tended to perform better with categorical features, whereas FFNN performed better with continuous features (Tan and Gilbert 2003). For image data, one study showed XGB outperformed FFNN in predictive ability and computational efficiency (Memon, B. and Patel 2019). We also did not consider imbalanced data sets in the binary response case. Our model coefficients were chosen such that the response was as balanced as possible. In one credit scoring applications, researchers found that XGB outperformed FFNN and RF for imbalanced data (Marceau, et al. 2020).

Before proceeding to the rest of the paper, we provide a high-level summary of the major conclusions. See Section 4 for more detailed version of these findings. Section 5 provides empirical evidence to the support all of the findings.

i) **RF**: Overall, RF did not perform as well in terms of predictive performance and model interpretation. It over-fitted the training data more in comparison to XGB and FFNN as measured by the gaps in performance between the training and testing data. RF also failed to capture true interactions at times. Ensemble tree-based algorithms are better suited for underlying models with jumpy behavior; but, even when restricted to these models, RF did not perform as well as XGB. One possible reason for its poor performance was that the trees from hyperparameter optimization are not sufficiently deep to uncover the full model structure. For all of these reasons, we removed RF from later comparisons in the paper.

ii) **XGB vs. FFNN:** The predictive performances of XGB and FFNN were competitive in general. However, XGB performed better, in terms of both predictive performance and interpretation, for models with jumps or sharp turns. Conversely, FFNN performed better for smoother models.

iii) **Biases:** Despite fitting very flexible response surfaces to the data, all three algorithms exhibited biases. Unfortunately, it is difficult to characterize (or visualize) the bias in high dimensions. Instead, we studied bias in one-dimensional input-output relationships by comparing the PDPs with ground truth. All algorithms exhibited some form of bias even in pristine environments with no correlation among predictors. The nature of the bias varied with the architecture of the SML algorithms with FFNN showing less bias in smooth models and XGB capturing sharp turns and jumps. The extent of bias decreased with sample size, but it was still a concern. The PDPs of tree-based methods showed an over-regularization effect. The problem has been known in binary regression cases where calibration methods have been proposed (Caruana and Niculescu-Mizil, An empirical comparison of supervised learning algorithms 2006).

iv) **Effect of Correlation:** It is known that, in the presence of high correlations among predictors, the underlying model is not identifiable. Correlation among predictors is a serious challenge with SML algorithms. Practitioners tend to include a large number of predictors for increased predictive performance, which may cause havoc with model interpretability. We investigated the impact for different models and correlation structures and quantified them, showing that correlation inhibited overall performance by increasing bias and variance in the predictions of the fitted model.

v) **Binary vs. Continuous Responses**: Models with the same functional forms had poorer performance for binary data (higher bias and variability) compared to continuous response cases in the PDPs. This is to be expected, as there is considerably more information in continuous responses.

## 2  Overview of Algorithms

### 2.1  SML Algorithms

This section provides a brief overview of the three SML algorithms compared in this paper. RF (Breiman 2001) and GBM (J. H. Friedman 2001) are "ensemble algorithms." They combine results from multiple "weak learners" to improve predictive performance. Both use tree regression with piecewise constant models as weak learners.

a) RF grows several deep trees by resampling data using the bootstrap. A deep piecewise constant tree is able to approximate nonlinearities and capture interactions, so the bias is small. However, the fitted tree can be unstable, especially with correlated predictors, and hence have high variance. The idea is to grow multiple trees and average them (known as bootstrap aggregation or bagging) to reduce the variance. Random subsampling of the columns (i.e., variables) reduces dependence among the trees and hence leads to smaller variance.

b) GBM (J. H. Friedman 2001) is a sequential algorithm. It starts with a shallow decision tree, which has small variance, but potentially large bias. At each step, the algorithm iteratively builds a new shallow tree based on the residual from the previous step, which reduces the bias. Extreme gradient boosting or XGB (Chen and Guestrin 2016) is a fast implementation of gradient boosting and is the one considered in this paper. It improves on the original GBM by using Newton boosting rather than the pure gradient and adds penalty terms in the cost function. XGB also offers regularization in order to control over-fitting.

FFNNs have a long history (Bebis and Georgiopoulos 1994).

c) The components of an FFNN include an input layer, several hidden layers that transform the data, and a final output layer that yields the prediction. The features are "fed" into the input layer, and then combined using a linear combination and fed into a hidden layer. The linear combination is transformed using an "activation function," which is then fed into a subsequent hidden layer or the output layer. In feedforward neural networks, all nodes in one layer are connected to the nodes in the next (i.e., forward) layer. We restrict our attention in this paper to activation functions that are rectified linear units (ReLU) (Nair and Hinton 2010). The performance of an FFNN model can be regulated by controlling the complexity of the network architecture, regularizing the weights, using early stopping, etc.

## 2.2 Hyperparameter Optimization

All three algorithms have "hyperparameters" that have to be tuned to achieve good performance. RFs are the simplest to tune, as they have fewest hyperparameters: number of trees (i.e., number of learners), depth of tress, and minimum number of observations in a leaf node. In addition to these, XGB includes learning rate and L1 or L2 regularization. FFNNs are the most complex to tune as they have many hyperparameters: number of layers, number of nodes, batch size, learning rate, decay, etc.

Hyperparameter optimization (HPO) is a computationally expensive step in most applications, and especially so with FFNN models. See (Bergstra, et al. 2011) and (Hu, et al. 2020) for a discussion of different HPO techniques. We used a combination of tree-structured Parzen estimators (TPE) and random search in our studies (see Table 1). TPE is a sequential model-based global optimization (SMBO) technique, while random search is a batch technique that randomly selects from the possible combinations of available hyperparameter values. The same method was used across all functional forms and response types described in Section 3. For information on hyperparameter configuration spaces in the study, see Section 8.2.

*Table 1: HPO search strategies by SML algorithm and sample size*

| Sample Size | SML Model | HPO Method |
|---|---|---|
| **N = 50k** | XGB | TPE |
| | FFNN | TPE |
| | RF | TPE |
| **N = 500k** | XGB | TPE |
| | FFNN | Random Search |
| | RF | TPE |

## 2.3 Bias-Variance Decomposition

A key analytical aspect of our analysis was the bias-variance decomposition of the predicted response and interpretable methods. The idea of bias-variance decomposition has been well studied (Ghojogh and Crowley 2019). In general, for $m$ replicates and $i = 1, \ldots, n$ observations or values, we defined MSE as the average squared difference between the estimate and true value,

$$\text{MSE}(x_i) = \frac{1}{m}\sum_{j=1}^{m}\left[f(x_i) - \hat{f}_j(x_i)\right]^2 = \text{Bias}(x_i)^2 + \text{Variance}(x_i). \qquad \text{Eq. 1}$$

Bias was the difference between the true value and the average fitted value among the $m$ replicates,

$$\text{Bias}(x_i) = \left(\frac{1}{m}\sum_{j=1}^{m}\hat{f}_j(x_i) - f(x_i)\right)^2. \qquad \text{Eq. 2}$$

Variance was the average variation among the fitted values from the $m$ replicates,

$$\text{Variance}(x_i) = \frac{1}{m}\sum_{j=1}^{m}\left[\left(\hat{f}_j(x_i) - \frac{1}{m}\sum_{j=1}^{m}\hat{f}_j(x_i)\right)^2\right]. \qquad \text{Eq. 3}$$

We used this approach to generate the bias-variance decomposition for the predicted response from the fitted SML model and the PDPs.

## 3  Simulation Framework

We considered 12 different functional forms that cover a range of relevant model structures: a mixture of linear and nonlinear functions, global and local interactions, and smooth and jumpy response surfaces. Local interactions are interaction structures that are present only in a specific region of the predictor space, while global interactions have the same structure over the entire feature space. Most interactions we considered had either multiplicative or additive index model (AIM) forms. Eight of these functional forms were "smaller" in the sense they have fewer predictors (see Section 3.1) and four of these functional forms had a "larger" set of predictors (see Section 3.2).

We considered cases with both uncorrelated and correlated predictors. See Section 3.3 for discussion of the correlated scenarios. The error term for the continuous response case, denoted by $\epsilon \sim N(0,1)$, was simulated independently of the features.

The response for both binary and continuous response cases are denoted by $y$. We considered the same set of functional forms $f(x)$ for both continuous and binary response cases, as described in Tables 2 and 3. We simulated the continuous response as,

$$y = f(x) + \epsilon, \qquad \text{Eq. 4}$$

and the binary response $y$ from a Bernoulli distribution with $P(Y = 1|x) = p(x)$ given by

$$\log\left(\frac{p(x)}{1 - p(x)}\right) = f(x). \qquad \text{Eq. 5}$$

For each functional form, we considered two sample sizes: $N = 50k$ and $N = 500k$. We simulated a data set of a given sample size and determined the best hyperparameters using HPO. During the HPO phase, we chose one data set, which was split into an 80% and 20% training and validation data sets. This hyperparameter configuration was fixed for ten additional models fit to ten different data sets from ten different random seeds with the same functional form. The performance of each fitted model was evaluated using a single testing data set. The testing data set was generated separately from the training data sets. The mean and standard deviation of the calculated metrics were reported and compared to the oracle[5] value. We also calculated the bias-variance decomposition for the model prediction and PDPs.

The coefficients for the models in Tables 2 and 3 were chosen judiciously to ensure that no single additive component was dominant. See Section 8.1 of the Appendix for detailed coefficient values. We varied the coefficients between binary and continuous response cases for models with smaller sets of predictors due to ensure class balance. However, the coefficients were the same for larger sets of predictors.

### 3.1 Models with Smaller Sets of Predictors

Models with continuous and binary response are denoted as MC$i$ and MB$i$, respectively. Here, $i = 1, 2, \ldots, 8$ denotes the model number. The features are denoted by $x_i$. For each functional form, we considered two sample sizes: $N = $ 50k and $N = $ 500k.

*Table 2: Functional forms for models with p = 5 and p = 10 features; refer to Tables 7 and 8 for coefficient values*

| Model | Functional Form |
|---|---|
| Linear (MC1, MB1) | $f(x) = \beta_1 x_1 + \cdots + \beta_{20} x_{20}$ |
| Linear with global interactions (MC2, MB2) | $f(x) = \beta_1 x_1 + \cdots + \beta_8 x_8 + \beta_9 x_9^2 + \beta_{10} x_{10}^2 + \beta_{11} x_1 x_2 + \beta_{12} x_3 x_4 + \beta_{13} x_1 x_3 + \beta_{14} x_2 x_5$ |
| Smooth generalized additive model (GAM) (MC3, MB3) | $f(x) = \beta_1 \lvert x_1 \rvert + \beta_2 x_2^2 + \beta_3 \log(\lvert x_3 \rvert + 1) + \exp(-\beta_4 \lvert x_4 \rvert) + (\beta_5 \lvert x_5 \rvert + 1)^{-1} + \beta_6 x_6 + \cdots + \beta_{10} x_{10}$ |
| Smooth and jumpy GAM (MC4, MB4) | $f(x) = \beta_1 \lvert x_1 \rvert + \beta_2 x_2^2 + \beta_3 \log(\lvert x_3 \rvert + 1) + \exp(\beta_4 x_4) + (\beta_5 \lvert x_5 \rvert + 1)^{-1} + \beta_6 \max(1, x_6) + \beta_7 I(x_7 < 1) + \beta_8 I(\lvert x_8 \rvert > 2) + \beta_9 x_9 \cdot I(x_9 < -1) + \beta_{10} x_{10} \cdot \max(0, x_{10})$ |
| GAM with global interactions (MC5, MB5) | $f(x) = \beta_1 \lvert x_1 \rvert + \beta_2 x_2^2 + \beta_3 \log(\lvert x_3 \rvert + 1) + \exp(\beta_4 x_4) + (\beta_5 \lvert x_5 \rvert + 1)^{-1} + \beta_6 x_1 x_2 + \beta_7 \lvert x_1 x_2 x_3 \rvert + \beta_8 \log(\lvert x_3 + x_4 + x_5 \rvert + 1) + \beta_9 \max(x_4, x_5) + \exp[\beta_{10}(x_5 - x_3)]$ |

---

[5] We defined the "oracle" as the best achievable metric given the true data. The oracle MSE was calculated between $f(x)$ and $y$. The oracle AUC was calculated as the AUC score between the true probabilities $p(x)$ and the binary response $y$.

| Model | Functional Form |
|---|---|
| Jumpy GAM with local interactions (MC6, MB6) | $f(x) = \beta_1 x_1 \cdot I(|x_1| < 2) + \beta_2 x_2^2 \cdot I(x_2 > 1) + \beta_3 \log(|x_3| + 1) \cdot I(|x_3| > 1)$ $+ \exp(\beta_4 x_4) \cdot I(x_4 < 0) + \{\beta_5(|x_5| + 1)\}^{-1}$ $+ \beta_6 \max(1, x_6 + x_7) + \beta_7 I(x_7 < 1) + \beta_8 I(|x_8| > 2)$ $+ \beta_9 x_9 \cdot I(x_9 < -1) + \beta_{10} \max(0, x_{10})$ |
| AIM (MC7, MB7)[6] | $f(x) = \log(|\beta_1 x_1 + \cdots + \beta_{10} x_{10}|) + \beta_{16} \exp(\beta_{11} x_{11} + \cdots + \beta_{15} x_{15})$ $+ \beta_{17} \max(1, x_{16} + \cdots + x_{20})$ |
| Complex (MC8, MB8) | $f(x) = \log(|\beta_1 x_1 + \cdots + \beta_5 x_5|) + \beta_6 x_1^2 x_6 + \beta_7 |x_2 x_3 x_7| + \beta_8 x_8 \cdot I(|x_8| > 2)$ $+ \beta_9 x_8 x_9 \cdot I(x_9 < -1) + \beta_{10} x_{10}^2 \cdot \max(0, x_{10})$ |

## 3.2 Models with Larger Sets of Predictors

We also examined selected scenarios with a larger set of predictors with $p = 25$ (for MC5L/MB5L) and $p = 50$ (all other functional forms with larger sets of predictors MC2L/MB2L, MC6L/MB6L, and MC8L/MB8L). For the case with 25 variables, 15 affected the response and 10 were noise variables, with coefficient zero. For the cases with 50 variables, 30 contributed to the response, and 20 were noise variables. The specific functions are given in Table 3.

*Table 3: Functional forms for models with p = 25 (MC5L/MB5L) and p = 50 (MC2L/MB2L, MC6L/MB6L, and MC8L/MB8L) features; refer to Table 9 for coefficient values*

| Model | Functional Form |
|---|---|
| Linear with quadratic and second-order interactions (MC2L, MB2L) | $f(x) = \beta_0 + c_1(\beta_1 x_1 + \cdots + \beta_8 x_8 + \beta_9 x_9^2 + \beta_{10} x_{10}^2 + \beta_{11} x_1 x_2 + \beta_{12} x_3 x_4 + \beta_{13} x_1 x_3$ $+ \beta_{14} x_2 x_5)$ $+ c_2(\beta_{15} x_{11} + \cdots + \beta_{22} x_{18} + \beta_{23} x_{19}^2 + \beta_{24} x_{20}^2 + \beta_{25} x_{11} x_{12} + \beta_{26} x_{13} x_{14}$ $+ \beta_{27} x_{11} x_{13} + \beta_{28} x_{12} x_{15})$ $+ c_3(\beta_{29} x_{21} + \cdots + \beta_{36} x_{28} + \beta_{37} x_{29}^2 + \beta_{38} x_{30}^2 + \beta_{39} x_{31} x_{32} + \beta_{40} x_{23} x_{24}$ $+ \beta_{41} x_{21} x_{23} + \beta_{42} x_{22} x_{25})$ |
| GAM with global interactions (MC5L, MB5L) | $f(x) = \beta_0 + c_1(\beta_1|x_1| + \beta_2 x_2^2 + \beta_3 \log(|x_3| + 1) + \exp(\beta_4 x_4) + \beta_5(|x_5| + 1)^{-1} + \beta_6 x_1 x_2$ $+ \beta_7 |x_1 x_2 x_3| + \beta_8 \log(|x_3 + x_4 + x_5| + 1) + \beta_9 \max(x_4, x_5)$ $+ \exp[\beta_{10}(x_5 - x_3)])$ $+ c_2(\beta_{11}|x_6| + \beta_{12} x_7^2 + \beta_{13} \log(|x_8| + 1) + \exp(\beta_{14} x_9)$ $+ \beta_{15}(|x_{10}| + 1)^{-1} + \beta_{16} x_6 x_7 + \beta_{17} |x_6 x_7 x_8|$ $+ \beta_{18} \log(|x_8 + x_9 + x_{10}| + 1) + \beta_{19} \max(x_9, x_{10}) + \exp[\beta_{20}(x_{10} - x_8)])$ $+ c_3(\beta_{21}|x_{11}| + \beta_{22} x_{12}^2 + \beta_{23} \log(|x_{13}| + 1) + \exp(\beta_{24} x_{14})$ $+ \beta_{25}(|x_{15}| + 1)^{-1} + \beta_{26} x_{11} x_{12} + \beta_{27} |x_{11} x_{12} x_{13}|$ $+ \beta_{28} \log(|x_{13} + x_{14} + x_{15}| + 1) + \beta_{29} \max(x_{14}, x_{15})$ $+ \exp[\beta_{30}(x_{15} - x_{13})])$ |

---

[6] If MC7/MB7 or MC8/MB8 are to be used for any additional work, we recommend adding a small constant to the first additive components $\log(|\beta_1 x_1 + \cdots + \beta_{10} x_{10}|)$ and $\log(|\beta_1 x_1 + \cdots + \beta_5 x_5|)$, respectively, to ensure that the function is defined everywhere.

| | |
|---|---|
| Jumpy GAM with local interactions (MC6L, MB6L) | $f(x) = \beta_0 + c_1\big(\beta_1 x_1 \cdot I(|x_1| < 2) + \beta_2 x_2^2 \cdot I(x_2 > 1) + \beta_3 \log(|x_3| + 1) \cdot I(|x_3| > 1)$ $+ \exp(\beta_4 x_4) \cdot I(x_4 < 0) + \{\beta_5 \max[0.1, (|x_5| + 1)]\}^{-1}$ $+ \beta_6 \max(1, x_6 + x_7) + \beta_7 I(x_7 < 1) + \beta_8 I(|x_8| > 2) + \beta_9 x_9 \cdot I(x_9 < -1)$ $+ \beta_{10} \max(0, x_{10})\big)$ $+ c_2\big(\beta_{11} x_{11} \cdot I(|x_{11}| < 2) + \beta_{12} x_{12}^2 \cdot I(x_{12} > 1) + \beta_{13} \log(|x_{13}| + 1)$ $\cdot I(|x_{13}| > 1) + \exp(\beta_{14} x_{14}) \cdot I(x_{14} < 0) + \{\beta_{15} \max[0.1, (|x_{15}| + 1)]\}^{-1}$ $+ \beta_{16} \max(1, x_{16} + x_{17}) + \beta_{17} I(x_{17} < 1) + \beta_{18} I(|x_{18}| > 2) + \beta_{19} x_{19}$ $\cdot I(x_{19} < -1) + \beta_{20} \max(0, x_{20})\big)$ $+ c_3\big(\beta_{21} x_{21} \cdot I(|x_{21}| < 2) + \beta_{22} x_{22}^2 \cdot I(x_{22} > 1) + \beta_{23} \log(|x_{23}| + 1)$ $\cdot I(|x_{23}| > 1) + \exp(\beta_{24} x_{24}) \cdot I(x_{24} < 0) + \{\beta_{25} \max[0.1, (|x_{25}| + 1)]\}^{-1}$ $+ \beta_{26} \max(1, x_{26} + x_{27}) + \beta_{27} I(x_{27} < 1) + \beta_{28} I(|x_{28}| > 2) + \beta_{29} x_{29}$ $\cdot I(x_{29} < -1) + \beta_{30} \max(0, x_{30})\big)$ |
| Complex (MC8L, MB8L) | $f(x) = \beta_0 + c_1\big(\log(|\beta_1 x_1 + \cdots + \beta_5 x_5|) + \beta_6 x_1^2 x_6 + \beta_7 |x_2 x_3 x_7| + \beta_8 x_8 \cdot I(|x_8| > 2)$ $+ \beta_9 x_8 x_9 \cdot I(x_9 < -1) + \beta_{10} x_{10}^2 \cdot \max(0, x_{10})\big)$ $+ c_2\big(\log(|\beta_{11} x_{11} + \cdots + \beta_{15} x_{15}|) + \beta_{16} x_{11}^2 x_{16} + \beta_{17} |x_{12} x_{13} x_{17}| + \beta_{18} x_{18}$ $\cdot I(|x_{18}| > 2) + \beta_{19} x_{18} x_{19} \cdot I(x_{19} < -1) + \beta_{20} x_{20}^2 \cdot \max(0, x_{20})\big)$ $+ c_3\big(\log(|\beta_{21} x_{21} + \cdots + \beta_{25} x_{25}|) + \beta_{26} x_{21}^2 x_{26} + \beta_{27} |x_{22} x_{23} x_{27}| + \beta_{28} x_{28}$ $\cdot I(|x_{28}| > 2) + \beta_{29} x_{28} x_{29} \cdot I(x_{29} < -1) + \beta_{30} x_{30}^2 \cdot \max(0, x_{30})\big)$ |

### 3.3   Correlation Structures

To study the effects of correlation among the predictors, we used a block-diagonal covariance matrix $\Sigma$. The correlation blocks occurred in a 2×2, 3×3, … pattern as shown below. Only a subset of the cases were used. For $p = 10$, we considered two blocks total,

$$\Sigma = \begin{bmatrix} 1 & r & 0 & 0 & 0 & 0 & 0 & 0 & 0 & 0 \\ r & 1 & 0 & 0 & 0 & 0 & 0 & 0 & 0 & 0 \\ 0 & 0 & 1 & r & r & 0 & 0 & 0 & 0 & 0 \\ 0 & 0 & r & 1 & r & 0 & 0 & 0 & 0 & 0 \\ 0 & 0 & r & r & 1 & 0 & 0 & 0 & 0 & 0 \\ 0 & 0 & 0 & 0 & 0 & 1 & r & 0 & 0 & 0 \\ 0 & 0 & 0 & 0 & 0 & r & 1 & 0 & 0 & 0 \\ 0 & 0 & 0 & 0 & 0 & 0 & 0 & 1 & r & r \\ 0 & 0 & 0 & 0 & 0 & 0 & 0 & r & 1 & r \\ 0 & 0 & 0 & 0 & 0 & 0 & 0 & r & r & 1 \end{bmatrix}_{10 \times 10}$$

For $p = 25$, we had five blocks total, resulting in a $25 \times 25$ matrix $\Sigma$; and for $p = 50$, we had ten blocks total, resulting in $50 \times 50$ matrix $\Sigma$. We considered correlated predictors for functional forms MC2/MB2, MC5/MB5, MC6/MB6, and MC8/MB8 for the smaller functional forms and all of the larger functional forms. We considered two correlation levels $r = 0.25$ and $r = 0.5$ in addition to the uncorrelated models.

# 4  Summary of Findings

This section summarizes all the findings from our investigation. We elaborate on each finding in Section 5 and provide supporting empirical evidence.

**Predictive Performance and Related Aspects**:

**Finding 1:** (a) **Binary responses**: Predictive performances based on AUC were comparable among the three algorithms. While RF performed the worst, the loss in terms of AUC was modest.

(b) **Continuous responses**: All performance assessments were based on MSE. XGB and FFNN were competitive, with FFNNs performing better for smoother models, and XGBs performing better for non-smooth cases. RF objectively had the worst performance.

**Finding 2: Over- and Under-fitting:** We assessed over-fitting by the gap in performance metrics between training and testing data sets. A large gap is likely due to over-fitting as the model is too flexible and adapts to the local structure in the training set and does not generalize as well to the testing data. In our study, FFNN showed the least amounts of over-fitting, which is a likely result of the smooth, piecewise-linear nature of the fitted models. Tree-based methods exhibited more over-fitting behavior. They are piecewise constant and discontinuous at the boundary, and hence more jumpy. RF was worse than XGB, and likely suffered from a combination of both over-fitting and under-fitting. If we used deeper trees in RF to improve predictive performance, under-fitting could decrease in severity, but over-fitting could worsen.

We used a permutation-based variable importance technique to identify important variables.

**Finding 3: Identifying Important Variables:** In our empirical studies, all three methods correctly recovered the variables that were part of the underlying model. However, in practice, any method can miss active variables with small effects. XGB and FFNN had similar rankings and variable importance scores. However, RF tended to under-estimate the scores of less important variables.

**Recovering Input-Output Relationships In Terms of PDPs and Interactions**

**Finding 4: Influence of Variable Importance:** As expected, the ability to recover the true PDP curve of a variable varied with its importance score. The estimated PDP lines were closer to the truth when the importance score was high. While this held true for all three algorithms, it was particularly noticeable for RF.

**Finding 5: Over-Regularization Effect:** Tree-based algorithms tended to under- and over-estimate the true PDPs in the upper and lower tails, respectively, resulting in an over-regularization or "squeezing" effect. This effect was particularly large for RF.

**Finding 6:** (a) **RF's Failure to Recover True PDPs:** RF was consistently the worst in recovering the true PDPs, across both jumpy and smooth response surfaces.

(b) **Inability to Capture Interactions:** There were several instances where RF failed to capture true interactions. In addition, it falsely captured interactions that were not part of the true model.

Based on these findings, we removed RF from additional comparisons.

**Further Comparisons of XGB and FFNN on Identifying Input-Output Relationships Based on PDPs:**

**Finding 7:** (a) **Jumpy Response Surfaces:** XGB performed better in recovering functional forms with jumps, discontinuities, and sharp turns.

(b) **Smooth Response Surfaces:** FFNN performed better in recovering smooth functional forms.

**Finding 8: Interactions in Additive Index Models:** FFNN captured interactions in additive index models, while XGB generally failed in this area.

**Finding 9: Noise Variables:** Noise variables were generally captured correctly for both algorithms, although XGB did slightly better in this regard.

**Effect of Sample Size:**

**Finding 10:** As expected, the performances of all algorithms improved as sample size increased from 50k to 500k. This included predictive performance, reduction in variability across simulations, and reduction in bias and variance in the one-dimensional PDPs of individual predictors. In addition, bias and variance of PDPs for the noise terms decreased so that these terms were properly identified as unimportant.

**Response Types:**

**Finding 11:** As expected, for the same functional forms, the algorithms performed better with continuous responses in comparison to the binary responses. For binary response, the predictive performance was lower and there was higher bias and variability in the PDPs.

**Effect of Correlation:**

**Finding 12:** As expected, correlation inhibited the performance of all algorithms due to increasing lack of identifiability among features, which resulted in larger bias and variability in model predictions and interpretability tools.

Finally, we address an issue in the performance of the popular PDP.

**Model Bias as Measured by PDPs:**

**Finding 13:** All algorithms exhibited some bias in the PDPs even in pristine environments such as uncorrelated predictors. The problems were worse in small samples for binary responses.

# 5 Comprehensive Discussion of Empirical Evidence

The discussion section is organized in the same order as the findings in Section 4, and are now supported by empirical evidence.

## 5.1 Predictive Performance and Related Aspects

We use the visual summaries in Figures 1 to 6 to discuss the predictive performance of the three algorithms. The center dots in the figures show the average values from the eight simulations (four of which include correlated cases) and the widths of the bars show the variation ($1.96 \cdot sd$) across ten replicates. See Tables 13 to 24 in the Appendix in Section 8.3 for corresponding numerical values. The "best-" and "worst-in-class" results are bolded in these tables.

### 5.1.1 Finding 1a: Binary Responses

**Predictive performances based on AUC were comparable among the three algorithms. While RF performed the worst, the loss in terms of AUC was modest.**

Figure 1 shows testing AUC values (higher is better) for four different functional forms, with smaller sets of predictors, for different levels of correlation. The oracle AUC values are shown in red. Figure 2 gives corresponding results for models with a larger set of predictors.

*Figure 1: Performance metrics for binary response models MB2, MB5, MB6, and MB8*

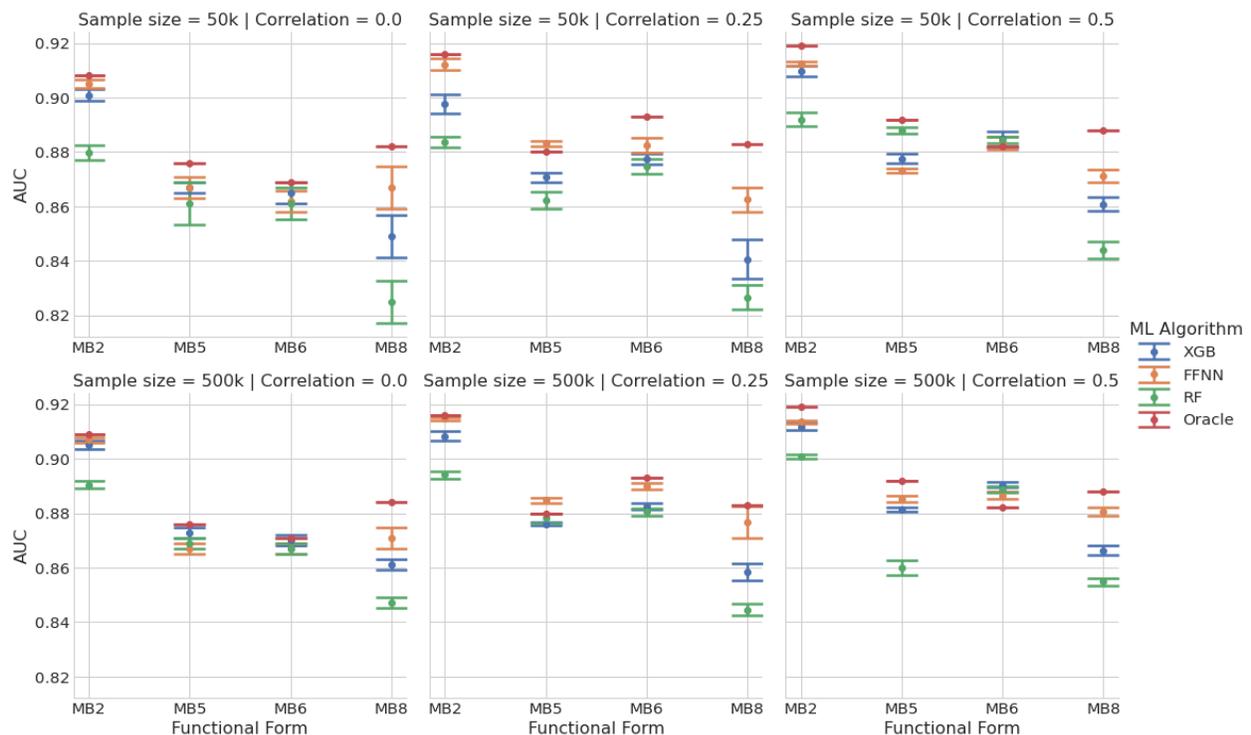

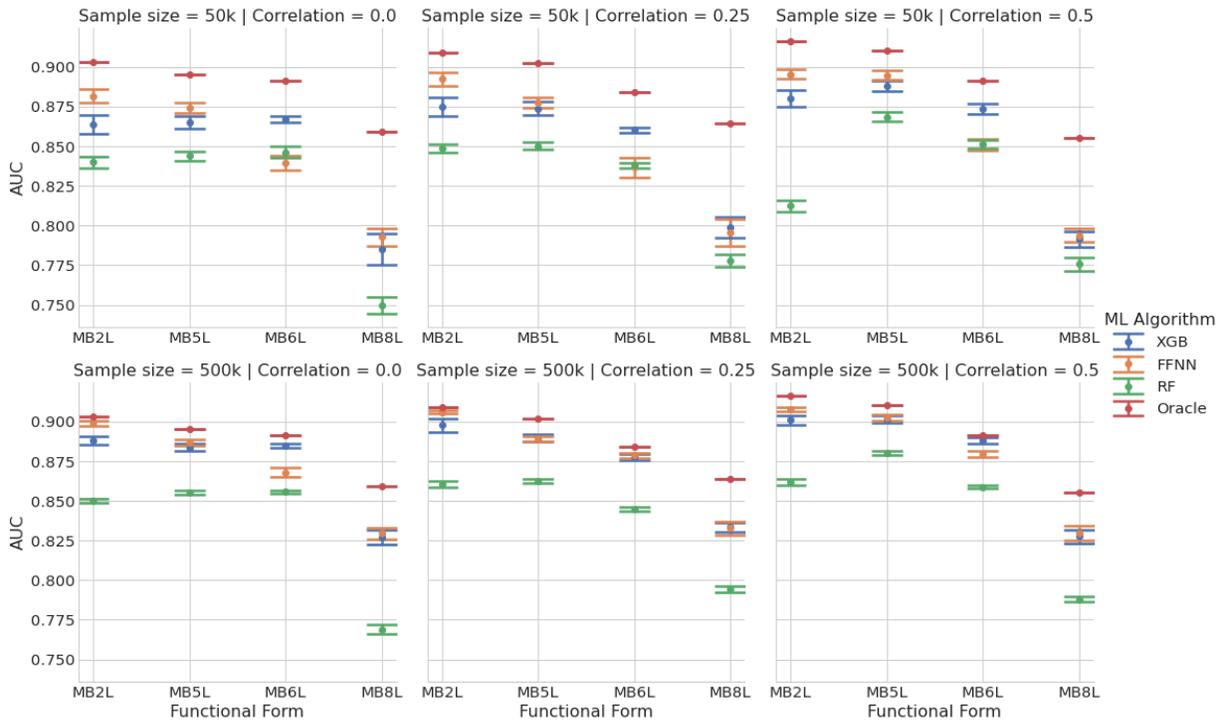
*Figure 2: Performance metrics for binary response models MB2L, MB5L, MB6L, and MB8L*

We can see that, overall, XGB (blue) and FFNN (orange) were competitive. XGB was better for MB6 (jumpy GAM with local interactions). FFNN was better for MB2 and MB5, which were, respectively, linear models with global interactions and a GAM with global interactions. MB8 had a combination of smooth forms and some discontinuities. For this, XGB and FFNN were competitive in some instances, while the latter was better in others. Note, however, that the differences in AUC values were small. RF (green) generally had lower AUCs, with the exception of MB6, where it was sometimes competitive. These behaviors have also been observed in the literature (Xiao, et al. 2019).

Turning to a comparison of Figures 1 versus 2 with larger versus smaller sets of predictors, MB2 and MB2L had similar oracle AUC values of around 0.9, but the testing AUC values were relatively worse for MB2L. There was also more variability. Note also the relative difference in performance between FFNN and XGB for MB8 versus MB8L. This suggests that FFNN may potentially have challenges in fitting a model with a larger number of predictors. We do not explore this issue further in this paper.

There were minor changes with correlation levels, but there was no consistent pattern.

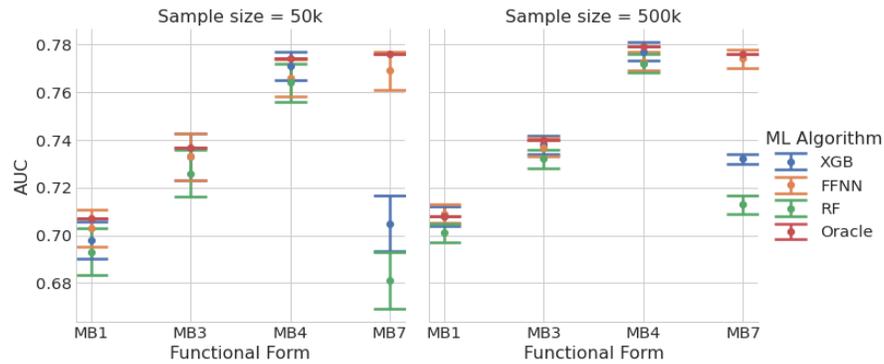

Figure 3: Performance metrics for binary response models MB1, MB3, MB4, and MB7

Figure 3 highlights the comparisons for MB1, MB3, MB4, and MB7 for the uncorrelated case. The first three models were additive: linear, GAM, and smooth plus jumpy GAM. MB7 was an additive index model (AIM) that had specialized forms of interactions. FFNN and XGB were competitive for the first three, but FFNN outperformed XGB for MB7. The performance of RF was reasonable for the first three, but poor for the AIM.

### 5.1.2  Finding 1b: Continuous Response

**All performance assessments were based on MSE. XGB and FFNN were competitive, with FFNNS performing better for smoother models and XGBs performing better for non-smooth cases. RF objectively had the worst performance.**

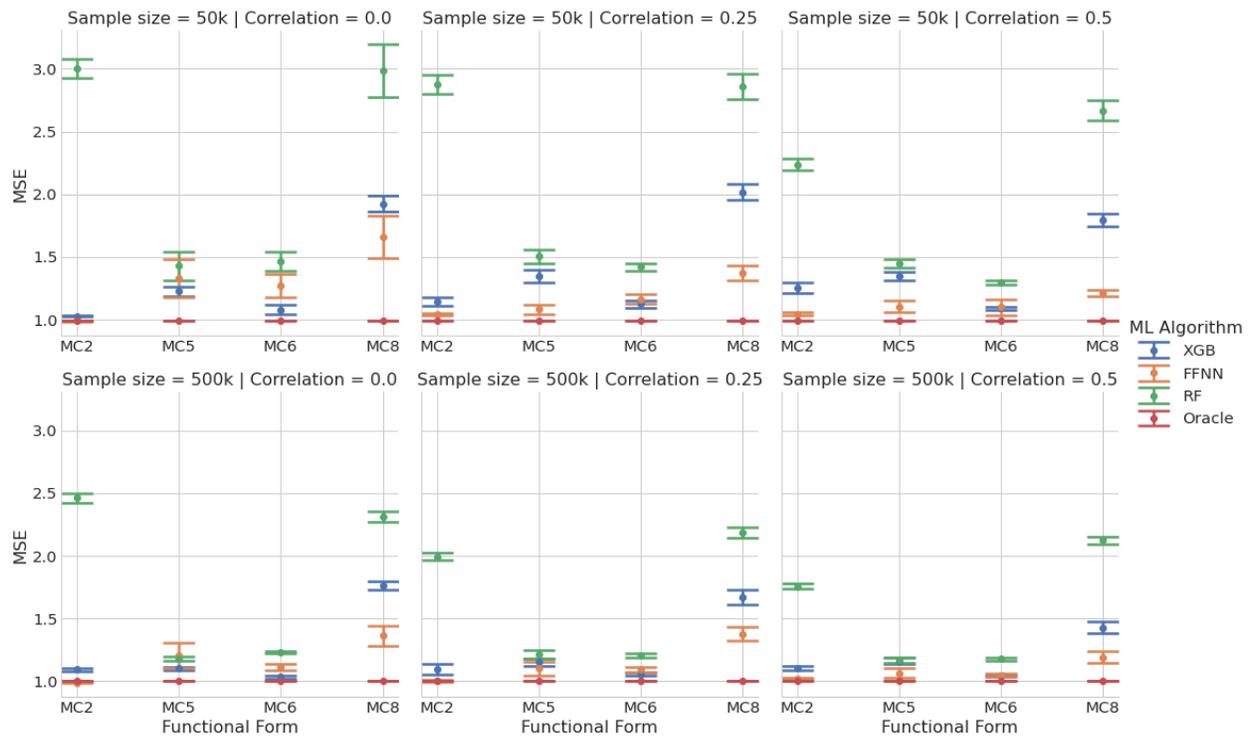

Figure 4: Performance metrics for continuous response models MC2, MC5, MC6, and MC8

*Figure 5: Performance metrics for correlated functional forms (continuous response) MC2L, MC5L, MC6L, and MC8L*

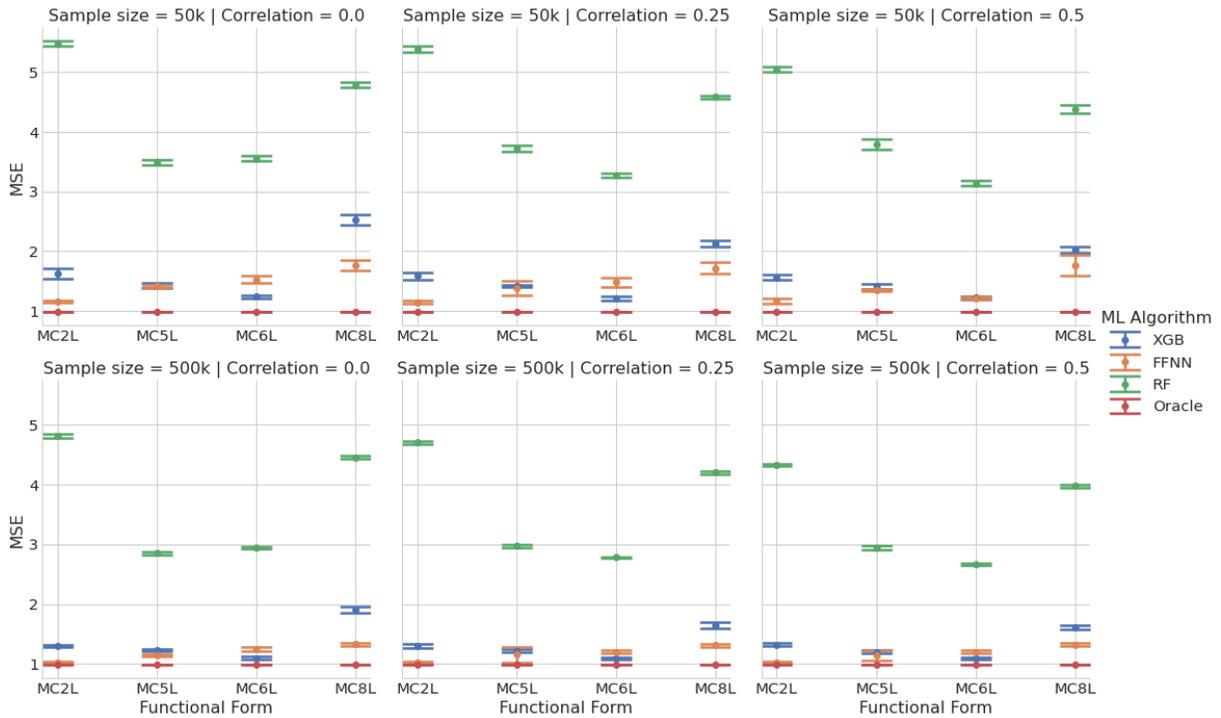

Figures 4, 5, and 6 provide testing MSE values (lower is better) for continuous responses. Compared to XGB, FFNN was better for MC2 (linear with global interactions) and competitive for MC5 (GAM with global interactions). FFNN was clearly better than XGB for MC8 (a combination of smooth forms and some discontinuities). XGB was better than FFNN for MC6 (jumpy GAM with local interactions). These conclusions were qualitatively similar to those for the binary responses. However, with the exception of MC5, RF was significantly worse than the other two models.

Comparing Figures 4 and 5, the MSE values for XGB and FFNN were about the same for models with smaller and larger sets of predictors. The numerical values in Tables 19 and 20 (in the Appendix) showed more switching of "best-in-class" performance between XGB and FFNN models. For RF, the MSE was significantly larger for models with a large number of predictors.

*Figure 6: Performance metrics for continuous response models MC1, MC3, MC4, and MC7*

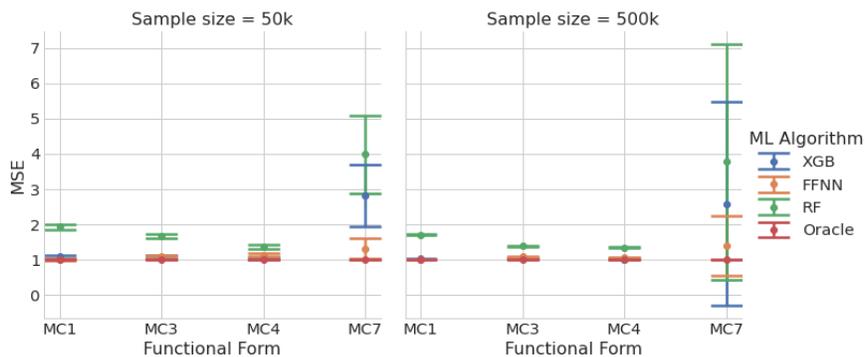

Figure 6 provides a comparison of MC1, MC3, MC4, and MC7 for the uncorrelated case. Recall that the first three correspond to additive models: linear, GAM, and smooth plus jumpy GAM. MC7 was an additive index model. As with the binary response case, FFNN and XGB were competitive for the first three models, but FFNN significantly outperformed XGB for MC7 (AIM). RF performed poorly in comparison, especially for MC7. Note also that, for $N = 500k$, MC7 had relatively large standard deviations. This could have been cause by the term $\log(|\beta_1 x_1 + \cdots + \beta_{10} x_{10}|)$ in MC7/MB7 to be undefined (i.e., the log additive component is singular).

## 5.2 Finding 2: Over- and Under-fitting

**We assessed over-fitting by the gap in performance metrics between training and testing data sets. A large gap is likely due to over-fitting as the model is too flexible and adapts to the local structure in the training set. In our study, FFNN showed the least amounts of over-fitting, which is a likely result of the smooth, piecewise-linear nature of the fitted models. Tree-based methods exhibited more over-fitting behavior. They are piecewise constant and discontinuous at the boundary, and hence more jumpy. RF was worse than XGB, and likely suffered from a combination of both over-fitting and under-fitting. If we used deeper trees in RF to improve predictive performance, under-fitting could decrease in severity, but over-fitting could worsen.**

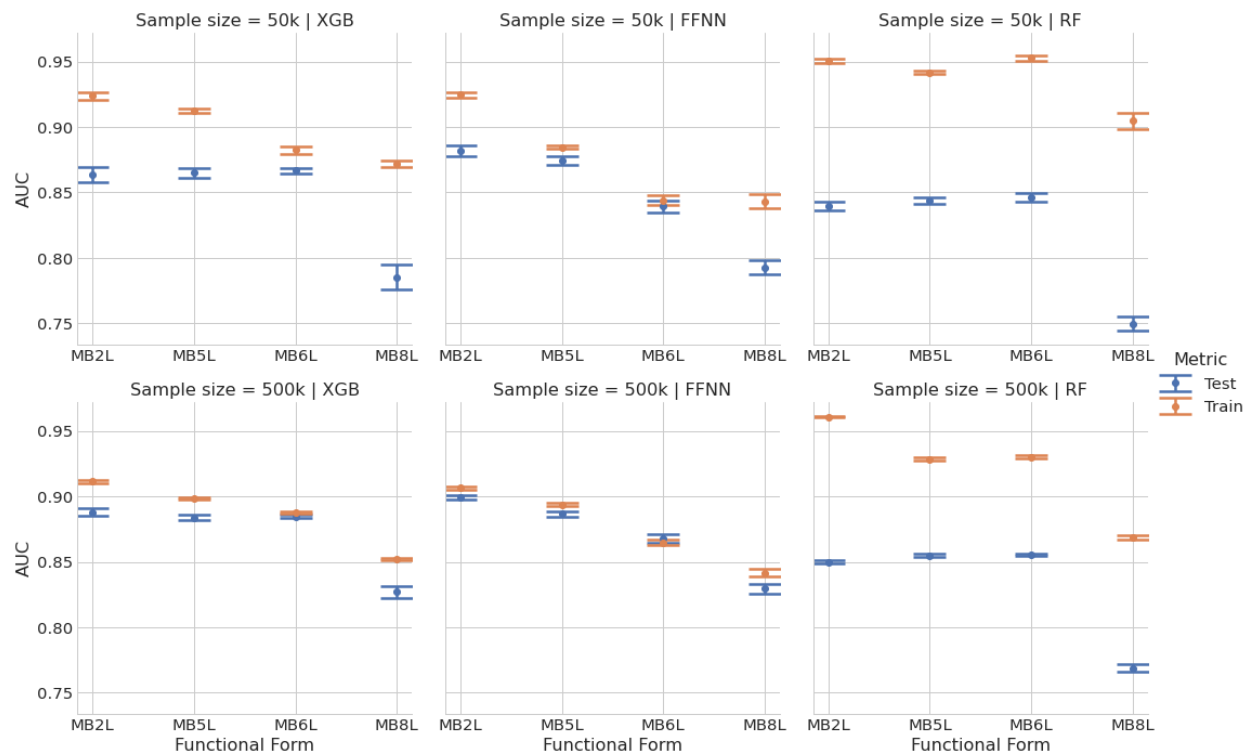

*Figure 7: Training and testing performance metrics for selected binary response models MB2L, MB5L, MB6L, and MB8L*

*Figure 8: Training and testing performance metrics for selected continuous response models MC2L, MC5L, MC6L, and MC8L*

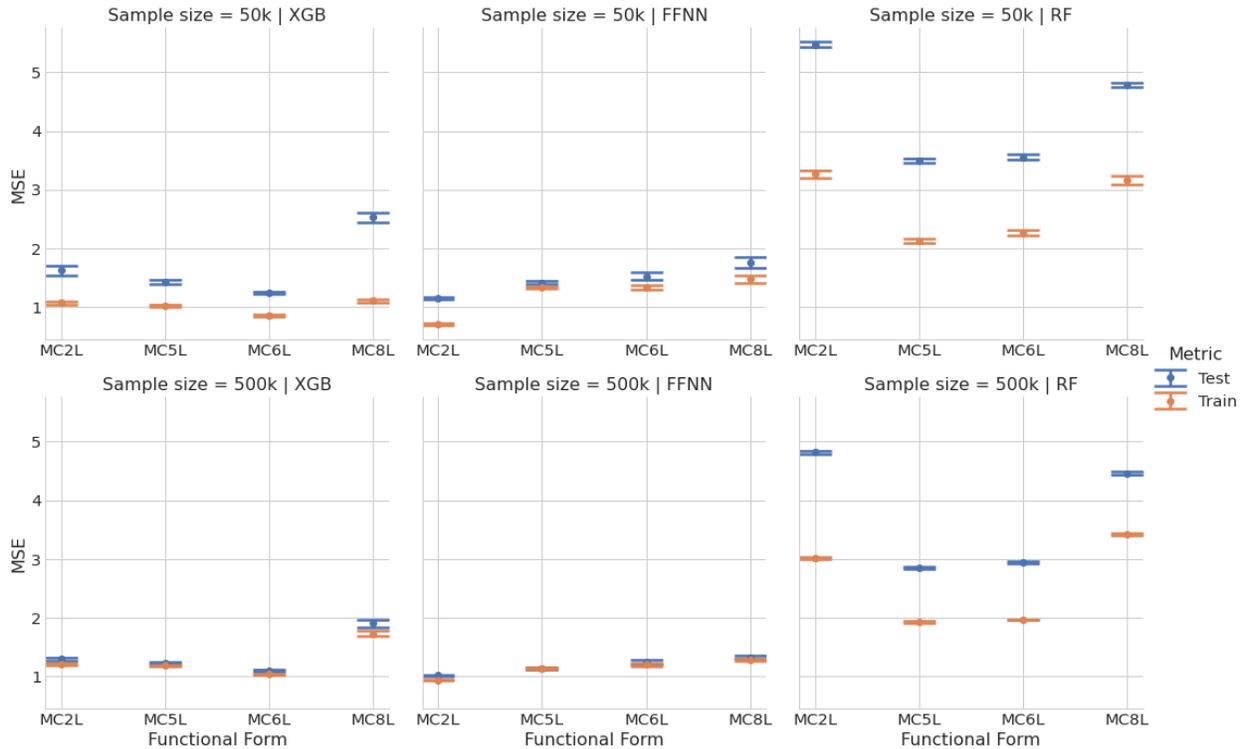

Figures 7 and 8 show the AUC values for training and testing data sets for the binary and continuous response cases, respectively.[7] RF had the largest gap. XGB and FFNN had less over-fitting, with FFNN performing better. RF also showed more over-fitting with larger sets of predictors. It is known (Hastie, Tibhsirani and Friedman 2009) that, when the number of variables is large, there is a smaller probability for each split that a relevant or important variable may be chosen.

It is possible to use regularization methods to reduce further over-fitting, but we did not explore these approaches in detail. FFNN and XGB were allowed to include L1 and L2 regularization hyperparameters via HPO, but did not always choose to set non-zero values.

### 5.3   Finding 3: Identifying Important Variables

**In our empirical studies, all three methods correctly recovered the variables that were part of the underlying model. However, in practice, all methods can miss active variables with small effects. XGB and FFNN had similar rankings and variable importance scores. However, RF tended to underestimate the scores of less important variables.**

Figure 9 shows the variable importance plots for the three algorithms for MB2L (linear with quadratic and second-order interactions) for both the uncorrelated and correlated case. There were 50 total predictors in this case, of which only 30 contribute to the response; the remaining

---

[7] See Tables 21 to 24 in Section 8.3.2.2 of the Appendix for the numeric values of the performance metrics.

20 were noise variables. Among the 30 contributing variables, there were three groups of 10 each with relative importances of 60%, 30%, and 10%. Only the 30 active predictors were identified as important by all three algorithms, but their rankings vary.

When correlation $r = 0.5$, XGB and RF algorithms followed the uncorrelated pattern exactly. However, the results for FFNN show some slight changes, where some noise variables show non-zero importance scores. However, this was not a significant concern overall.

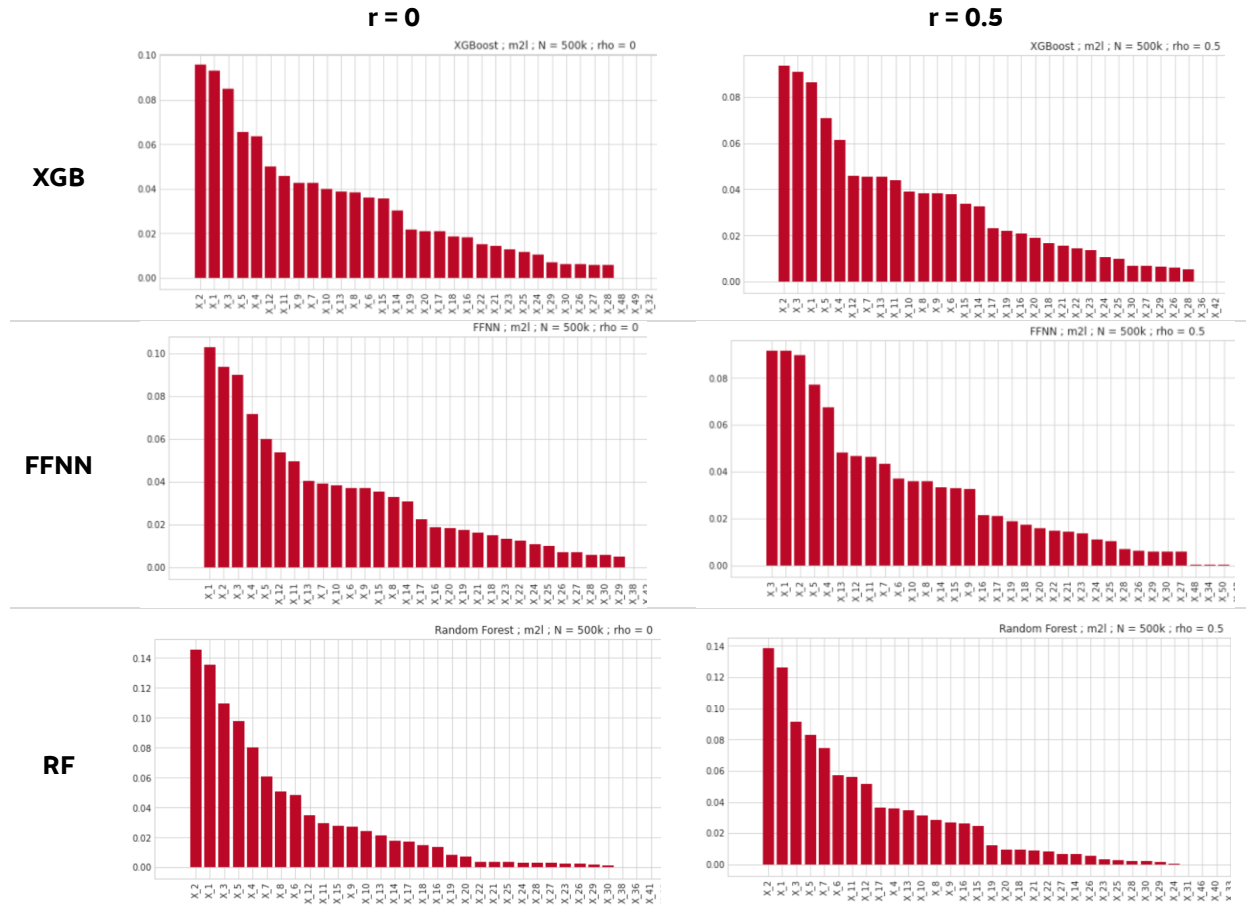

Figure 9: Variable importance for MB2L (linear with global interactions), N = 500k and r = 0 and r = 0.5

## 5.4 Recovering Input-Output Relationships In Terms of PDPs and Interactions

### 5.4.1 Finding 4: Influence of Variable Importance

**As expected, the ability to recover the true PDP curve of a variable varied with its importance score. The estimated PDP lines were closer to the truth when the importance score was high. While this held true for all three algorithms, it was particularly noticeable for RF.**

Figure 10 shows PDPs for $X_4$, $X_{14}$, and $X_{24}$ from MB2L (linear with global interactions). The variables had different levels of importance in the model. As we move from left to right and from top to bottom in the figure, the PDP curves deviate further from the true curve. The most faithfully fitted PDP curves sit in the top left of the figure for high importance and $r = 0$. The least faithfully fitted PDP curves sit in the bottom right of the figure for low importance and $r = 0.5$.

This indicates that as importance increased, so did the reproducibility of the true PDP curve. As correlation increased, the reproducibility of the true PDP curve decreased. Overall, the variable importance played a significant role.

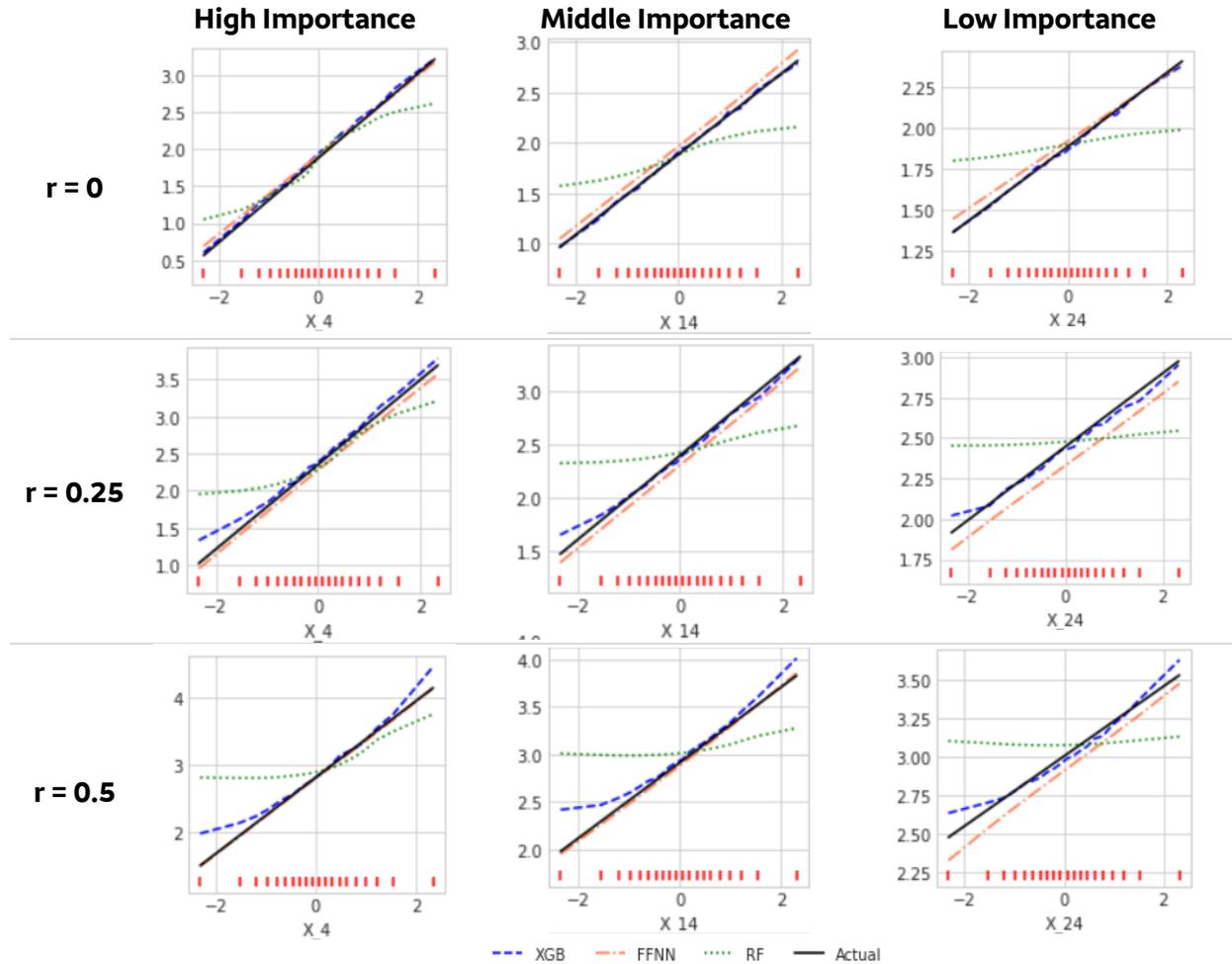

Figure 10: PDPs for MB2L (linear with global interactions) for three correlation levels, N = 500k

In addition, as true importance decreased, the relative bias increased for PDPs. Figure 11 shows an example from MB5L (GAM with global interactions), with three variables $X_3, X_8$, and $X_{13}$ of different levels of importance. As importance decreased, the PDP curve lost some of its distinctive shape and deviated further from the true PDP curve. This aligned with the influence of variable importance on predictors.

Figure 12 decomposes the deviation (error) of the PDPs in Figure 11 into bias and variance components. Bias (blue line) is the largest contributor to the overall error (red line).

Figure 11: PDPs for three variables ($X_3$, $X_8$, and $X_{13}$) of different importance levels for uncorrelated MB5L (GAM with global interactions), N = 50k; **bold red line is the true PDP**; lighter red lines are PDPs from the ten replicates

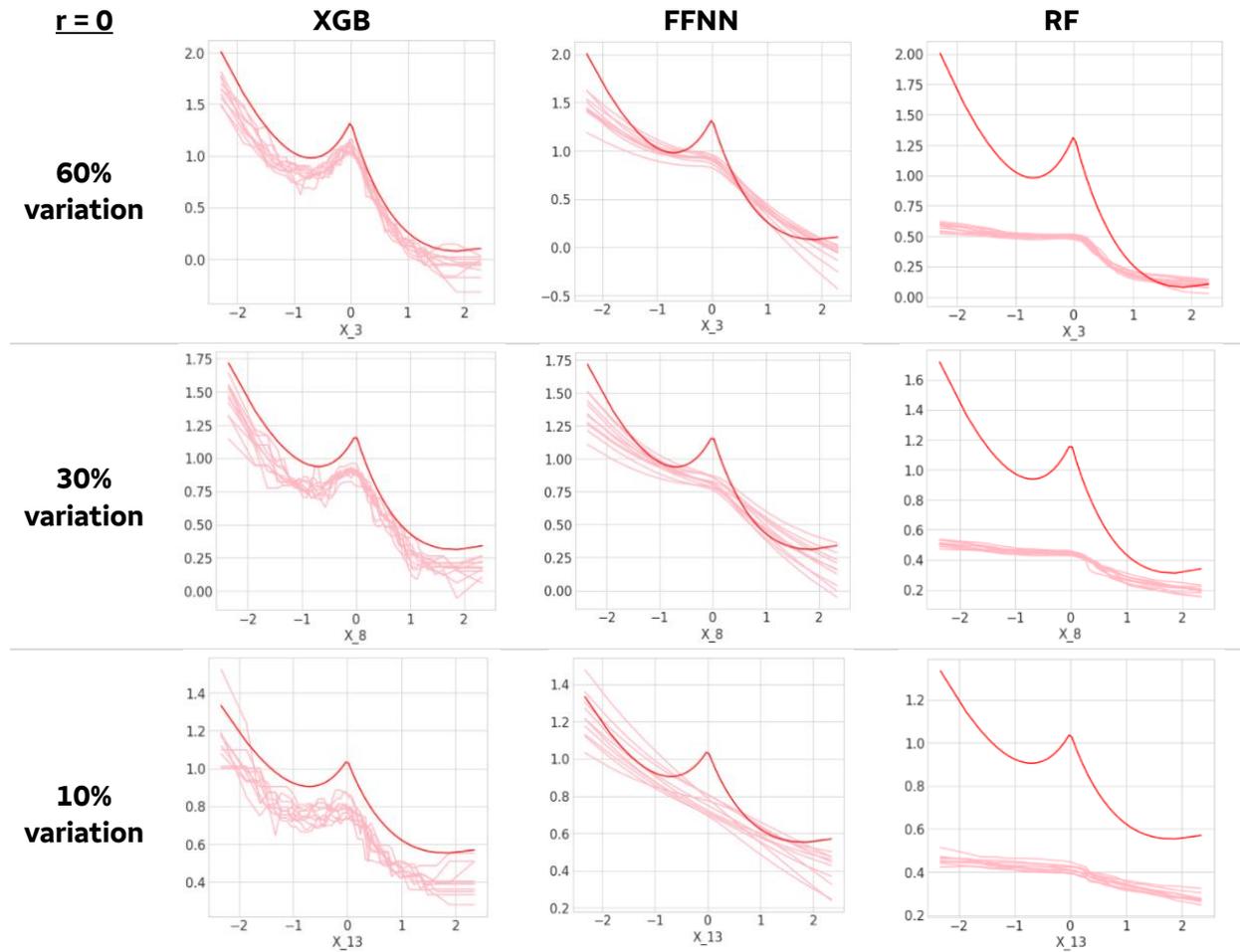

Figure 12: Bias-variance decomposition for three variables ($X_3$, $X_8$, and $X_{13}$) of different importance levels for uncorrelated MB5L (GAM with global interactions), N = 50k

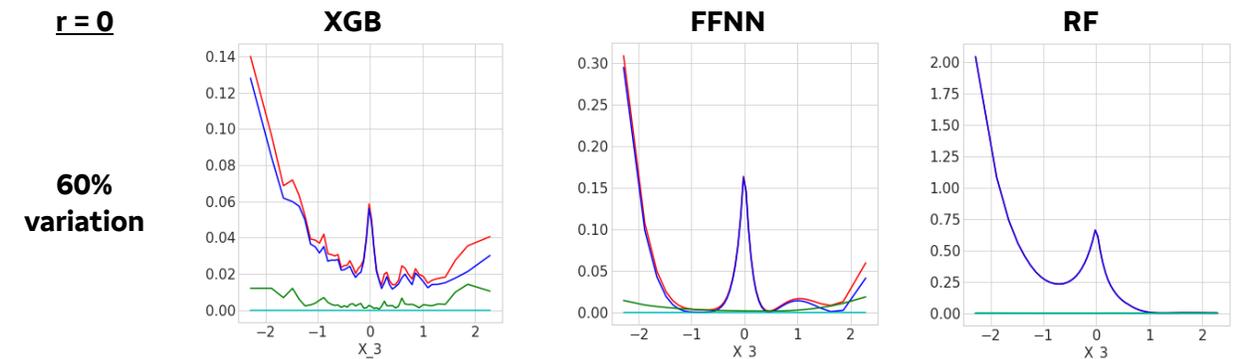

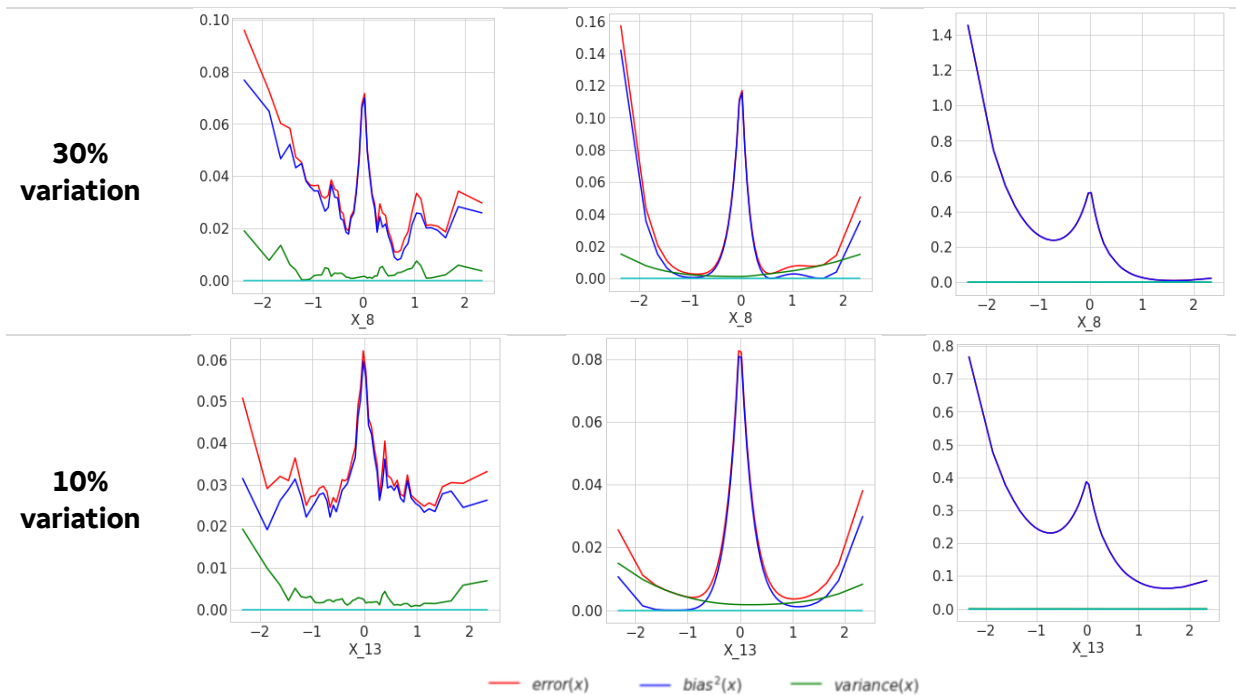

### 5.4.2 Finding 5: Over-Regularization Effect

**Tree-based algorithms tended to under- and over-estimate the true PDPs in the upper and lower tails, respectively, resulting in an over-regularization or "squeezing" effect. This effect was particularly large for RF.**

RF and XGB fit local constant models for each tree, whereas FFNN fits local linear models (e.g., the ReLU activation function). As such, tree-based methods suffer from the type of local model fitted. The effect was particularly evident for binary response cases for RF models due to the bagging algorithm.

Since RFs rely on averaging trees in the forest, the models can have difficulty making probability predictions near probabilities 0 and 1 (Caruana and Alexandru, Predicting good probabilities with supervised learning 2005). If individual trees make varying predictions, the average prediction from the entire forest moves away from the probabilities 0 and 1. We do not observe this in XGB as boosting does not rely on the same algorithmic mechanics. XGB fits shallow trees with larger bias in each tree, and boosting corrects for some or most of those biases.

Figure 13 shows the effect for the binary and continuous response cases MB1 and MC1 (linear), respectively. Figure 14 shows the effect for RF, but not for XGB, which tended to exhibit the behavior with binary responses rather than continuous responses.

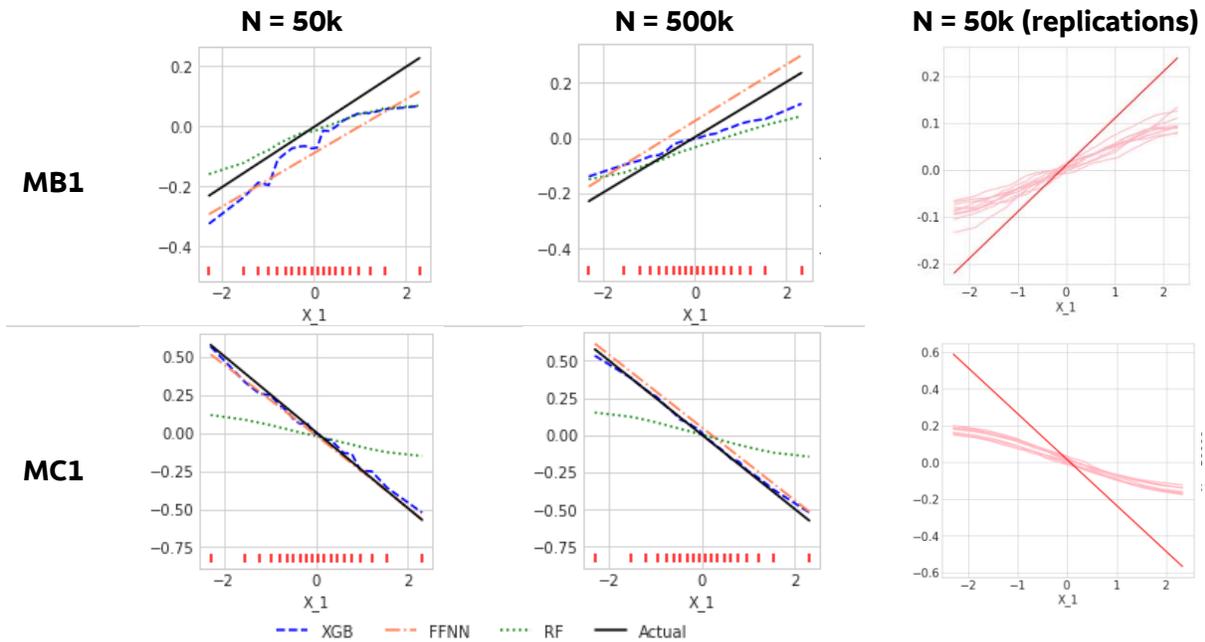

*Figure 13: PDP for MB1/MC1 (linear), focusing on variable $X_1$; (left) N = 50k; (middle) 500k; (right) N = 50k, RF model replications;* **bold red line is the true PDP**; *lighter red lines are PDPs from the ten replicates*

*Figure 14: True vs. fitted continuous response for uncorrelated MC1 (linear), sample size N = 500k; (left) XGB model; (right) RF model*

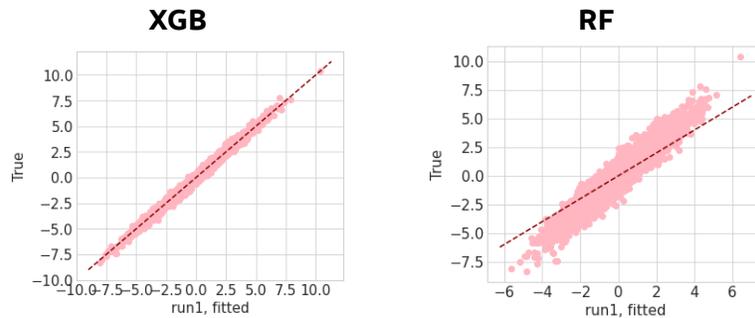

As correlation increased, the over-regularization effect became more severe, even accounting for sample size. Figure 15 shows the increasingly stronger effect as correlation grows. A larger sample size may help to offset the over-regularization effect to a certain extent, but it does not necessarily counteract the effect of correlation.

Tree-based methods are greedy when choosing a splitting variable, as the method will choose variables that contribute the most every time (i.e. important variables). As correlation increases, less important variables will appear deeper within the tree, after the more important variables have already been included. Weaker signals may not be fully captured, which may result in an underestimation of the variable's true importance and overall impact on the predicted response from the fitted model.

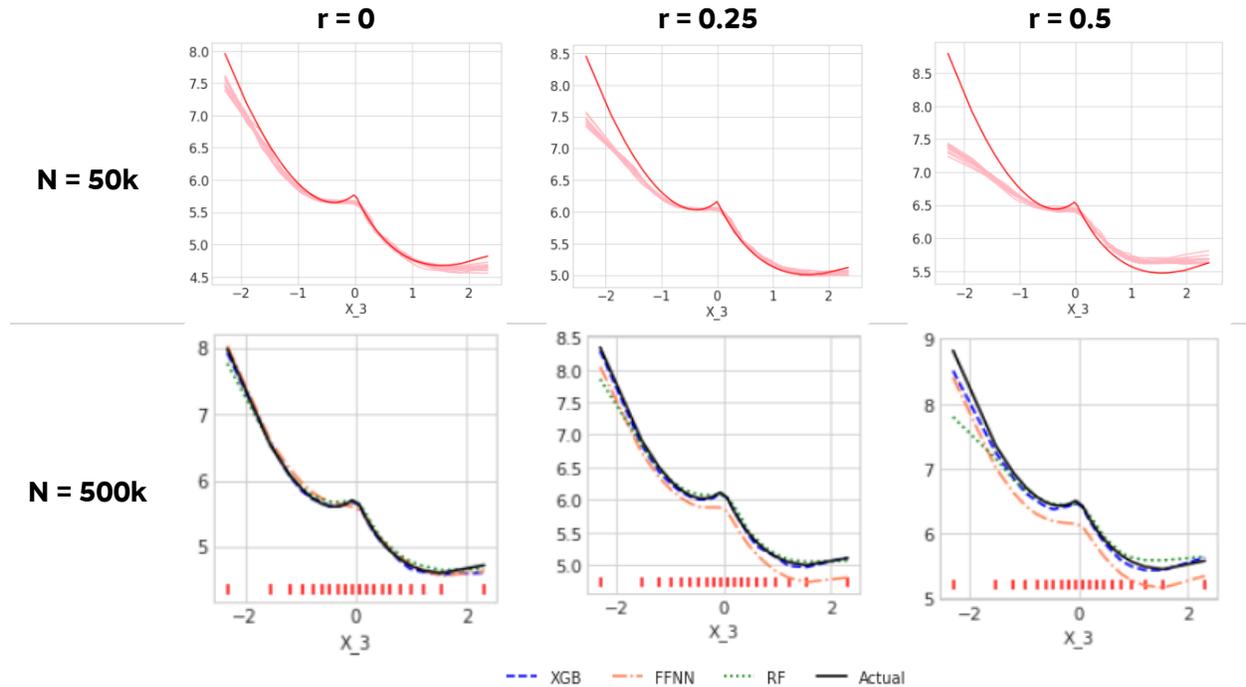
*Figure 15: PDPs for $X_3$ from uncorrelated MC5 from the RF model for sample sizes N = 50k and N = 500k and three correlation levels; (top row)* **bold red line is the true PDP**; *lighter red lines are PDPs from the ten replicates*

Bias in the probability scores could be due to the over-regularization effect in binary response cases, which can be diagnosed with reliability plots. Probability calibration approaches (e.g., Platt scaling, isotonic regression, spline calibration, etc.) can correct for bias in probability scores (Caruana and Niculescu-Mizil, An empirical comparison of supervised learning algorithms 2006). However, these methods complicate the interpretability of the ML model.

### 5.4.3 Finding 6a: RF's Failure to Recover True PDPs

**RF was consistently the worst in recovering the true PDPs, across both jumpy and smooth response surfaces.**

Figure 16 shows that XGB and FFNN captured the overall behavior of $X_7$ from MB6 (jumpy GAM with local interactions), with some small bias, particularly from $X_7 > 1$; however, the bias was more significant for RF. Increasing sample size improved the behavior of the fitted PDP curve.

*Figure 16: PDP for uncorrelated MB6 (jumpy GAM with local interactions), focusing on variable $X_7$; (top left) N = 50k, (top right) 500k, all three SML algorithms; (bottom left) N = 50k, XGB model; (bottom middle) N = 50k, FFNN model; (bottom right) N = 50k, RF model; **bold red line is the true PDP**; lighter red lines are PDPs from the ten replicates*

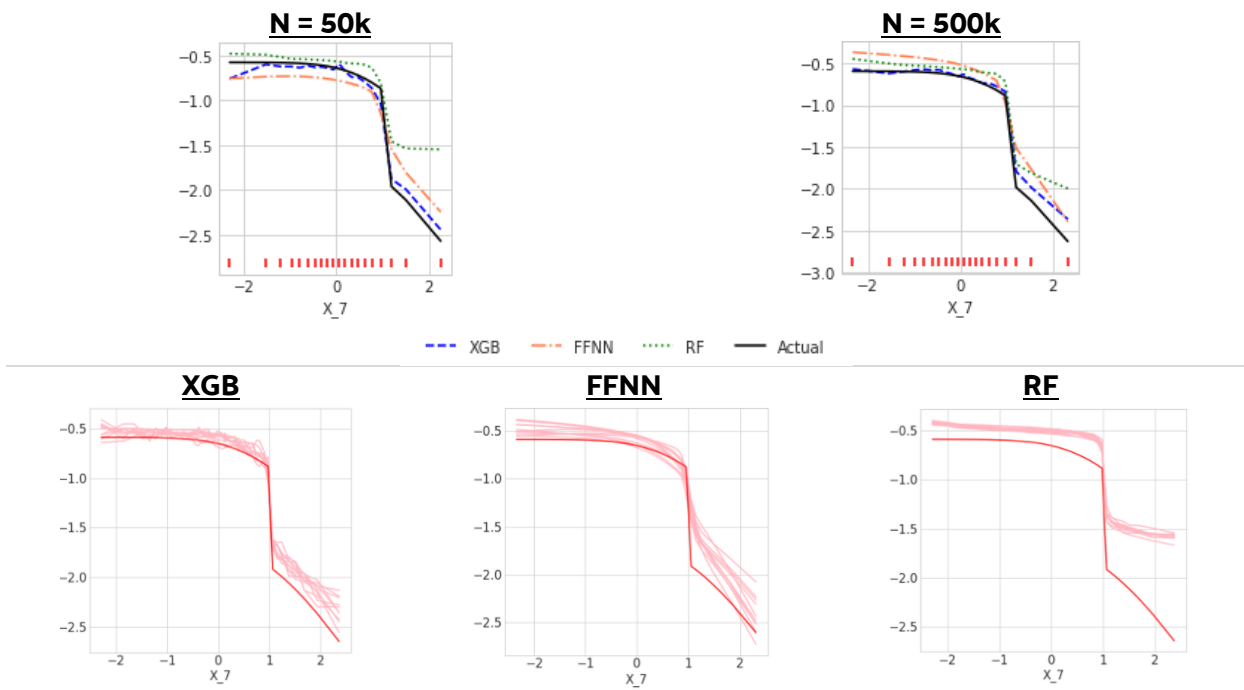

Figure 17 shows that XGB and FFNN captured the quadratic behavior of $X_{10}$ from MB2 (linear with global interactions). However, RF showed a severe flattening of the quadratic shape, which could be attributed to the over-regularization effect (see Section 5.4.2). In this case, the problem does not disappear with increasing sample size.

*Figure 17: PDP for uncorrelated MB2 (linear with global interactions), focusing on variable $X_{10}$; (top left) N = 50k, (top right) 500k, all three SML algorithms; (bottom left) N = 50k, XGB model; (bottom middle) N = 50k, FFNN model; (bottom right) N = 50k, RF model; **bold red line is the true PDP**; lighter red lines are PDPs from the ten replicates*

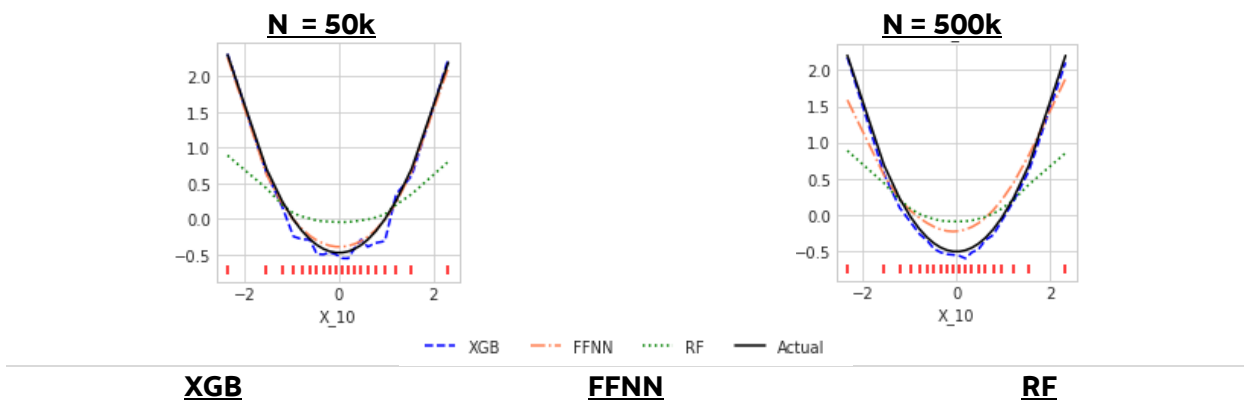

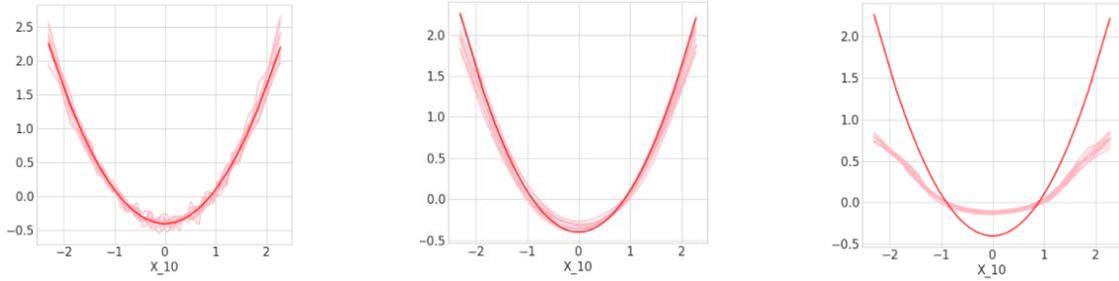

For noise variables that were not part of the model, PDPs for RF exhibited bias and false structure. The PDPs showed bias for RF, which increased as correlation increased.

Figure 18 shows that as correlation increased, the bias increased for noise variable $X_{35}$ in MC8L/MB8L (complex model).

*Figure 18: Noise variable $X_{35}$ in MC8L/MB8L (complex model) for RF models, N = 50k, three correlation levels;* **bold red line is the true PDP**; *lighter red lines are PDPs from the ten replicates*

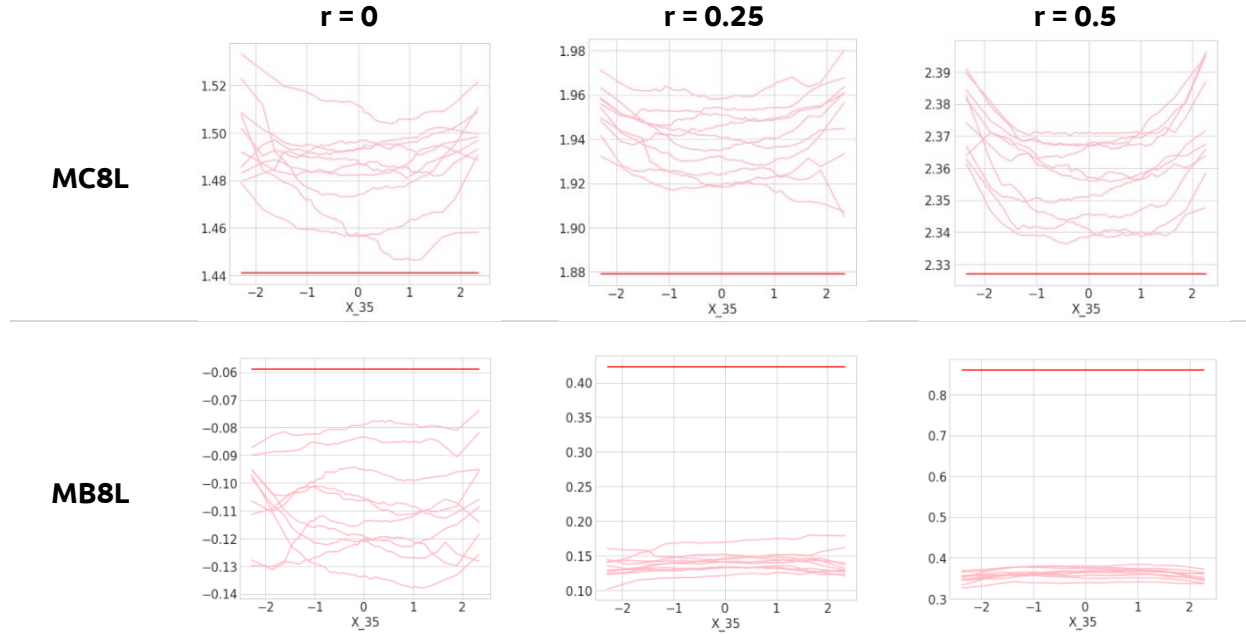

### 5.4.4 Finding 6b: Inability to Capture True Interactions

**There were several instances where RF failed to capture true interactions. In addition, it falsely captured interactions for that were not part of the true model.**

Figure 19 shows the variable importance plots from MC2 (linear with global interactions), RF returned scores that were different from those for XGB and FFNN, which put more weight on the top two interacting variables $X_2$ and $X_5$. For RF, one ($X_7$) of the top five variables was a non-interacting variable. Equal importance was given to all five most important variables for RF, resulting in more evenly weighted importance scores.

*Figure 19: Variable importance of all three SML models for uncorrelated MC2 (linear with global interactions), sample size N = 500k; (left) XGB model; (middle) FFNN model; (right) RF model*

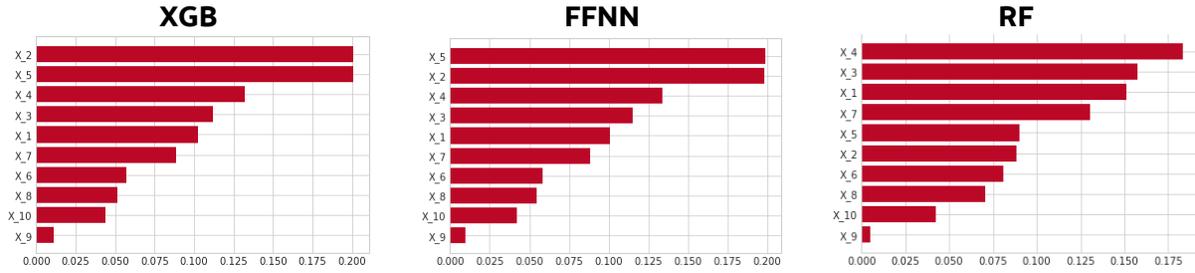

Figure 20 shows the H-statistics for MC8 (complex model), which are characterized by a gradient from green to red (high to low score of the presence of an interaction, respectively). The interactions between variables in the pairs $(X_2, X_4)$ and $(X_2, X_5)$ were truly present in the model. Both pairs in Figure 20 have relatively small H-statistic values for RF in comparison to those for XGB and FFNN.

As interactions increased in complexity, the trees in RF had fewer opportunities to split on the relevant features. For example, if there was an interaction between two features $X_1$ and $X_2$, the tree would first need to split on $X_1$ and then, at splits deeper in the tree, on $X_2$ differently for different levels of $X_1$. However, if $X_1$ was not split on at all, then the interaction, which we could define as the impact of $X_2$ on the response depends on the value of $X_1$, would not be captured. In some cases, for correlated features, a lower order effect may capture the same information as a higher order effect (Wright, Ziegler and König 2016).

*Figure 20: Unnormalized H-statistics of all three methods for uncorrelated MC8 (complex model), sample size N = 500k; (top left) XGB model; (top right) FFNN model; (bottom row) RF model*

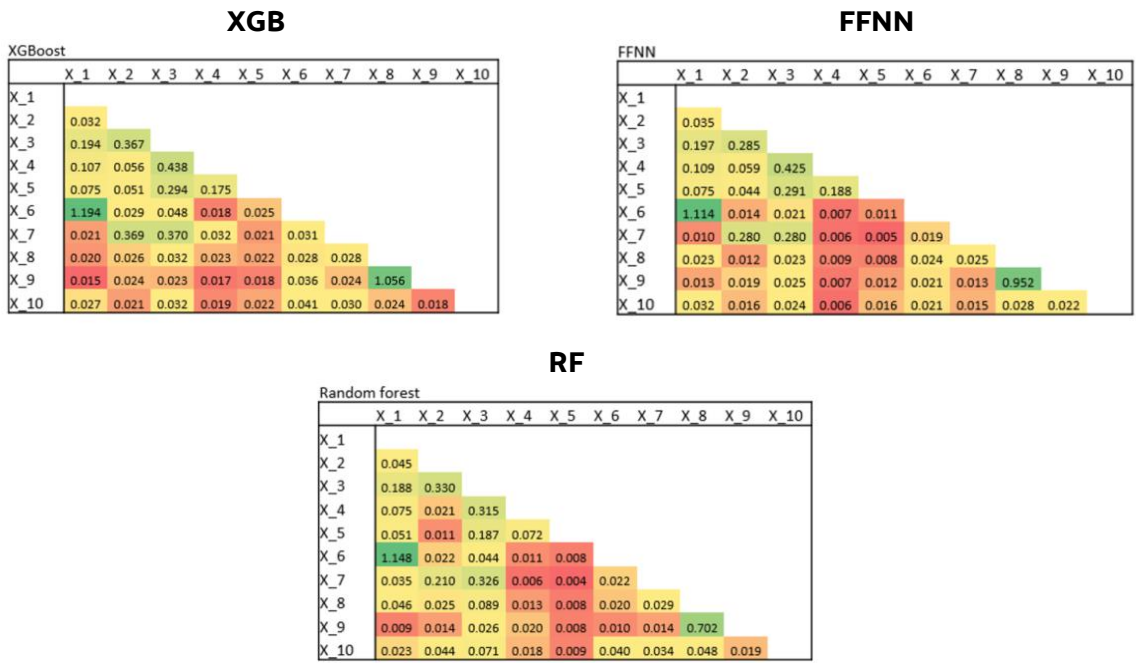

Tree-based models, in particular RF, could also induce artificial interactions. The CICE[8] plots for XGB and RF showed the presence of interactions that do not exist in the data. We do not see the same behavior for FFNN. Figure 21 shows the CICE plots for MC1 (linear), which suggested no interactions for FFNN; however, the plots for XGB and RF suggested the presence of an artificial interaction term. Bias in the model could result in splitting on one or more incorrect variables, culminating in the presence of artificial interactions. From the H-statistics for MC8 (complex model) in Figure 20, there was no interaction between $X_1$ and $X_7$, but the H-statistics for RF could suggest an interaction, relative to the magnitude of the H-statistics from the other models.

*Figure 21: CICE plots for uncorrelated MC1 (linear), focusing on variables $X_1$ (top row) and $X_2$ (bottom row), sample size N = 500k; (first column) XGB model CICE plots; (second column) FFNN model CICE plots; (third column) RF model CICE plots; (fourth column) actual CICE plots*

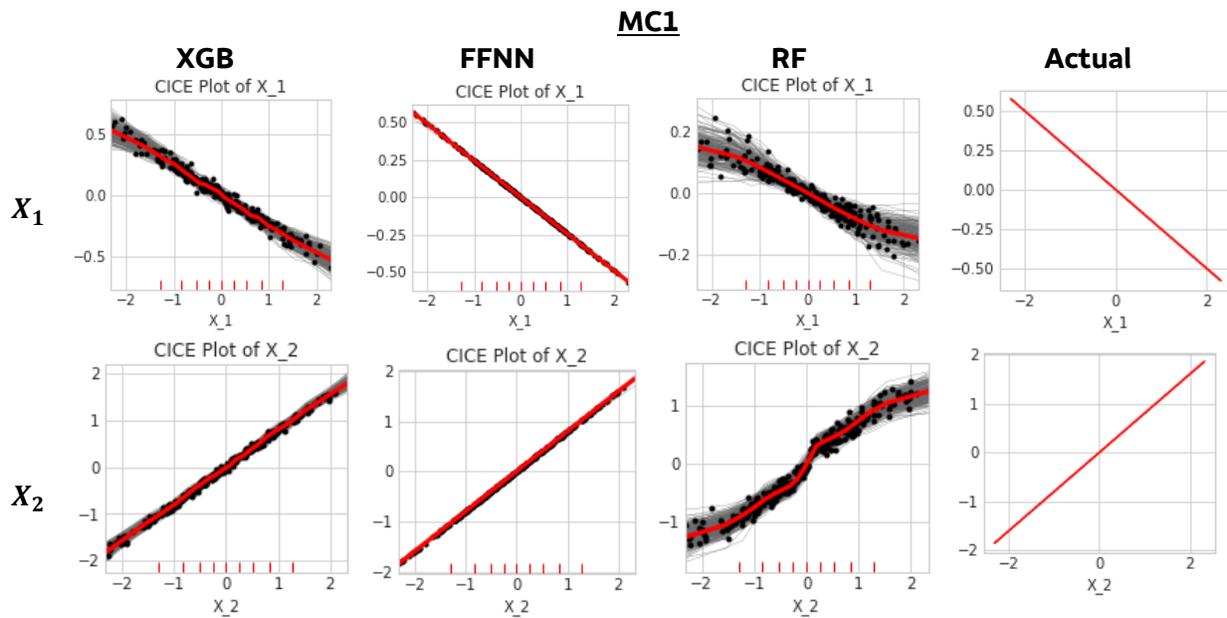

Overall, the presence and severity of artificial interactions increased as correlation increased. Figure 22 shows $X_8$ from MC2 (linear with global interactions), in which patterns (i.e., curves in the CICE lines) in the plots for RF and XGB indicated the possible presence of interactions.

---

[8] ICE plots display how each observation's prediction changes when a particular variable changes. Each line on an ICE plot represents a single observation in a data set of $N$ observations. The centered ICE or CICE plot centers the plot by anchoring the observations at a certain point in the predictor, and only displays the difference in the prediction to the chosen point. For additional details, please refer to (Goldstein 2015).

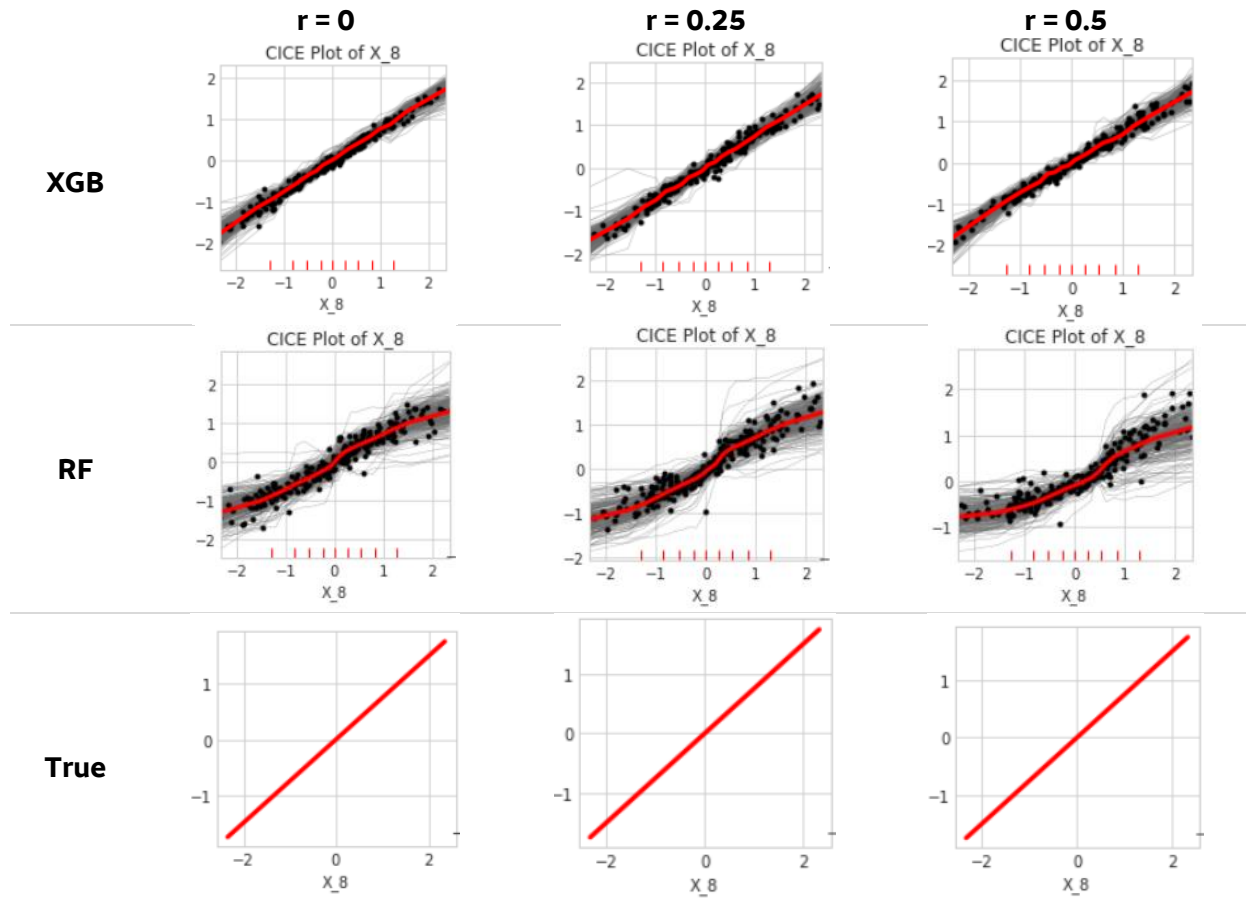

*Figure 22: CICE plots for $X_8$ from MC2 (linear with global interactions) at three correlation levels, N = 500k*

## 5.5 Further Comparison of XGB and FFNN on Identifying Input-Output Relationships Based on PDPs

### 5.5.1 Response Surfaces

#### 5.5.1.1 Finding 7a: Jumpy Response Surfaces

**XGB performed better in recovering functional forms with jumps, discontinuities, and sharp turns.**

Figure 23 shows the PDPs for MC6/MB6 (jumpy GAM with local interactions). Variable $X_3$ had a sharp turn at zero from an indicator function, which FFNN had difficulty capturing, but XGB captured well. FFNN improved when sample size is increased to $N = 500k$. Variable $X_7$ in MB6 also had a discontinuity due to a max($\cdot$) function, which was more accurately captured by XGB. The behavior was consistent across sample size.

*Figure 23: PDP for uncorrelated MC6/MB6 (jumpy GAM with local interactions), focusing on variable $X_3$ and $X_7$, respectively; (left) N = 50k, (second to the left) 500k, all three SML algorithms; (second from the right) N = 50k, XGB model; (right) N = 50k, FFNN model; **bold red line is the true PDP**; lighter red lines are PDPs from the ten replicates*

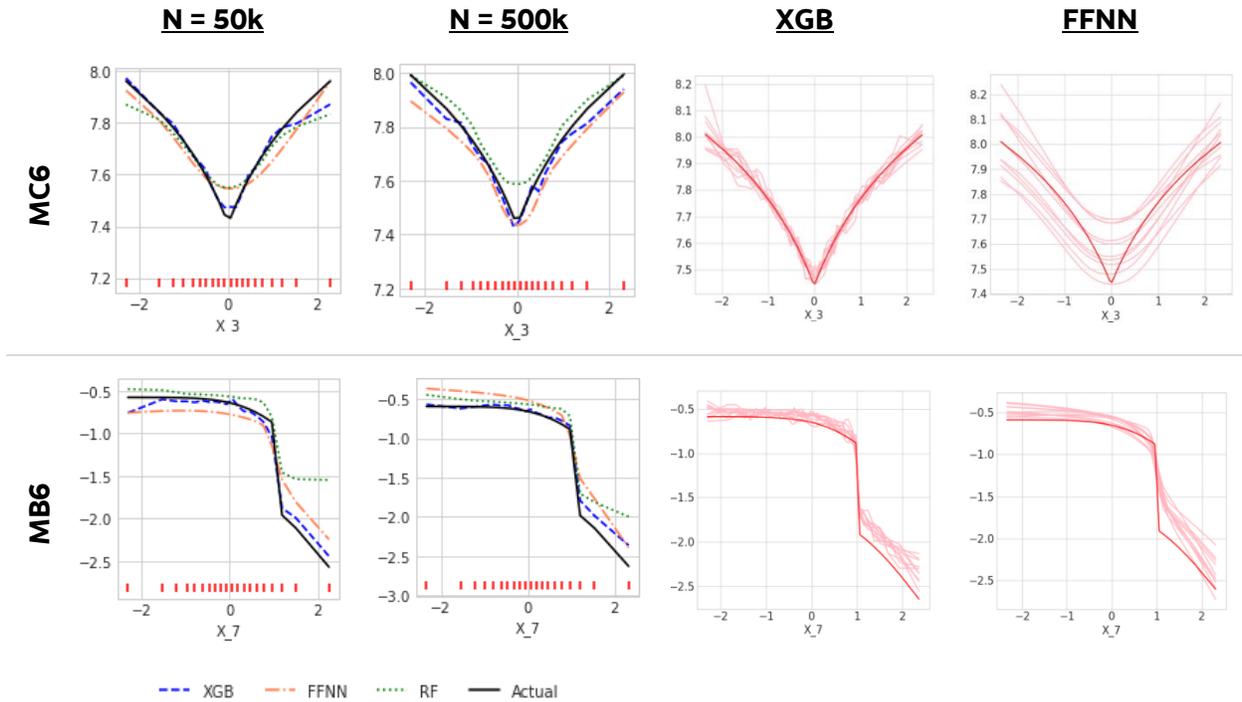

XGB continued to outperform FFNN as correlation increased. Figures 24 and 25 show the PDPs for $X_3$ from MC6. As sample size increased, FFNN became more faithful to the true PDP curve; however, XGB still did better regardless of sample size. Figure 25 shows that FFNN smoothed out sharp turns in the response surface for larger correlation levels. Although it appears FFNN decreased in variability as correlation increases, this phenomenon is not observed in all cases.

*Figure 24: PDPs for $X_3$ from MC6 (jumpy GAM with local interactions) for three correlation levels, N = 50k and N = 500k*

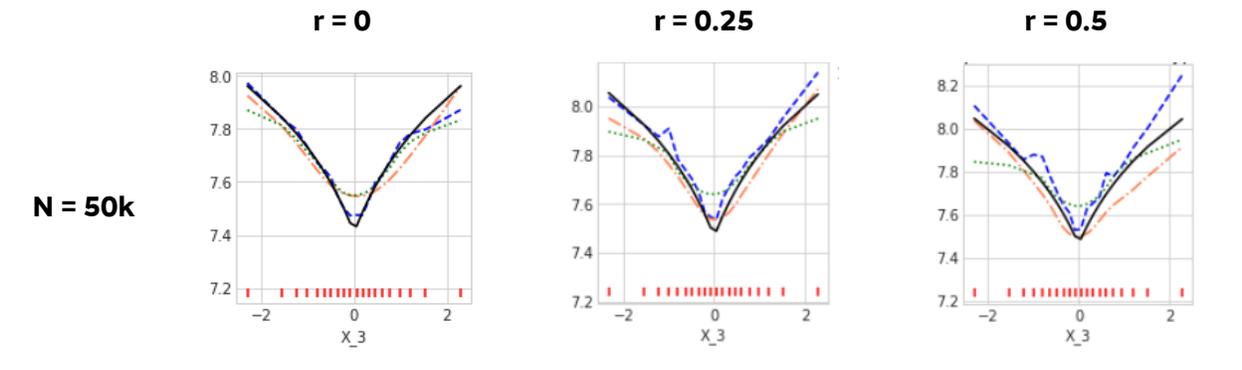

**N = 500k**

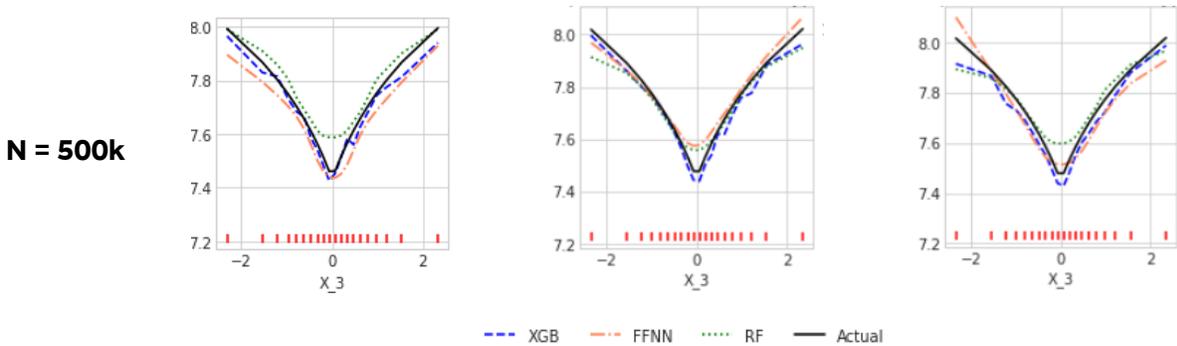

Figure 25: PDPs for $X_3$ for MC6 (jumpy GAM with local interactions) for three correlation levels for XGB and FFNN models, N = 50k; **bold red line is the true PDP**; lighter red lines are PDPs from the ten replicates

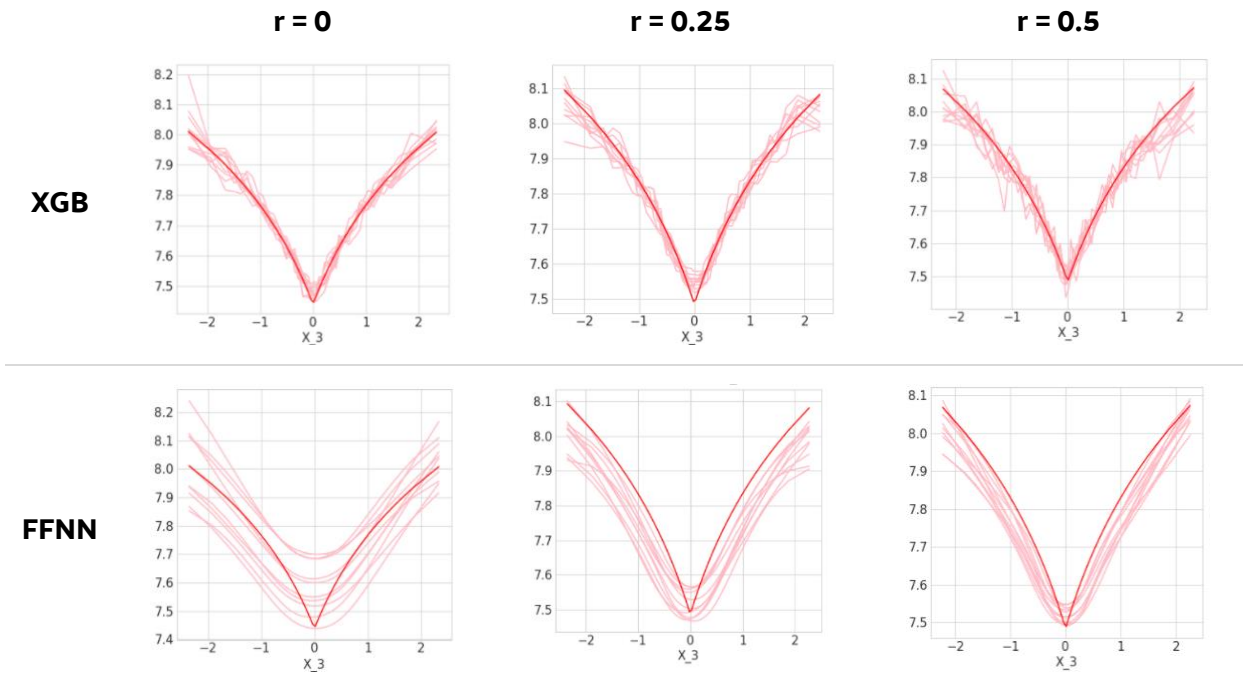

#### 5.5.1.2 Finding 7b: Smooth Response Surfaces
**FFNN performed better in recovering smooth functional forms.**

FFNN uses the ReLU activation function, which creates local linear models, allowing a smooth transition from one local region of the data to the next. FFNN can fit smoother surfaces with more accuracy than XGB may be able to. Figure 26 shows FFNN capturing the quadratic curve from the $X_{10}$ term in MC2/MB2 (linear with global interactions), while XGB had some difficulty picking up the smoothness of the curve. The binary response case exhibited more bumps along the surface than was seen for FFNN, which was a less accurate representation of the true, smooth quadratic curve.

*Figure 26: PDP for uncorrelated MC2/MB2 (linear with global interactions), focusing on variable $X_{10}$; (left) N = 50k; (second from the left) 500k, all three SML algorithms; (second from the right) N = 50k, XGB model; (right) N = 50k, FFNN model; **bold red line is the true PDP**; lighter red lines are PDPs from the ten replicates*

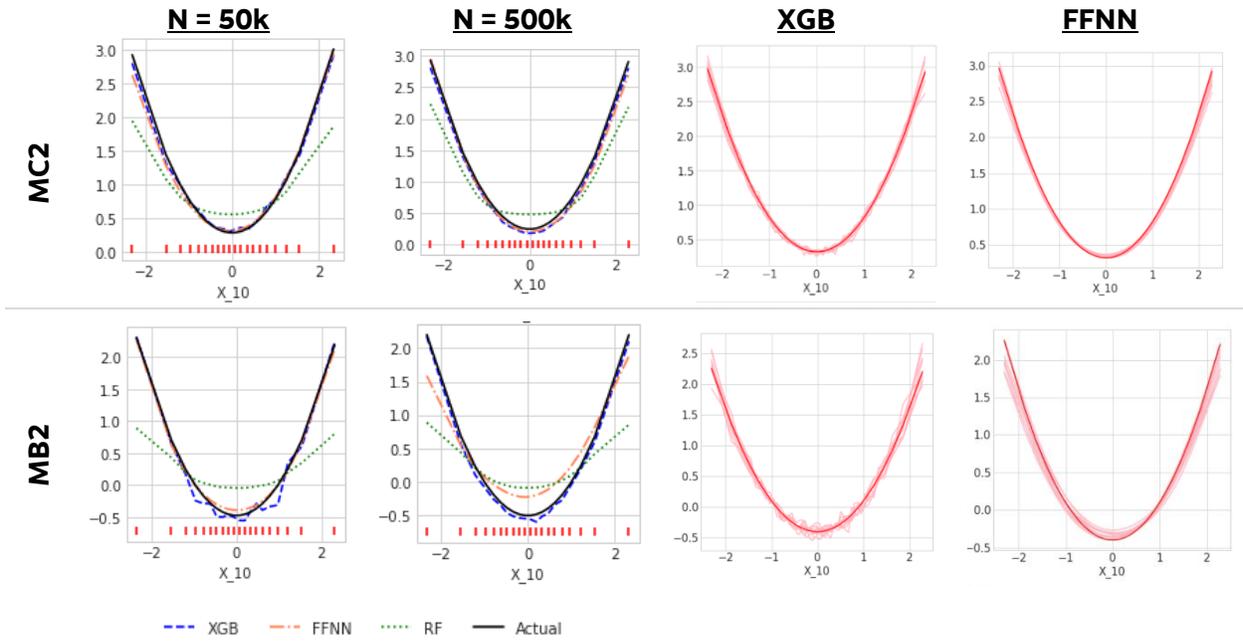

*Figure 27 shows the bias-variance decomposition of the PDPs for $X_1$ from MC1 (linear). FFNN had perceivably smaller bias and lower variance.*

Table 4 displays some summary statistics of the bias-variance decomposition of the predictions. The residuals showed clear improvement from XGB to FFNN, as the variability of the residuals were smaller in FFNN in comparison to XGB.

*Figure 27: PDP for uncorrelated MC1 (linear), focusing on variable $X_1$; (left) N = 50k; (second from the left) 500k, all three SML algorithms; (second from the right) N = 50k, XGB model; (right) N = 50k, FFNN model; **bold red line is the true PDP**; lighter red lines are PDPs from the ten replicates*

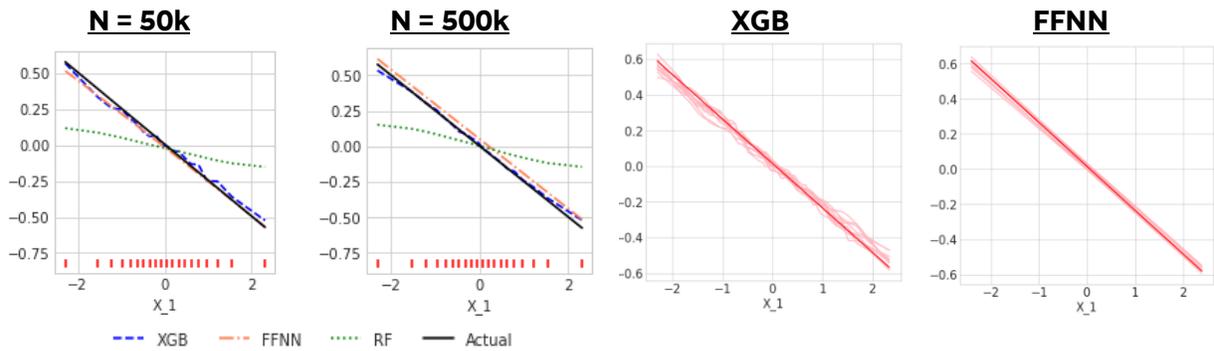

*Table 4: Summary statistics (minimum, mean, maximum, and standard deviation) for the bias-variance decomposition of the residuals for MC1 (linear); (top) XGB model, N = 500k; (bottom) FFNN model, N = 500k*

|     |             | Min.   | Mean   | Max.   | Std. Dev. |
| --- | ----------- | ------ | ------ | ------ | --------- |
| **XGB** | *Total Error* | 0.0012 | 0.0435 | 1.0905 | 0.0443 |
|     | *Bias²*     | 0.0000 | 0.0102 | 1.0503 | 0.0383 |

|  |  |  | 0.0012 | 0.0333 | 0.1590 | 0.0182 |
|  |  | Variance |  |  |  |  |

|  |  | | | | |
|---|---|---|---|---|---|
|  | Total Error | 0.0000 | 0.0011 | 0.1633 | 0.0029 |
| **FFNN** | Bias² | 0.0000 | 0.0007 | 0.1605 | 0.0028 |
|  | Variance | 0.0000 | 0.0004 | 0.0028 | 0.0002 |

FFNN continued to outperform XGB for smooth functional forms as correlation increased. Table 5 shows the bias-variance decomposition of the predicted output for MC2 (linear with global interactions). XGB generated larger bias values and residuals in comparison to FFNN. The majority of the total error originated from the squared bias. These large values originated from the extreme tails, where tree-based methods flatten out due to the piecewise constant nature of the model.

Table 5: Summary statistics (minimum, mean, maximum, and standard deviation) for the bias-variance decomposition of the residuals for MC2 (linear with global interactions); (top) XGB model, two correlation levels, N = 500k; (bottom) FFNN model, two correlation levels, N = 500k

|  |  |  | Min. | Mean | Max. | Std. Dev. |
|---|---|---|---|---|---|---|
| **XGB** | r = 0.25 | Total Error | 0.0037 | 0.1110 | 10.6655 | 0.3041 |
|  |  | Bias² | 0.0000 | 0.0510 | 10.4277 | 0.2868 |
|  |  | Variance | 0.0030 | 0.0600 | 0.6524 | 0.0510 |
|  | r = 0.5 | Total Error | 0.0050 | 0.1293 | 12.5799 | 0.2885 |
|  |  | Bias² | 0.0000 | 0.0357 | 12.0393 | 0.2492 |
|  |  | Variance | 0.0033 | 0.0937 | 1.4183 | 0.0962 |
| **FFNN** | r = 0.25 | Total Error | 0.0006 | 0.0181 | 1.6655 | 0.0519 |
|  |  | Bias² | 0.0000 | 0.0079 | 1.5865 | 0.0449 |
|  |  | Variance | 0.0003 | 0.0101 | 0.2683 | 0.0120 |
|  | r = 0.5 | Total Error | 0.0004 | 0.0356 | 3.6703 | 0.1316 |
|  |  | Bias² | 0.0000 | 0.0301 | 3.6414 | 0.1296 |
|  |  | Variance | 0.0002 | 0.0055 | 0.0762 | 0.0054 |

### 5.5.2 Finding 8: AIM Interactions

**FFNN captured interactions present in additive index models, while XGB generally failed in this area.**

Figures 28 and 29 show the CICE plots for the binary and continuous response cases for MB8/MC8 (complex model), respectively. FFNN captured the wavy[9] patterns in the CICE lines of $X_3$ and $X_4$ that results from the AIM term $\log(|\beta_1 x_1 + \cdots + \beta_5 x_5|)$. XGB failed to capture the wavy pattern. FFNN captured AIM-style interactions regardless of response type and correlation level.

---

[9] We define "wavy" as the peaks and valleys present towards the center of the plots, below the average CICE line.

*Figure 28: Actual and estimated CICE lines for uncorrelated MB8 (complex model), focusing on $X_3$ and $X_4$, sample size N = 500k; (first column) XGB model CICE plots; (second column) FFNN model CICE plots; (third column) actual CICE plots*

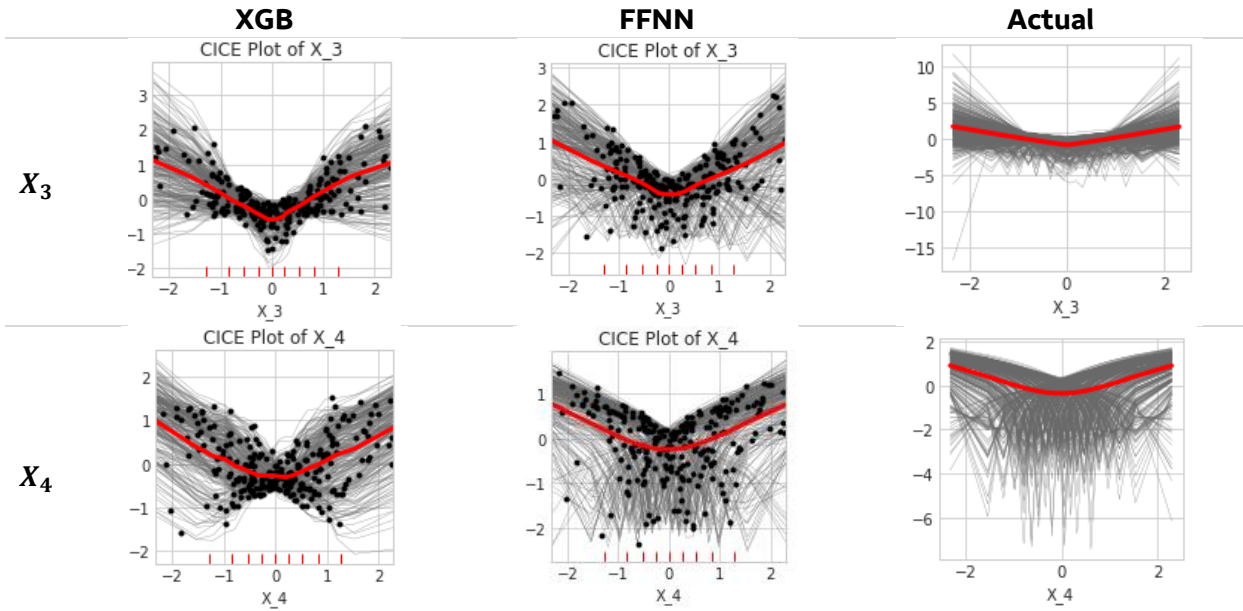

FFNN captured the wavy pattern as correlation increased. However, the magnitude and prevalence of the waves decreased as correlation increased, most notably in the tails of the plots.

Figure 29 shows a lower density of peaks and valleys towards the more extreme values of $X_4$, indicating as correlation increased, FFNN lost some of its ability to identify AIM-style interaction behavior, perhaps due to variable identifiability issues that arise with higher levels of correlation.

*Figure 29: Actual and estimated CICE plots for $X_4$ from MB8 (complex model) for three correlation levels, N = 50k*

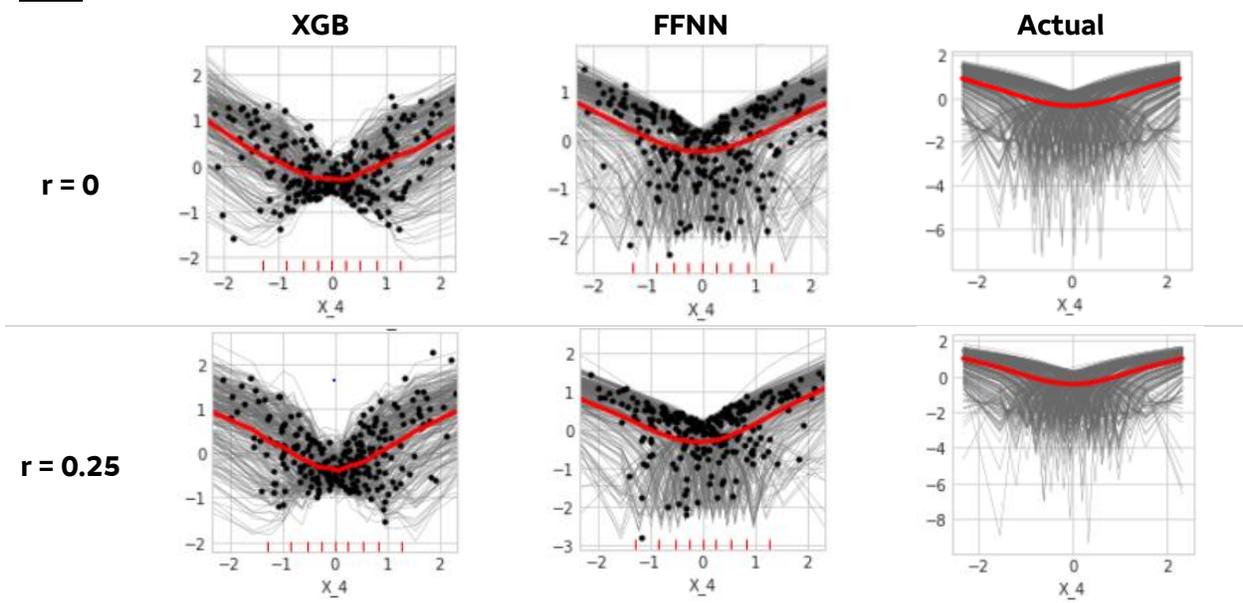

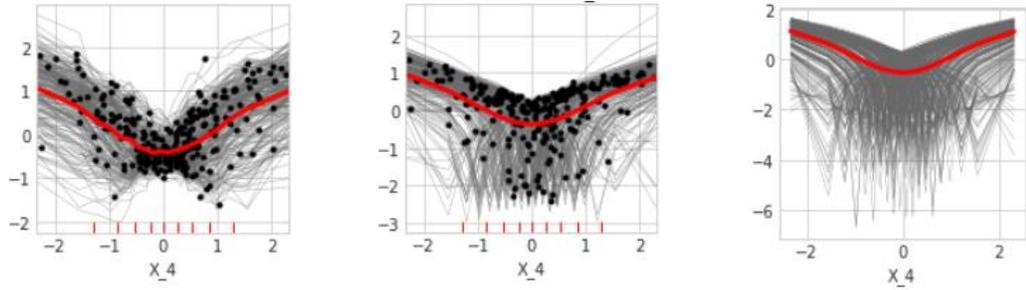

### 5.5.3 Finding 9: Noise Variables

**Noise variables were generally captured correctly for both algorithms. XGB did slightly better in this regard.**

Figure 30 shows the same noise variable $X_{35}$ for XGB and FFNN from MC8L/MB8L (complex model), where we see minimal bias for both algorithms. XGB more properly represented the PDPs for noise terms than FFNN does, as FFNN captured artificial nonlinear (quadratic) information. The overall bias for the noise terms in all cases tended to increase as correlation increased.

*Figure 30: Noise variable $X_{35}$ in MC8L/MB8L for XGB and FFNN models, N = 50k; bold red line is the true PDP; lighter red lines are PDPs from the ten replicates*

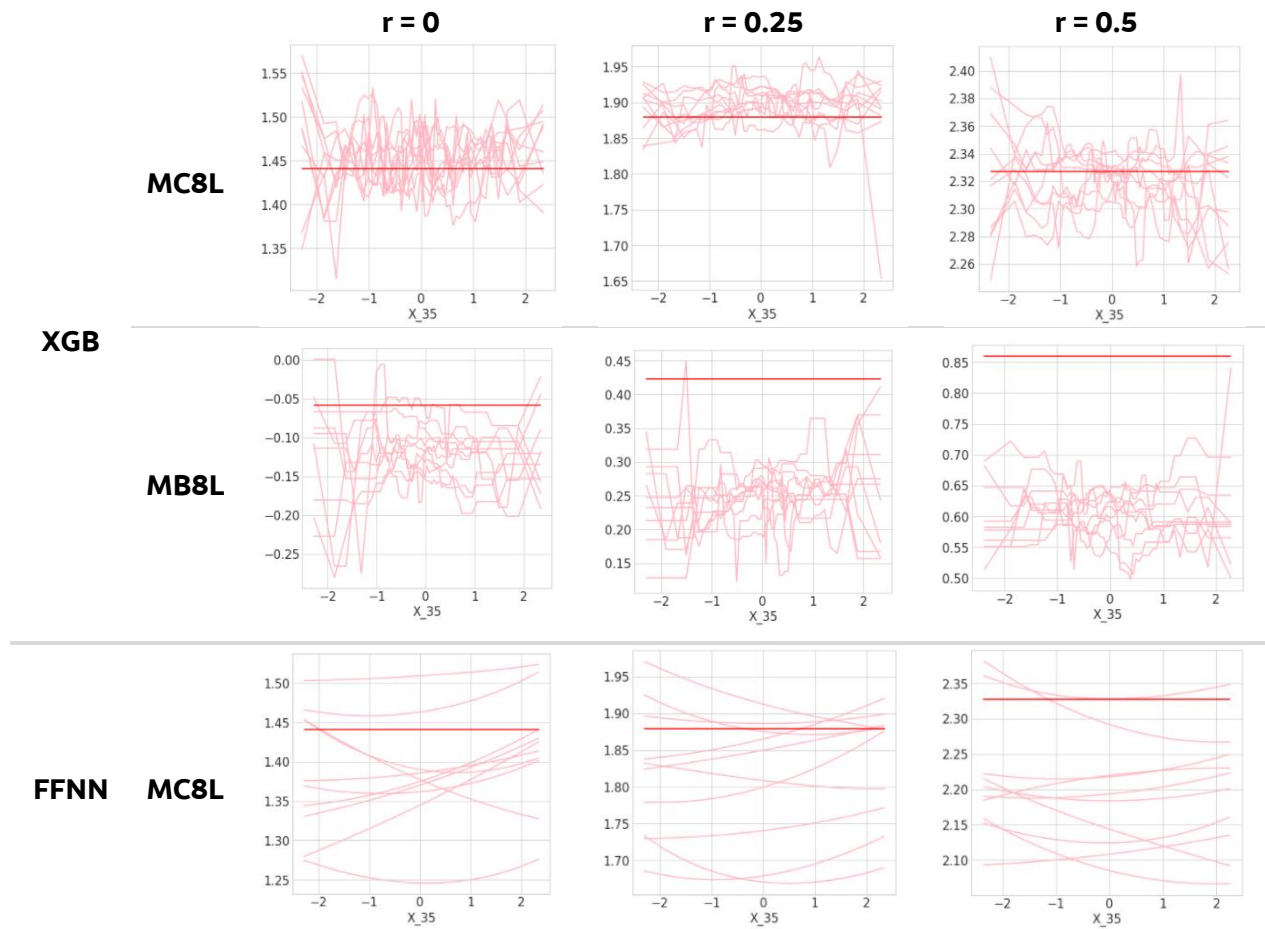

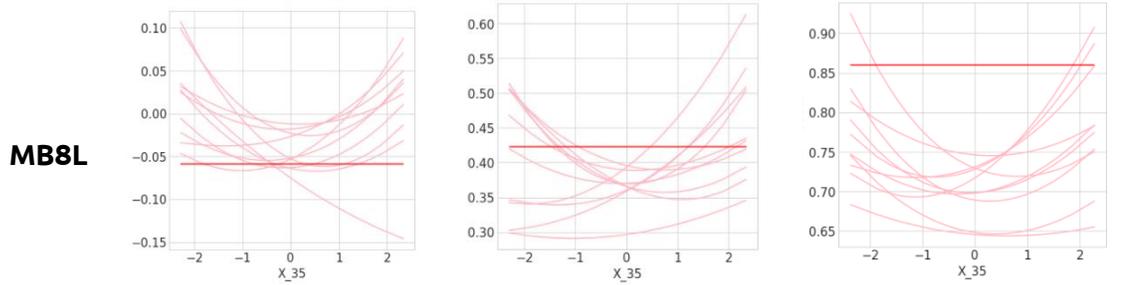

**MB8L**

### 5.6 Finding 10: Sample Size

**As expected, the performances of all algorithms improved as sample size increased from 50k to 500k. This included predictive performance, reduction in variability across simulations, and reduction in bias and variance in the one-dimensional PDPs of individual predictors. In addition, bias and variance of PDPs for the noise terms decreased so that these terms can be properly identified as unimportant.**

Figure 31 shows the PDPs for MC6 (jumpy GAM with local interactions), where stronger adherence to the true PDP curve was observed for $N = 500$k in comparison to $N = 50$k. The overall variability and bias among the ten replicates decreased.

Table 6 shows the average bias-variance decomposition of the fitted responses for MC6, which showed an overall decrease in error as sample size increased.

*Figure 31: PDP for uncorrelated MC6 (jumpy GAM with local interactions), focusing on variable $X_1$, fitted using all three SML algorithms and two sample sizes N = 50k and N = 500k; (first column) XGB model; (second column) FFNN model; **bold red line** is the true PDP; lighter red lines are PDPs from the ten replicates*

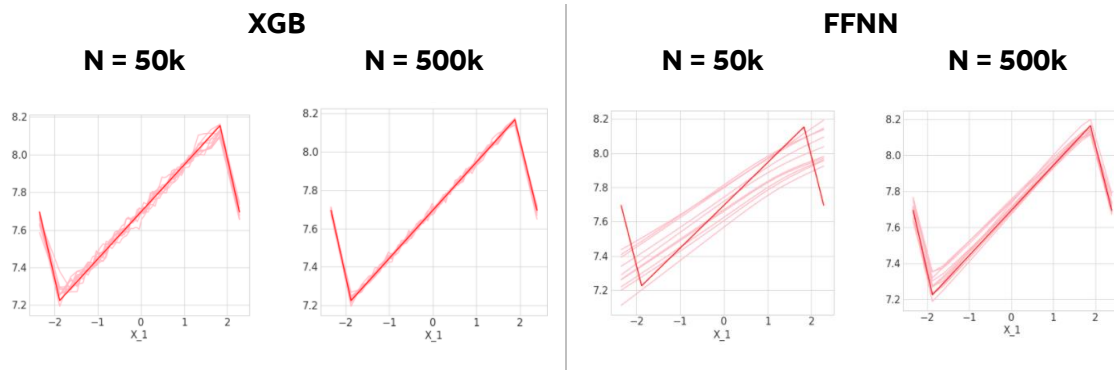

*Table 6: Average bias-variance decomposition for the fitted response for uncorrelated functional form MC6*

| Model | Sample Size | Error | Squared Bias | Variance |
|---|---|---|---|---|
| XGB | N = 50k | 0.097 | 0.050 | 0.047 |
|  | N = 500k | 0.072 | 0.041 | 0.031 |
| FFNN | N = 50k | 0.214 | 0.172 | 0.041 |
|  | N = 500k | 0.050 | 0.033 | 0.017 |

As correlation increased, increase in sample size showed a reduction in bias and variance for both XGB and FFNN. Figure 32 shows this for variable $X_5$ from MC2/MB2 (linear with global interactions), and

Figure 33 shows the bias-variance decomposition of these plots.

*Figure 32: PDPs for $X_5$ from MC2/MB2 (linear with global interactions) for three correlation levels, N = 50k; **bold red line** is the true PDP; lighter red lines are PDPs from the ten replicates*

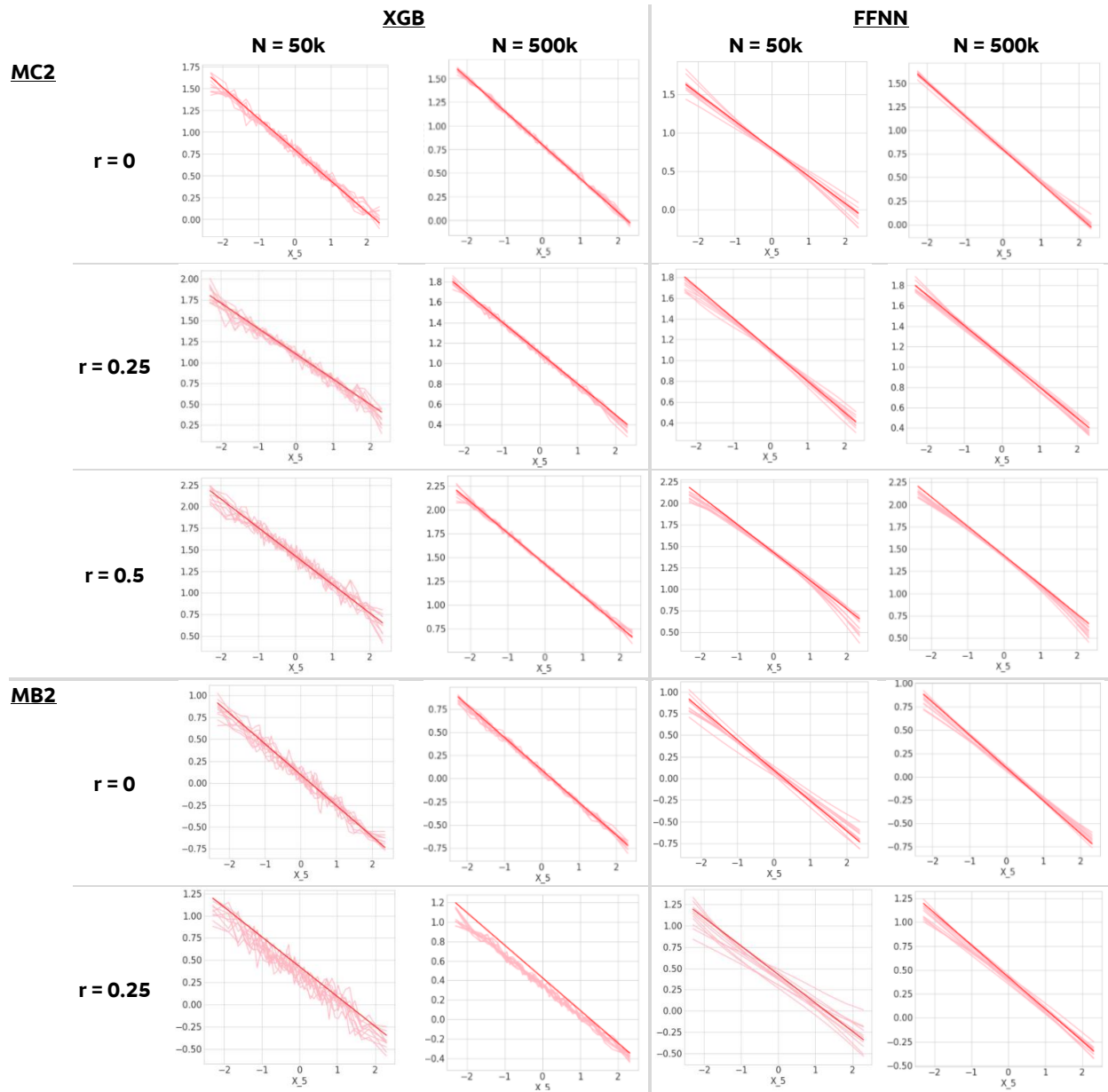

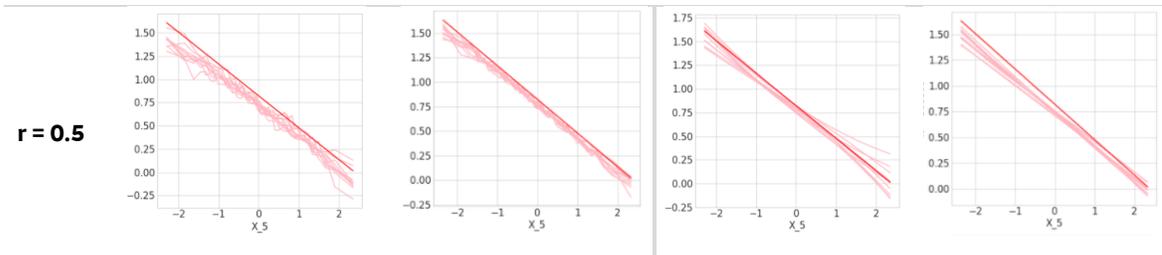

*Figure 33: Bias-variance decomposition of the PDP for MC2/MB2 (linear with global interactions), focusing on variable $X_5$, fitted using all three SML algorithms, sample size N = 50k and N = 500k, three correlation levels; (first column) XGB model; (second column) FFNN model*

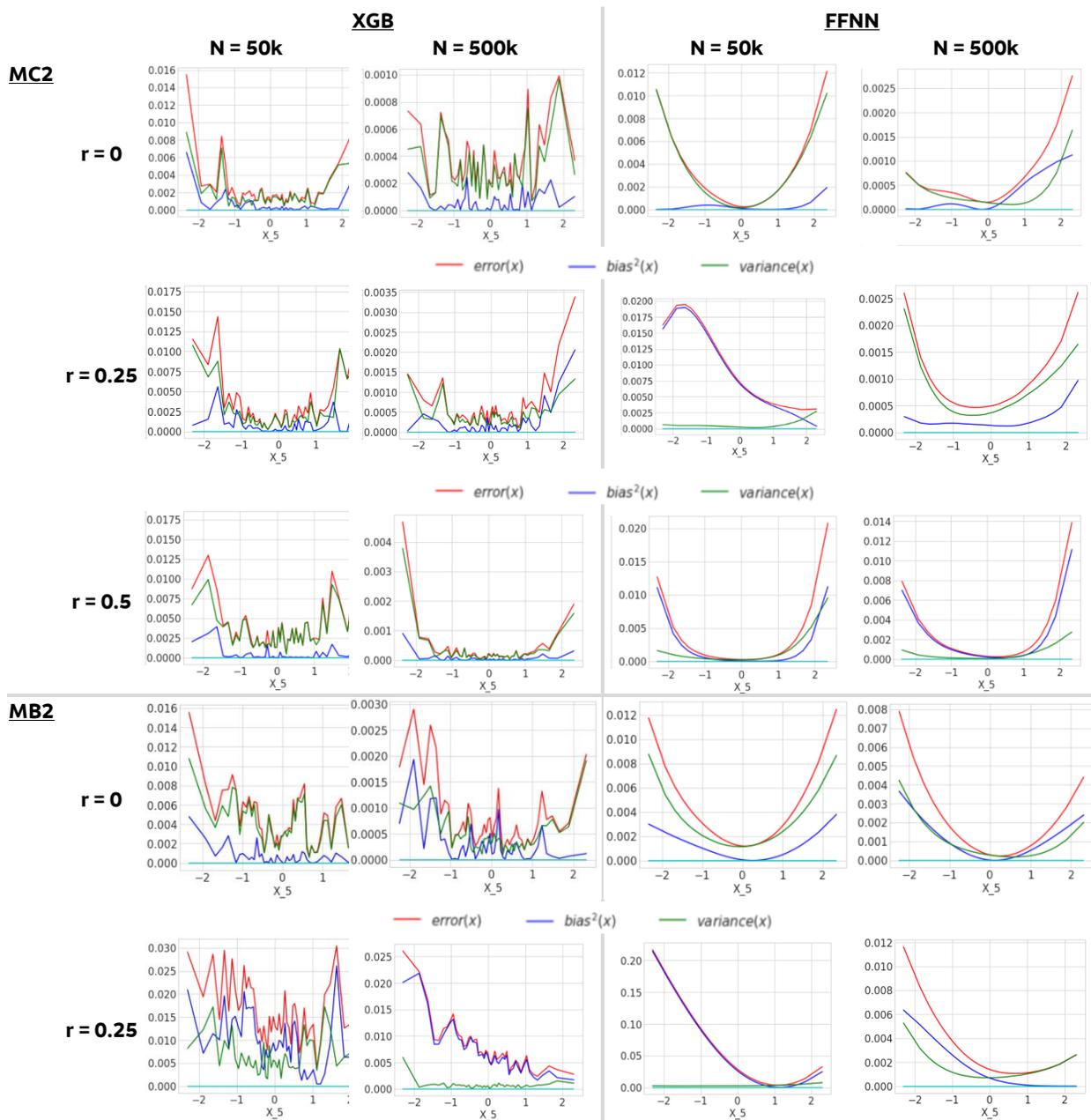

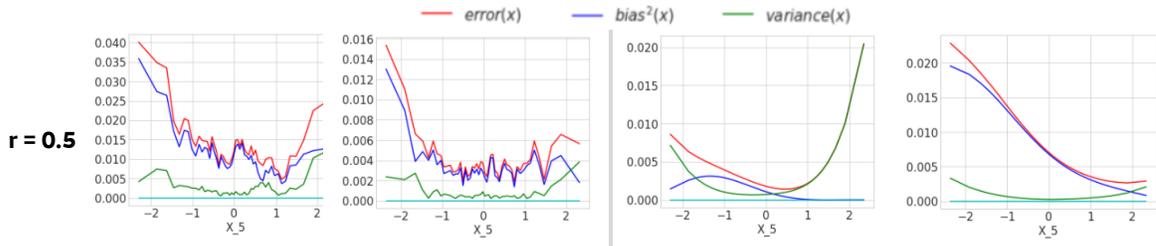

In addition, bias and variance of PDPs for the noise terms decreased so these terms could be properly identified as unimportant. Figure 34 shows noise term $X_{32}$ from MC2L/MB2L (linear with global interactions). The larger number of observations not only boosted performance in terms of bias and variance, but also enhanced the ability of the model to capture the true underlying behavior of the noise term.

*Figure 34: PDPs for noise term $X_{32}$ for MC2L/MB2L (linear with global interactions) for three correlation levels, N = 50k and N = 500k, for the (left column) XGB and (right column) FFNN algorithms; **bold red line** is the true PDP; lighter red lines are PDPs from the ten replicates*

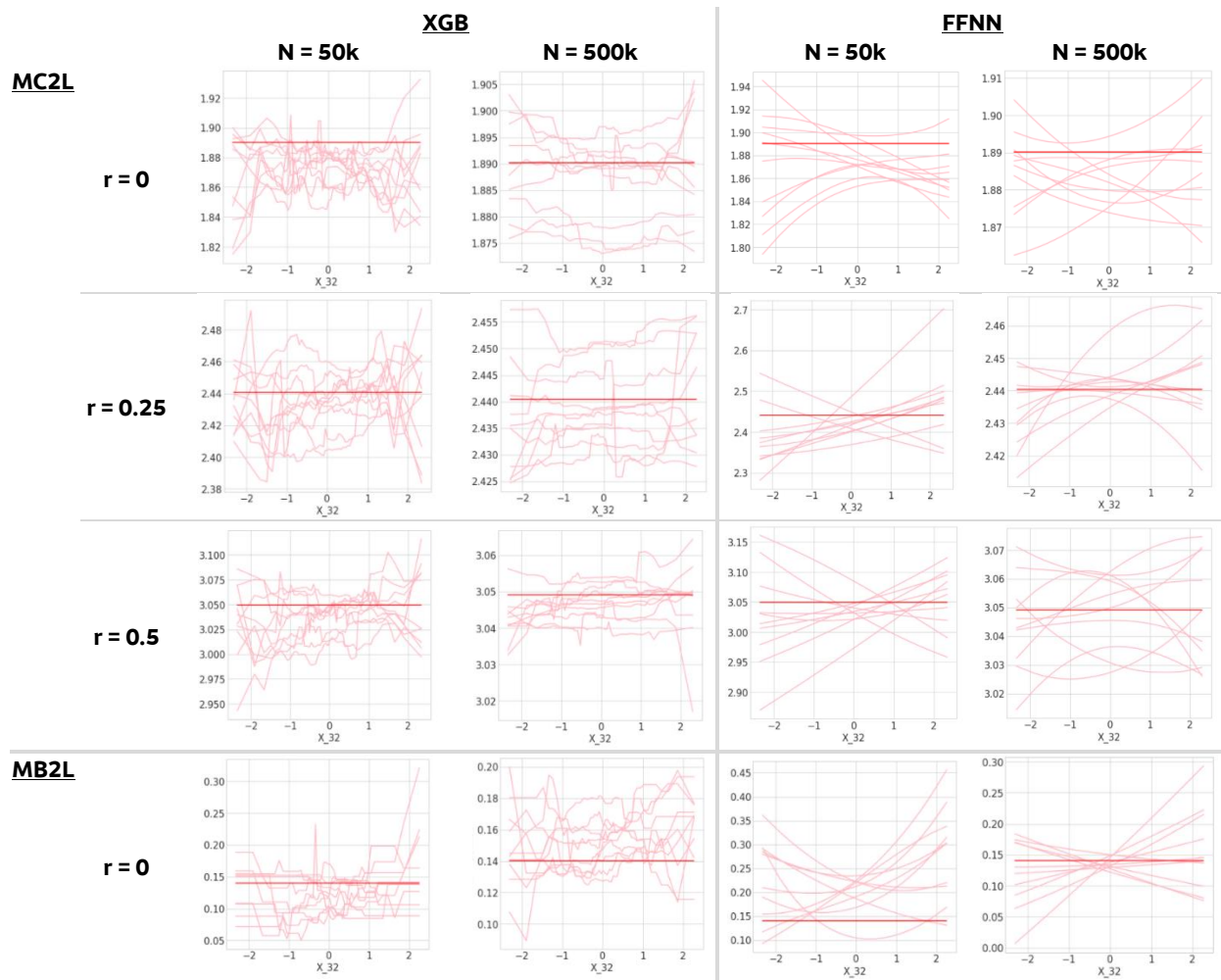

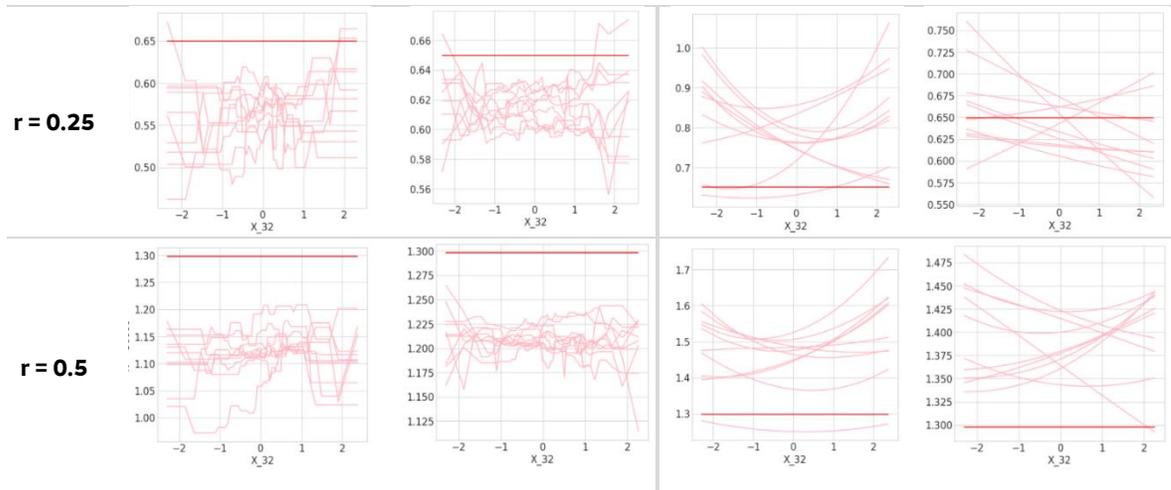

## 5.7 Finding 11: Bias and Variability Relative to Response Type

**As expected, for the same functional forms, the algorithms performed better with continuous responses compared to the binary responses. For binary response, the predictive performance was lower and there was higher bias and variability for the PDPs.**

Figure 35 shows PDPs and their respective bias-variance decomposition from MB1/MC1 (linear). The PDPs were not directly comparable between MC1 and MB1. However, when accounting for scale, MC1 showed smaller error and variance than MB1.

*Figure 35: PDP for MC1/MB1 (linear), focusing on variable $X_5$, fitted using all three SML algorithms, sample size N = 50k; (first column) XGB model; (second column) FFNN model; bold red line is the true PDP; lighter red lines are PDPs from the ten replicates*

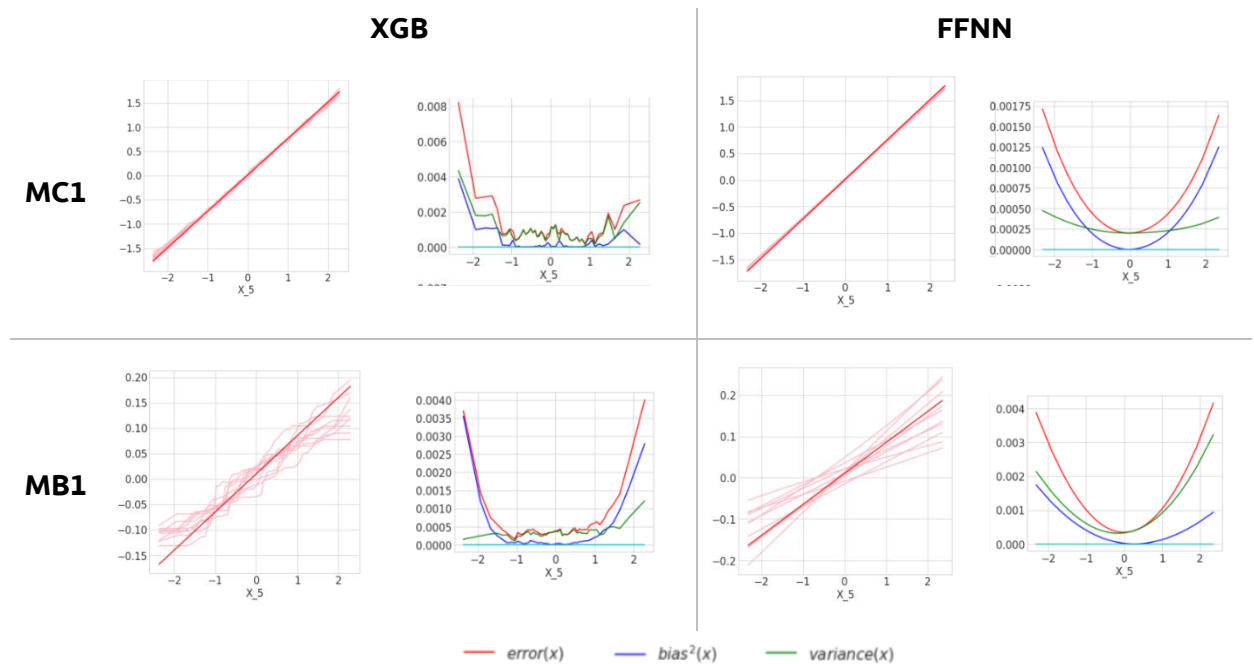

Figure 36 shows that as correlation increased for MC6L/MB6L (jumpy GAM with local interactions) for XGB, the $y$-axis range increased for noise variable $X_{34}$. While the increase is small, we expected no change given the relative range of values in the uncorrelated case. The binary response case showed larger deviations from the true PDP as correlation increased.

*Figure 36: PDPs for noise term $X_{34}$ for MC6L/MB6L (jumpy GAM with local interactions) for three correlation levels, N = 50k, for the (left column) XGB and (right column) FFNN models; bold red line is the true PDP; lighter red lines are PDPs from the ten replicates*

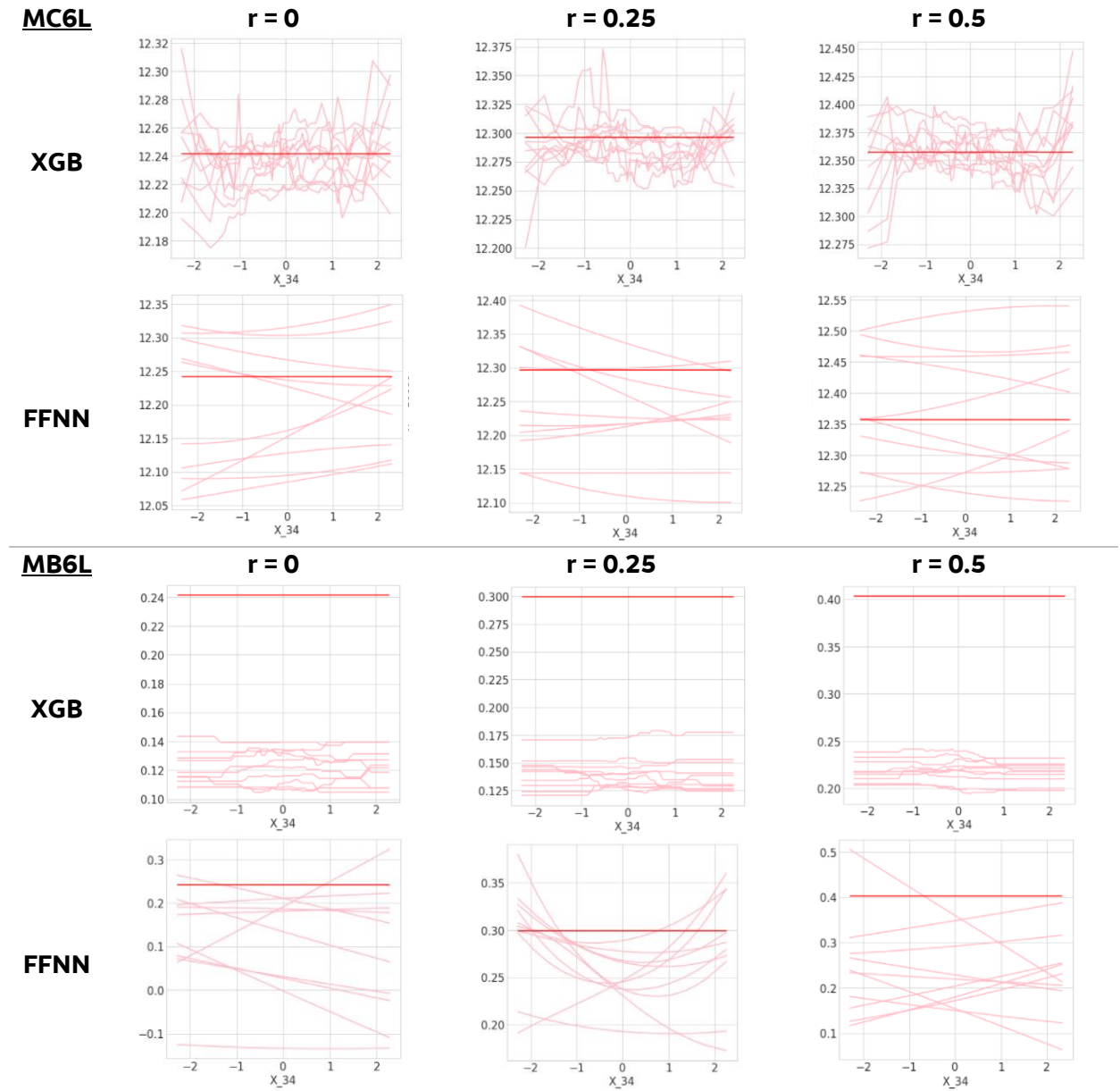

## 5.8 Finding 12: Correlation

**As expected, correlation inhibited the performance of XGB, FFNN, and RF due to increasing lack of identifiability among features, which resulted in larger bias and variability in model predictions and interpretability tools.**

We did not see any distinct patterns in the performance metrics as correlation increases. The general, overarching findings regarding interpretability tools described in this paper were mostly preserved regardless of the correlation value. What was not necessarily preserved was the quality of the finding, as that deteriorated as correlation increases. As observed in previous sections,[10] as correlation increased, the replicability of the true PDP curve decreased. In such cases, the effect of interactions could bleed into main effects, resulting in inaccurate interpretations from important variables. If the data set was affected by correlation among the predictors, the model could find lower order interactions captured the same information as higher order interactions, which limited the ability to capture fully the higher order interactions.

Figures 32 and 33 show the PDPs across a combination of sample sizes and correlation levels for $X_5$ from MC2/MB2 (linear with global interactions). Increasing sample size aided in model performance and PDP reproducibility. However, as correlation increased, regardless of sample size, the overall bias and variability of the estimated PDPs grew as well. As expected,[11] the bias and variability showed larger increases for binary response cases in comparison to continuous response cases. This behavior was also reflected in the CICE plots found in

Figure 29.

Table 5 shows the numerical increases of the total error, squared bias, and variance as correlation increased. The estimated PDPs increasingly deviated from the true PDP and had measurably greater variability. Figure 34 visualizes this behavior, where, as correlation grew, the larger the $y$-axis range grew in conjunction with the distance of the estimated PDPs from the true PDP. Bias and variance increased, as well as the presence of more pronounced structures (see Figures 18, 30, and 36).

If there was existing poor behavior from the model or in the PDPs, that behavior was exacerbated and magnified by increasing correlation. The over-regularization effect discussed in Section 5.4.2 worsened as correlation increased (see Figure 15). In the presence of correlated variables, RF may result in different subsets of variables that may have high correlation among the column subsamples, resulting in potentially correlated trees. Greater levels of correlation increased the presence of artificial interactions as well.

Tree-based methods are greedy in choosing a splitting variable. They will almost always choose the variables that contribute greatly over a variable that contributes more modestly. As correlation increases, less important variables will uniquely contribute to the model at a lower rate. These less important variables will appear deeper within the tree, after the more important variables have already taken their share. At this point, the less important variables may have little to nothing to explain, which can also result in a poor behavior from a fitted model.

---

[10] See Sections 5.4.1 and 5.5.1 for discussion on the estimation of the PDP respective to correlation level.
[11] See Sections 5.5.3 and 5.6 for discussion on the bias-variance decomposition relative to response type.

## 5.9 Finding 13: PDPs

**All algorithms exhibited some bias in the PDPs even in pristine environments such as uncorrelated predictors. The problems were worse in small samples for binary responses.**

The bias observed in estimated PDPs worsened as the complexity of the functional form increased. Many of the findings were reliant on the overall lack of good performance in the PDPs, which showed deficiencies or weaknesses in the model fit and interpretability. In simple cases, such as functional forms MC1/MB1[12] (linear) and MC2/MB2[13] (linear with global interactions), the true PDP was mostly recovered by XGB and FFNN. However, there were cases where the simple functional form was not captured by the PDP, even given a larger number of observations.

Functional forms that were more complicated tended to return estimated PDPs that had large biases. Examples[14] throughout this paper showed irreconcilable differences between the true and estimated PDP curves. The estimated PDP showed more degeneration as the true importance of the variable decreased.

# 6 Additional Observations

## 6.1 Observation 1: Usefulness of AUC

**AUC does not necessarily provide a true picture of model performance and can be misleading. There were instances where RF had competitive AUCs with XGB and FFNN. However, the PDPs from the FFNN and XGB were considerably closer to the truth when compared to those from RF, which exhibited significant deviation.**

The AUC values for MB6L (jumpy GAM with local interactions) suggested that RF was better than FFNN. However, the diagnostic plots tell a very different story. Figure 37 shows that the PDP curves for FFNN were more faithful to the true curve in comparison to RF.

*Figure 37: PDPs of $X_1, \ldots, X_5$ for uncorrelated functional form MB6L, N = 50k; **bold red line is the true PDP**; lighter red lines are PDPs from the ten replicates*

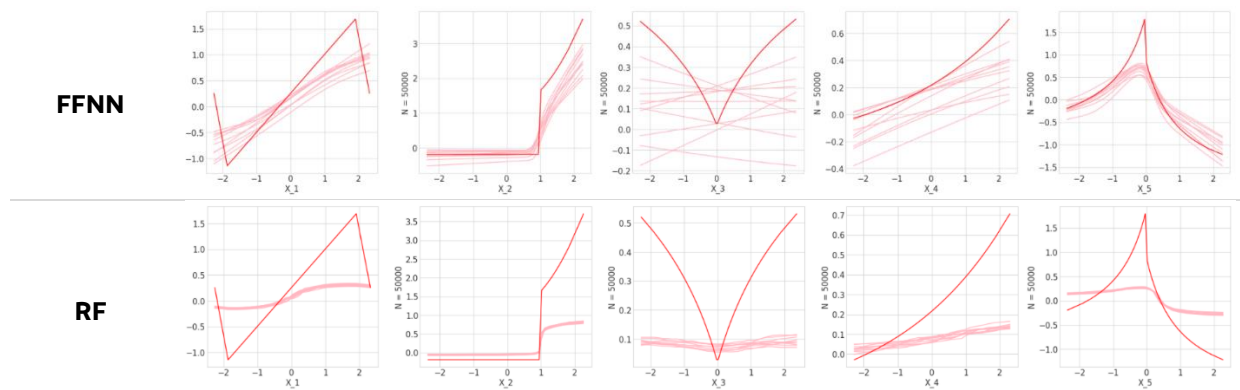

---

[12] See Figures 13, 27, and 35 for examples from functional forms MC1/MB1 (linear).
[13] See Figures 17 and 26 for examples from functional forms MC2/MB2 (linear with global interactions).
[14] See Figures 11, 15, and 16 for examples.

This behavior was also evident when we examined the distributions of residuals of the log-odds for MB6L. This example demonstrated one of the limitations of using AUC as a performance metric. It focused on accurate ranking of the probabilities, but appeared to fail in taking into account other aspects of model performance.

# 7   Concluding Remarks

We compared three popular SML algorithms based on predictive performance and behavior in *post-hoc* interpretability tools. As expected, these algorithms performed differently across functional forms. Based on the predictive performance in Section 5.1, we saw that XGB and FFNN supported better model prediction performance than RF across a variety of functional forms.

XGB and FFNN were competitive, with some differences for some of the functional forms. XGB performed better in cases of jumpy response surfaces; whereas FFNN was better suited for smooth response surfaces or if the underlying functional form is suspected of containing one or more AIM-style interactions. We do not recommend the use of RF, as it was not competitive in terms of model performance or interpretability. RF tended to exhibit over-fitting behaviors that were more severe in comparison to XGB and FFNN.

As we have seen, correlation negatively affected model performance and the efficacy of *post-hoc* interpretability methods. The results from the fitted models may be unreliable in the presence of correlated variables. This suggested that users should exercise more care in the model development stage and not blindly include many highly correlated predictors in the hopes for small improvements in predictive performance. While this may be standard practice in Kaggle competitions and automated ML development systems, it could create havoc for model interpretation.

Many of the applications in banking deal with binary responses. It is worth reiterating that binary data are considerably less informative and hence it is more challenging to identify and interpret the underlying models unless the sample sizes are large.

Finally, users should keep in mind that, despite the non-parametric nature of the SML algorithms, there can be bias in the results, significantly so in some cases. We saw this even with PDPs which summarize lower dimensional input-output relationships. The nature of the bias varies with the architecture of the SML algorithms with FFNN having smaller bias in smooth models. However, it cannot capture jumps or sharp turns, which tend to be smoothed out. Conversely, XGB is better at capturing non-smooth surfaces; but its jumpy nature can make interpretation of PDPs challenging.

# 8 Appendix

## 8.1 Coefficients

We performed some basic data checks while deciding on the coefficients. The coefficients for the smaller continuous and binary response cases are shown in Tables 7 and 8, respectively; the coefficients for the larger continuous and binary response cases are shown in Table 9.

For binary response cases, we ensured that the AUC is beyond a certain threshold (for example, 0.65). We also chose coefficients appropriately to ensure that the classes are not imbalanced (i.e., the proportion of observations belonging to a particular class should not exceed 0.65).

For continuous response cases, we ensured that $R^2$ is greater than 0.7. We also selected the signal-to-noise ratio (SNR) to be between 4 and 10, where SNR is defined as

$$SNR = \frac{\sigma[f(X)]}{\sigma(\varepsilon)}. \qquad \text{Eq. 6}$$

Here $f(x)$ is the functional form, $\sigma(\cdot)$ is the variance function, and $\varepsilon$ is the noise term. Further, we ensured that the variance of any single additive component does not dominate. To describe this, consider a model given by $Y = f_1(X_1) + f_2(X_2) + \cdots + f_p(X_p)$. Define the importance score for the $k$-th component as

$$Imp_k = \frac{Var[f_p(X_p)]}{Var[f_1(X_1) + f_2(X_2) + \cdots + f_p(X_p)]}. \qquad \text{Eq. 7}$$

The coefficients were chosen such that all $Imp_k \leq 0.4$ for all predictors.

*Table 7: Coefficients for smaller continuous response cases*

| Functional Form | Coefficients | | | | | | | | | |
|---|---|---|---|---|---|---|---|---|---|---|
| | $\beta_1$ | $\beta_2$ | $\beta_3$ | $\beta_4$ | $\beta_5$ | $\beta_6$ | $\beta_7$ | $\beta_8$ | $\beta_9$ | $\beta_{10}$ |
| **Linear (MC1)** | -0.25 | 0.8 | 1 | -0.25 | 0.75 | -0.2 | -0.5 | -0.3 | 0.75 | 1 |
| | $\beta_{11}$ | $\beta_{12}$ | $\beta_{13}$ | $\beta_{14}$ | $\beta_{15}$ | $\beta_{16}$ | $\beta_{17}$ | $\beta_{18}$ | $\beta_{19}$ | $\beta_{20}$ |
| | 0.25 | 0.3 | -0.25 | 0.3 | -0.35 | -0.2 | -0.2 | 0.75 | 0.25 | 0.5 |
| | $\beta_1$ | $\beta_2$ | $\beta_3$ | $\beta_4$ | $\beta_5$ | $\beta_6$ | $\beta_7$ | $\beta_8$ | $\beta_9$ | $\beta_{10}$ |
| **Linear with global interactions (MC2)** | 1 | 0.2 | -0.5 | 0.75 | -0.35 | -0.8 | -1 | 0.75 | 0.25 | 0.5 |
| | $\beta_{11}$ | $\beta_{12}$ | $\beta_{13}$ | $\beta_{14}$ | | | | | | |
| | 0.25 | 1 | 0.3 | 1.5 | | | | | | |
| | $\beta_1$ | $\beta_2$ | $\beta_3$ | $\beta_4$ | $\beta_5$ | $\beta_6$ | $\beta_7$ | $\beta_8$ | $\beta_9$ | $\beta_{10}$ |
| **Smooth GAM (MC3)** | 1 | 0.5 | 3 | -0.5 | 1 | 0.4 | 1 | 0.75 | -0.5 | 1 |
| **Smooth and jumpy GAM (MC4)** | 1.5 | 0.5 | 2.5 | 0.5 | 0.25 | 1.5 | 2 | 1.5 | 1.5 | 0.75 |

| Functional Form | | | | | | | | | | |
|---|---|---|---|---|---|---|---|---|---|---|
| GAM with global interactions (MC5) | 1 | 0.5 | -2 | 0.5 | 0.5 | 1.5 | 1 | 1.5 | 0.75 | 0.5 |
| Jumpy GAM with local interactions (MC6) | 0.25 | 0.5 | 3 | 0.75 | 5 | 2 | 1 | 3 | 2 | 1.5 |
| AIM (MC7) | $\beta_1$ | $\beta_2$ | $\beta_3$ | $\beta_4$ | $\beta_5$ | $\beta_6$ | $\beta_7$ | $\beta_8$ | $\beta_9$ | $\beta_{10}$ |
| | 0.5 | 1.5 | 3 | 5 | 4 | 4 | 2 | 2.5 | 1.5 | 2 |
| | $\beta_{11}$ | $\beta_{12}$ | $\beta_{13}$ | $\beta_{14}$ | $\beta_{15}$ | $\beta_{16}$ | $\beta_{17}$ | | | |
| | 1 | 0.25 | 0.75 | 0.25 | 0.5 | 0.35 | 2 | | | |
| Complex (MC8) | $\beta_1$ | $\beta_2$ | $\beta_3$ | $\beta_4$ | $\beta_5$ | $\beta_6$ | $\beta_7$ | $\beta_8$ | $\beta_9$ | $\beta_{10}$ |
| | 1 | 0.5 | 3 | 2 | 1.5 | 1 | 1.5 | 1.5 | 2 | 0.5 |

*Table 8: Coefficients for smaller binary response cases*

| Functional Form | Coefficients | | | | | | | | | |
|---|---|---|---|---|---|---|---|---|---|---|
| Linear (MB1) | $\beta_1$ | $\beta_2$ | $\beta_3$ | $\beta_4$ | $\beta_5$ | $\beta_6$ | $\beta_7$ | $\beta_8$ | $\beta_9$ | $\beta_{10}$ |
| | 0.1 | 0.04 | 0.035 | 0.2 | 0.075 | -0.025 | -0.2 | -0.03 | 0.2 | -0.35 |
| | $\beta_{11}$ | $\beta_{12}$ | $\beta_{13}$ | $\beta_{14}$ | $\beta_{15}$ | $\beta_{16}$ | $\beta_{17}$ | $\beta_{18}$ | $\beta_{19}$ | $\beta_{20}$ |
| | -0.05 | -0.35 | 0.35 | -0.03 | -0.25 | 0.095 | 0.075 | -0.15 | 0.25 | 0.15 |
| Linear with global interactions (MB2) | $\beta_1$ | $\beta_2$ | $\beta_3$ | $\beta_4$ | $\beta_5$ | $\beta_6$ | $\beta_7$ | $\beta_8$ | $\beta_9$ | $\beta_{10}$ |
| | 0.5 | 0.2 | -0.05 | 0.15 | -0.35 | -0.8 | -0.8 | 0.75 | 0.05 | 0.5 |
| | $\beta_{11}$ | $\beta_{12}$ | $\beta_{13}$ | $\beta_{14}$ | | | | | | |
| | 0.25 | 0.1 | 0.3 | 0.5 | | | | | | |
| Smooth GAM (MB3) | $\beta_1$ | $\beta_2$ | $\beta_3$ | $\beta_4$ | $\beta_5$ | $\beta_6$ | $\beta_7$ | $\beta_8$ | $\beta_9$ | $\beta_{10}$ |
| | -0.75 | 0.05 | 0.08 | 0.75 | 1.5 | -0.5 | -0.5 | 0.075 | -0.4 | -0.1 |
| Smooth and jumpy GAM (MB4) | -1 | -0.35 | 0.2 | -0.25 | 0.2 | -2 | 0.75 | -2.5 | 0.5 | 0.75 |
| GAM with global interactions (MB5) | -1 | -0.5 | -2 | -0.5 | 0.5 | -0.5 | -1 | -0.75 | -0.75 | -0.5 |
| Jumpy GAM with local interactions (MB6) | 0.25 | -0.5 | -3 | -0.5 | -0.5 | -0.75 | 1 | -3 | -2 | -2.5 |
| AIM (MB7) | $\beta_1$ | $\beta_2$ | $\beta_3$ | $\beta_4$ | $\beta_5$ | $\beta_6$ | $\beta_7$ | $\beta_8$ | $\beta_9$ | $\beta_{10}$ |
| | 0.5 | 1.5 | 0.3 | 2 | 0.75 | 0.4 | 0.2 | 2.5 | 1.5 | 2 |
| | $\beta_{11}$ | $\beta_{12}$ | $\beta_{13}$ | $\beta_{14}$ | $\beta_{15}$ | $\beta_{16}$ | $\beta_{17}$ | | | |
| | 1 | 0.25 | 0.75 | 0.25 | 0.5 | -0.1 | -0.75 | | | |
| **Complex (MB8)** | $\beta_1$ | $\beta_2$ | $\beta_3$ | $\beta_4$ | $\beta_5$ | $\beta_6$ | $\beta_7$ | $\beta_8$ | $\beta_9$ | $\beta_{10}$ |
| | 1 | -0.5 | 1 | -2 | -1.5 | 1 | 1.5 | -1.5 | -2.5 | -0.55 |

The larger functional forms include some additional considerations to generate three groups of additive components that contribute approximately 60%, 30%, and 10% to the variation in the response. In all cases, each group of additive components is multiplied by a constant to control for the variation. The constants are $c_1 = 1, c_2 = 0.7,$ and $c_3 = 0.4$.

Additionally, the binary response cases include intercepts to ensure class balance. In binary response cases, all intercept values $\beta_0$ is non-zero, and is specified in Table 9. In continuous response cases, all intercept values $\beta_0$ is equal to zero.

Recall that each of the larger functional forms include either 10 or 20 noise variables that do not contribute to the response. Functional forms MB5L/MC5L consist of 10 noise variables, whereas the remaining three sets of functional forms MB2L/MC2L, MB6L/MC5L, and MB8L/MC8L consist of 20 noise variables. Refer to Section 3 for details of the functional forms.

*Table 9: Coefficients for larger binary and continuous response cases; note that $\beta_0$ values are only valid for the binary response cases (all $\beta_0$ values for continuous response cases are zero)*

| Functional Form | Coefficients | | | | | | | | | | |
|---|---|---|---|---|---|---|---|---|---|---|---|
| | $\beta_0$ | $\beta_1$ | $\beta_2$ | $\beta_3$ | $\beta_4$ | $\beta_5$ | $\beta_6$ | $\beta_7$ | $\beta_8$ | $\beta_9$ | $\beta_{10}$ |
| **Linear with global interactions (MB2L/MC2L)** | -1.75 | 0.6 | 0.6 | -0.51 | 0.57 | -0.57 | -0.57 | -0.6 | 0.57 | 0.45 | 0.45 |
| | $\beta_{11}$ | $\beta_{12}$ | $\beta_{13}$ | $\beta_{14}$ | | | | | | | |
| | 0.57 | 0.6 | 0.57 | 0.6 | | | | | | | |
| | $\beta_0$ | $\beta_1$ | $\beta_2$ | $\beta_3$ | $\beta_4$ | $\beta_5$ | $\beta_6$ | $\beta_7$ | $\beta_8$ | $\beta_9$ | $\beta_{10}$ |
| **GAM with global interactions (MB5L/MC5L)** | -13 | 0.5 | -2 | 0.5 | 0.5 | 3.5 | 0.65 | 0.75 | 1.5 | 0.75 | 0.35 |
| **Jumpy GAM with local interactions (MB6L/MC6L)** | -12 | 0.75 | 0.5 | 2.5 | 0.3 | 3 | 1.25 | 1.5 | 3 | 1 | 1 |
| **Complex (MB8L/MC8L)** | -1.5 | 1 | 0.5 | 0.75 | 0.75 | 0.5 | -0.45 | 1 | 2 | 1.25 | 0.25 |

## 8.2 Hyperparameter Configuration Spaces

In this section, we provide the hyperparameter configuration spaces used for HPO for the SML algorithms.

### 8.2.1  XGB

The details for the hyperparameter configuration space for XGB is given in Table 10. Specifically, we created a grid of values for the maximum depth, learning rate, column sampling by tree, subsample rate per tree, minimum child weight, number of estimators or trees used in the model, and the L1 and L2 regularization terms (alpha and lambda, respectively). These hyperparameter values were used for both the smaller and larger cases.

*Table 10: Hyperparameter configuration space for the XGB algorithm*

| Hyperparameter | Range of Values |
|---|---|
| **Maximum depth** | 4, 5, …, 9 |
| **Learning rate** | 0.3, 0.2, 0.1, 0.05, 0.01, 0.005 |
| **Column sampling by tree** | 0.2, 0.4, 0.6, 0.8 |
| **Subsample rate** | 0.2, 0.4, 0.6, 0.8 |
| **Minimum child weight** | 10, 20, 50, 80, 100 |
| **Number of estimators** | 50, 100, 200, 400, 600 |
| **Lambda (L2)** | 0, 0.001, 0.01, 0.01, 1, 10, 100 |
| **Alpha (L1)** | 0, 0.001, 0.01, 0.01, 1, 10, 100 |

### 8.2.2  FFNN

For FFNN, we used different configurations based on the response type and some trial-and-error. The details for the hyperparameter configuration space for FFNN is given in Table 11. For binary response cases, both deep and shallow FFNN were tuned. We define "shallow" in this case as FFNN models with one or two fully connected layers. "Deep" is defined as three or four fully connected layers. The layers of the shallow FFNN model are allowed to be wider than the layers specified for the deep FFNN model. The FFNN model between these two options with the best predictive score was chosen.

For continuous response cases, only shallow FFNN models were tuned. In some continuous response cases, the predictive accuracy for the FFNN model worsened when sample size was increased from $N = 50k$ to $N = 500k$. In these cases, we retuned the models with sample size $N = 500k$ using a bigger configuration space. In particular, we used wider layers for the FFNN model. The functional forms for which the updated configuration space was used are the following: MC2, MC4, MC6, and MC7. These differences in layer sizes are specified in Table 11.

The original values for the shallow FFNN model layers is given in the first row of the "Layers" section, with the additional values for wider layers given in the second row. The values for the deep FFNN model is given in the third row of this section. The fourth row of this section provides additional layer configurations for the larger functional form, which include up to 50 predictors.

Additional consideration was given to the number of epochs available for the larger functional forms, allowing HPO to choose from among a range of 100 to 400 epochs rather than fixing the value at 200.

*Table 11: Hyperparameter configuration space for the FFNN algorithm*

| Hyperparameter | | Range of Values |
|---|---|---|
| **Activation function** | | ReLU |
| **Batch normalization** | | True |
| **Dropout rate** | | 0.1, 0.2, 0.3, 0.4, 0.5 |
| **Learning rate** | | $10^{-3}, 10^{-2}, 10^{-1}$ |
| **Layers** | *Original values for shallow FFNN* | [16], [32], [64], [128], [32, 16], [64, 32], [128, 64] |
| | *Additional values for shallow FFNN* | [512], [256, 128], [512, 256] |
| | *Values for deep FFNN* | [32, 16, 8], [16, 8, 4], [32, 16, 8, 4] |
| | *Additional values for larger cases* | [256], [512], [1024], [1024, 512], [64, 32, 16], [128, 64, 32], [256, 128, 64], [512, 256, 128], [1025, 512, 256], [64, 32, 16, 8] |
| **Epochs** | *Smaller cases* | 200 |
| | *Larger cases* | 100, 150, 200, 250, 300, 350, 400 |
| **Batch size** | | 256, 512, 1024, 4096 |
| **L1 regularization** | | $10^x$, where x = [-5, -4, -3, -2, -1] |
| **L2 regularization** | | $10^x$, where x = [-5, -4, -3, -2, -1] |
| **Early stopping** | | False |

### 8.2.3 RF

The details for the hyperparameter configuration space for RF is given in Table 12. Specifically, we created a grid of values for the maximum depth, number of trees or estimators, maximum number of samples for each bootstrap sample, minimum number of samples needed for splitting a node, and maximum features used for each tree. The maximum number of features used for each tree depends on the number of features for a functional form, and is the determining factor for the maximum number of features. In Table 12, we specify five cases for 5, 10, 20, 25, and 30 features in a functional form. These hyperparameter values were used for both the smaller and larger cases.

*Table 12: Hyperparameter configuration space for the RF algorithm*

| Hyperparameter | Range of Values |
|---|---|
| **Maximum depth** | 10, 11, ..., 25 |
| **Number of estimators** | 100, 120, 140, ..., 1000 |
| **Maximum sample size (sampling rate)** | 0.4, 0.5, ..., 0.9 |
| **Minimum number of samples when splitting** | 1, 2, ..., 100 |

| | | |
|---|---|---|
| **Maximum number of features** | If number of features = 5: | 3 |
| | If number of features = 10: | 5 |
| | If number of features = 20: | 7 |
| | If number of features = 25: | 10 |
| | If number of features = 30: | 20 |

## 8.3 Predictive Metrics

This section reports the predictive scores for all functional forms, organized by response type and sample size combination. Tables 13 and 14 provide the AUCs for the smaller binary response cases for sample sizes $N = 50k$ and $N = 500k$, respectively. Tables 15 and 16 provide the MSEs for the smaller continuous response cases for sample sizes $N = 50k$ and $N = 500k$, respectively.

Tables 17 and 18 provide the AUCs for the larger binary response cases for sample sizes $N = 50k$ and $N = 500k$, respectively. Tables 19 and 20 provide the MSEs for the larger continuous response cases for sample sizes $N = 50k$ and $N = 500k$, respectively.

In all tables, all correlation levels that were run for each functional form are reported. For each functional form in these tables, we provide the mean and standard deviation of the AUC or MSE across ten fitted models based on a single testing data set. The oracle AUC or MSE for the testing set is also provided in all cases, as a reference to the testing metrics.

For additional details and analysis of these predictive scores, see Section 5.1.

### 8.3.1 Tables: Models with Smaller Sets of Predictors

The testing performance metrics presented in Tables 13 to 16 correspond to the plots provided in Sections 5.1.1 and 5.1.2 to support Finding 1. These tables include metrics for models with smaller sets of predictors.

Table 13: Testing AUC of all models with a smaller set of predictors for binary response cases with sample size N = 50k (the best value is bolded; the worst value is italicized)

| Functional Form | Correlation | Oracle AUC | XGB | | FFNN | | RF | |
|---|---|---|---|---|---|---|---|---|
| | | | Mean | SD | Mean | SD | Mean | SD |
| **Linear (MB1)** | r = 0 | 0.707 | 0.698 | 0.004 | **0.703** | 0.004 | *0.693* | 0.005 |
| **Linear with global interactions (MB2)** | r = 0 | 0.908 | 0.9009 | 0.0011 | **0.9050** | 0.0008 | *0.8796* | 0.0014 |
| | r = 0.25 | 0.916 | 0.8975 | 0.0018 | **0.9121** | 0.0011 | *0.8836* | 0.0010 |
| | r = 0.5 | 0.919 | 0.9096 | 0.0010 | **0.9126** | 0.0004 | *0.8919* | 0.0013 |
| **Smooth GAM (MB3)** | r = 0 | 0.737 | **0.733** | 0.005 | **0.733** | 0.005 | *0.726* | 0.005 |
| **Smooth + Jumpy GAM (MB4)** | r = 0 | 0.774 | **0.771** | 0.003 | 0.766 | 0.004 | *0.764* | 0.004 |
| | r = 0 | 0.876 | 0.8703 | 0.0012 | *0.8661* | 0.0005 | **0.8753** | 0.0005 |

| Functional Form | | Correlation | Oracle AUC | XGB | | FFNN | | RF | |
|---|---|---|---|---|---|---|---|---|---|
| | | | | Mean | SD | Mean | SD | Mean | SD |
| GAM with global interactions (MB5) | | r = 0.25 | 0.880 | 0.8707 | 0.0009 | **0.8832** | 0.0005 | *0.8623* | 0.0016 |
| | | r = 0.5 | 0.892 | 0.8775 | 0.0009 | *0.8731* | 0.0004 | **0.8880** | 0.0006 |
| Jumpy GAM with local interactions (MB6) | | r = 0 | 0.865 | **0.8599** | 0.0014 | *0.8574* | 0.0010 | 0.8581 | 0.0007 |
| | | r = 0.25 | 0.893 | 0.8776 | 0.0010 | **0.8826** | 0.0014 | *0.8747* | 0.0013 |
| | | r = 0.5 | 0.882 | **0.8848** | 0.0015 | 0.8832 | 0.0012 | *0.8844* | 0.0006 |
| AIM (MB7) | | r = 0 | 0.776 | 0.705 | 0.006 | **0.769** | 0.004 | *0.681* | 0.006 |
| Complex (MB8) | | r = 0 | 0.884 | 0.8583 | 0.0015 | **0.8663** | 0.0025 | *0.8369* | 0.0023 |
| | | r = 0.25 | 0.883 | 0.8407 | 0.0036 | **0.8625** | 0.0022 | *0.8266* | 0.0023 |
| | | r = 0.5 | 0.888 | 0.8609 | 0.0013 | **0.8713** | 0.0012 | *0.8441* | 0.0016 |

*Table 14: Testing AUC of all models with a smaller set of predictors for binary response cases with sample size N = 500k (the best value is bolded; the worst value is italicized)*

| Functional Form | Correlation | Oracle AUC | XGB | | FFNN | | RF | |
|---|---|---|---|---|---|---|---|---|
| | | | Mean | SD | Mean | SD | Mean | SD |
| **Linear (MB1)** | r = 0 | 0.708 | 0.708 | 0.002 | **0.709** | 0.002 | *0.701* | 0.002 |
| **Linear with global interactions (MB2)** | r = 0 | 0.908 | 0.9051 | 0.0007 | **0.9069** | 0.0005 | *0.8904* | 0.0007 |
| | r = 0.25 | 0.916 | 0.9084 | 0.0009 | **0.9148** | 0.0003 | *0.8941* | 0.0007 |
| | r = 0.5 | 0.919 | 0.9119 | 0.0007 | **0.9136** | 0.0003 | *0.9009* | 0.0004 |
| **Smooth GAM (MB3)** | r = 0 | 0.740 | **0.738** | 0.002 | 0.737 | 0.002 | *0.732* | 0.002 |
| **Smooth and jumpy GAM (MB4)** | r = 0 | 0.779 | **0.777** | 0.002 | 0.773 | 0.002 | *0.772* | 0.002 |
| **GAM with global interactions (MB5)** | r = 0 | 0.876 | **0.8758** | 0.0007 | 0.8731 | 0.0008 | *0.8713* | 0.0014 |
| | r = 0.25 | 0.880 | *0.8760* | 0.0003 | **0.8847** | 0.0005 | 0.8781 | 0.0009 |
| | r = 0.5 | 0.892 | 0.8812 | 0.0004 | **0.8854** | 0.0006 | *0.8599* | 0.0014 |
| **Jumpy GAM with local interactions (MB6)** | r = 0 | 0.865 | **0.8634** | 0.0008 | 0.8618 | 0.0004 | *0.8615* | 0.0005 |
| | r = 0.25 | 0.893 | 0.8826 | 0.0006 | **0.8900** | 0.0006 | *0.8805* | 0.0007 |
| | r = 0.5 | 0.882 | **0.8904** | 0.0005 | *0.8866* | 0.0007 | 0.8888 | 0.0006 |
| **AIM (MB7)** | r = 0 | 0.776 | 0.732 | 0.001 | **0.774** | 0.002 | *0.713* | 0.002 |
| **Complex (MB8)** | r = 0 | 0.884 | 0.8680 | 0.0012 | **0.8743** | 0.0027 | *0.8534* | 0.0014 |
| | r = 0.25 | 0.883 | 0.8583 | 0.0016 | **0.8767** | 0.0030 | *0.8446* | 0.0011 |

| | r = 0.5 | 0.888 | 0.8663 | 0.0009 | **0.8807** | 0.0008 | *0.8548* | 0.0007 |

Table 15: *Testing MSE of all models with a smaller set of predictors for continuous response cases with sample size N = 50k (the top value is bolded; the worst value is italicized)*

| Functional Form | Correlation | Oracle MSE | XGB | | FFNN | | RF | |
|---|---|---|---|---|---|---|---|---|
| | | | Mean | SD | Mean | SD | Mean | SD |
| **Linear (MC1)** | r = 0 | 0.995 | 1.078 | 0.017 | **1.003** | 0.019 | *1.924* | 0.032 |
| **Linear with global interactions (MC2)** | r = 0 | 0.995 | 1.1448 | 0.0177 | **1.0388** | 0.0048 | *3.0036* | 0.0399 |
| | r = 0.25 | 0.995 | 1.2534 | 0.0202 | **1.0426** | 0.0069 | *2.8784* | 0.0379 |
| | r = 0.5 | 0.995 | 1.5465 | 0.0286 | **1.0200** | 0.0040 | *2.2357* | 0.0236 |
| **Smooth GAM (MC3)** | r = 0 | 0.995 | **1.075** | 0.019 | 1.079 | 0.018 | *1.662* | 0.033 |
| **Smooth and jumpy GAM (MC4)** | r = 0 | 0.995 | **1.029** | 0.02 | 1.125 | 0.023 | *1.363* | 0.029 |
| **GAM with global interactions (MC5)** | r = 0 | 0.995 | 1.3648 | 0.0303 | **1.0776** | 0.0121 | *1.5951* | 0.0311 |
| | r = 0.25 | 0.995 | 1.3492 | 0.0261 | **1.0811** | 0.0180 | *1.5049* | 0.0275 |
| | r = 0.5 | 0.995 | 1.3447 | 0.0160 | **1.1020** | 0.0232 | *1.4495* | 0.0179 |
| **Jumpy GAM with local interactions (MC6)** | r = 0 | 0.995 | **1.0625** | 0.0089 | 1.1871 | 0.0238 | *1.3373* | 0.0426 |
| | r = 0.25 | 0.995 | **1.1238** | 0.0155 | 1.1618 | 0.0186 | *1.4182* | 0.0171 |
| | r = 0.5 | 0.995 | **1.0884** | 0.0078 | 1.0980 | 0.0327 | *1.2950* | 0.0076 |
| **AIM (MC7)** | r = 0 | 0.995 | 2.81 | 0.45 | **1.311** | 0.147 | *3.978* | 0.565 |
| **Complex (MC8)** | r = 0 | 0.995 | 1.7707 | 0.0192 | **1.6920** | 0.0854 | *3.0343* | 0.0365 |
| | r = 0.25 | 0.995 | 2.0177 | 0.0305 | **1.3708** | 0.0290 | *2.8617* | 0.0513 |
| | r = 0.5 | 0.995 | 1.7937 | 0.0244 | **1.2108** | 0.0147 | *2.6663* | 0.0404 |

Table 16: *Testing MSE of all models with a smaller set of predictors for continuous response cases with sample size N = 500k (the top value is bolded; the worst value is italicized)*

| Functional Form | Correlation | Oracle MSE | XGB | | FFNN | | RF | |
|---|---|---|---|---|---|---|---|---|
| | | | Mean | SD | Mean | SD | Mean | SD |
| **Linear (MC1)** | r = 0 | 0.998 | 1.022 | 0.005 | **1.002** | 0.005 | *1.705* | 0.008 |
| **Linear with global interactions (MC2)** | r = 0 | 0.998 | 1.0548 | 0.0051 | **1.0212** | 0.0020 | *2.4606* | 0.0190 |
| | r = 0.25 | 0.998 | 1.0950 | 0.0227 | **1.0032** | 0.0044 | *1.9923* | 0.0142 |
|  | r = 0.5 | 0.998 | 1.1043 | 0.0078 | **1.0217** | 0.0024 | *1.7561* | 0.0118 |
| **Smooth GAM (MC3)** | r = 0 | 0.998 | **1.034** | 0.004 | 1.077 | 0.010 | *1.392* | 0.008 |
| **Smooth and jumpy GAM (MC4)** | r = 0 | 0.998 | **1.013** | 0.004 | 1.057 | 0.005 | *1.338* | 0.008 |
| **GAM with global interactions (MC5)** | r = 0 | 0.998 | 1.1608 | 0.0167 | **1.0992** | 0.0197 | *1.2379* | 0.0112 |
|  | r = 0.25 | 0.998 | 1.1505 | 0.0150 | **1.0982** | 0.0279 | *1.2115* | 0.0163 |
|  | r = 0.5 | 0.998 | 1.1589 | 0.0137 | **1.0626** | 0.0182 | *1.1669* | 0.0110 |
| **Jumpy GAM with local interactions (MC6)** | r = 0 | 0.998 | 1.0582 | 0.0085 | **1.0375** | 0.0065 | *1.2136* | 0.0466 |
|  | r = 0.25 | 0.998 | **1.0536** | 0.0056 | 1.0900 | 0.0108 | *1.2042* | 0.0095 |
|  | r = 0.5 | 0.998 | **1.0420** | 0.0051 | 1.0463 | 0.0064 | *1.1762* | 0.0074 |
| **AIM (MC7)** | r = 0 | 0.998 | 2.586 | 1.473 | **1.394** | 0.426 | *3.769* | 1.699 |
| **Complex (MC8)** | r = 0 | 0.998 | 1.6032 | 0.0372 | **1.3004** | 0.0318 | *2.3165* | 0.1005 |
|  | r = 0.25 | 0.998 | 1.6684 | 0.0294 | **1.3755** | 0.0281 | *2.1869* | 0.0210 |
|  | r = 0.5 | 0.998 | 1.4267 | 0.0235 | **1.1905** | 0.0224 | *2.1214* | 0.0147 |

### 8.3.2 Tables: Models with Larger Sets of Predictors

The training and testing performance metrics presented in Tables 17 to 24 correspond to the plots provided in Section 5.1. These tables include metrics for models with larger sets of predictors.

#### 8.3.2.1 Testing Performance Metrics

*Table 17: Testing AUC of all models with a larger set of predictors for binary response cases with sample size N = 50k (the largest value is bolded; the smallest value is italicized)*

| **Functional Form** | **Correlation** | **Oracle AUC** | **XGB** | | **FFNN** | | **RF** | |
|---|---|---|---|---|---|---|---|---|
|  |  |  | **Mean** | **SD** | **Mean** | **SD** | **Mean** | **SD** |
| **Linear with global interactions (MB2L)** | r = 0 | 0.903 | 0.8637 | 0.0030 | **0.8816** | 0.0022 | *0.8398* | 0.0017 |
|  | r = 0.25 | 0.909 | 0.8747 | 0.0031 | **0.8923** | 0.0022 | *0.8487* | 0.0013 |
|  | r = 0.5 | 0.916 | 0.8801 | 0.0027 | **0.8954** | 0.0015 | *0.8123* | 0.0019 |
| **GAM with global interactions (MB5L)** | r = 0 | 0.895 | 0.8649 | 0.0020 | **0.8740** | 0.0017 | *0.8437* | 0.0014 |
|  | r = 0.25 | 0.902 | 0.8737 | 0.0022 | **0.8775** | 0.0017 | *0.8501* | 0.0011 |
|  | r = 0.5 | 0.910 | 0.8877 | 0.0017 | **0.8947** | 0.0014 | *0.8685* | 0.0016 |
|  | r = 0 | 0.891 | **0.8668** | 0.0011 | *0.8393* | 0.0023 | 0.8461 | 0.0018 |

| Functional Form | Correlation | | | | | | |
|---|---|---|---|---|---|---|---|
| Jumpy GAM with local interactions (MB6L) | r = 0.25 | 0.884 | **0.8601** | 0.0009 | *0.8365* | 0.0032 | 0.8378 | 0.0008 |
| | r = 0.5 | 0.891 | **0.8736** | 0.0016 | *0.8509* | 0.0018 | 0.8513 | 0.0013 |
| Complex (MB8L) | r = 0 | 0.859 | 0.7853 | 0.0050 | **0.7928** | 0.0028 | *0.7497* | 0.0027 |
| | r = 0.25 | 0.864 | **0.7988** | 0.0033 | 0.7955 | 0.0044 | *0.7779* | 0.0021 |
| | r = 0.5 | 0.855 | 0.7914 | 0.0026 | **0.7938** | 0.0021 | *0.7757* | 0.0021 |

Table 18: Testing AUC of all models with a larger set of predictors for binary response cases with sample size N = 500k (the largest value is bolded; the smallest value is italicized)

| Functional Form | Correlation | Oracle AUC | XGB | | FFNN | | RF | |
|---|---|---|---|---|---|---|---|---|
| | | | Mean | SD | Mean | SD | Mean | SD |
| Linear with global interactions (MB2L) | r = 0 | 0.903 | 0.8879 | 0.0014 | **0.8990** | 0.0008 | *0.8500* | 0.0008 |
| | r = 0.25 | 0.909 | 0.8976 | 0.0022 | **0.9059** | 0.0005 | *0.8604* | 0.0011 |
| | r = 0.5 | 0.916 | 0.9009 | 0.0016 | **0.9074** | 0.0007 | *0.8617* | 0.0009 |
| GAM with global interactions (MB5L) | r = 0 | 0.895 | 0.8837 | 0.0011 | **0.8866** | 0.0011 | *0.8550* | 0.0007 |
| | r = 0.25 | 0.902 | **0.8896** | 0.0013 | 0.8890 | 0.0007 | *0.8626* | 0.0006 |
| | r = 0.5 | 0.910 | 0.9014 | 0.0012 | **0.9021** | 0.0010 | *0.8800* | 0.0006 |
| Jumpy GAM with local interactions (MB6L) | r = 0 | 0.891 | **0.8846** | 0.0007 | 0.8679 | 0.0016 | *0.8556* | 0.0005 |
| | r = 0.25 | 0.884 | 0.8774 | 0.0009 | **0.8785** | 0.0007 | *0.8450* | 0.0007 |
| | r = 0.5 | 0.891 | **0.8879** | 0.0010 | 0.8794 | 0.0011 | *0.8588* | 0.0006 |
| Complex (MB8L) | r = 0 | 0.859 | 0.8272 | 0.0023 | **0.8295** | 0.0019 | *0.7689* | 0.0015 |
| | r = 0.25 | 0.864 | **0.8334** | 0.0015 | 0.8327 | 0.0023 | *0.7943* | 0.0009 |
| | r = 0.5 | 0.855 | 0.8275 | 0.0022 | **0.8298** | 0.0023 | *0.7881* | 0.0009 |

Table 19: Testing MSE of all models with a larger set of predictors for continuous response cases with sample size N = 50k (the smallest value is bolded; the largest value is italicized)

| Functional Form | Correlation | Oracle MSE | XGB | | FFNN | | RF | |
|---|---|---|---|---|---|---|---|---|
| | | | Mean | SD | Mean | SD | Mean | SD |
| Linear with global interactions (MC2L) | r = 0 | 0.984 | 1.6251 | 0.0421 | **1.1601** | 0.0101 | *5.4733* | 0.0237 |
| | r = 0.25 | 0.984 | 1.5839 | 0.0281 | **1.1485** | 0.0105 | *5.3797* | 0.0283 |
| | r = 0.5 | 0.984 | 1.5642 | 0.0221 | **1.1677** | 0.0189 | *5.0412* | 0.0205 |

| Functional Form | Correlation | Oracle MSE | XGB | | FFNN | | RF | |
|---|---|---|---|---|---|---|---|---|
| | | | Mean | SD | Mean | SD | Mean | SD |
| GAM with global interactions (MC5L) | r = 0 | 0.984 | 1.4258 | 0.0206 | **1.4117** | 0.0144 | *3.4925* | 0.0216 |
| | r = 0.25 | 0.984 | 1.4195 | 0.0099 | **1.3876** | 0.0638 | *3.7229* | 0.0250 |
| | r = 0.5 | 0.984 | 1.4092 | 0.0223 | **1.3488** | 0.0113 | *3.7893* | 0.0409 |
| Jumpy GAM with local interactions (MC6L) | r = 0 | 0.984 | **1.2425** | 0.0124 | 1.5279 | 0.0339 | *3.5525* | 0.0239 |
| | r = 0.25 | 0.984 | **1.2134** | 0.0194 | 1.4818 | 0.0398 | *3.2671* | 0.0183 |
| | r = 0.5 | 0.984 | 1.2235 | 0.0137 | **1.2126** | 0.0133 | *3.1371* | 0.0220 |
| Complex (MC8L) | r = 0 | 0.984 | 2.5267 | 0.0417 | **1.7577** | 0.0453 | *4.7827* | 0.0207 |
| | r = 0.25 | 0.984 | 2.1258 | 0.0255 | **1.7158** | 0.0494 | *4.5791* | 0.0152 |
| | r = 0.5 | 0.984 | 2.0251 | 0.0282 | **1.7570** | 0.0876 | *4.3792* | 0.0357 |

*Table 20: Testing MSE of all models with a larger set of predictors for continuous response cases with sample size N = 500k (the smallest value is bolded; the largest value is italicized)*

| Functional Form | Correlation | Oracle MSE | XGB | | FFNN | | RF | |
|---|---|---|---|---|---|---|---|---|
| | | | Mean | SD | Mean | SD | Mean | SD |
| Linear with global interactions (MC2L) | r = 0 | 0.984 | 1.2960 | 0.0113 | **1.0246** | 0.0054 | *4.8163* | 0.0156 |
| | r = 0.25 | 0.984 | 1.2989 | 0.0167 | **1.0264** | 0.0067 | *4.7084* | 0.0131 |
| | r = 0.5 | 0.984 | 1.3226 | 0.0104 | **1.0255** | 0.0042 | *4.3272* | 0.0112 |
| GAM with global interactions (MC5L) | r = 0 | 0.984 | 1.2339 | 0.0075 | **1.1428** | 0.0095 | *2.8529* | 0.0133 |
| | r = 0.25 | 0.984 | 1.2179 | 0.0148 | **1.1553** | 0.0687 | *2.9718* | 0.0123 |
| | r = 0.5 | 0.984 | 1.2033 | 0.0099 | **1.1433** | 0.0477 | *2.9395* | 0.0194 |
| Jumpy GAM with local interactions (MC6L) | r = 0 | 0.984 | **1.0981** | 0.0111 | 1.2440 | 0.0204 | *2.9380* | 0.0094 |
| | r = 0.25 | 0.984 | **1.0961** | 0.0079 | 1.2028 | 0.0127 | *2.7837* | 0.0054 |
| | r = 0.5 | 0.984 | **1.0926** | 0.0059 | 1.2025 | 0.0138 | *2.6648* | 0.0087 |
| Complex (MC8L) | r = 0 | 0.984 | 1.9044 | 0.0303 | **1.3272** | 0.0137 | *4.4562* | 0.0138 |
| | r = 0.25 | 0.984 | 1.6419 | 0.0245 | **1.3128** | 0.0134 | *4.2009* | 0.0121 |
| | r = 0.5 | 0.984 | 1.6152 | 0.0185 | **1.3222** | 0.0148 | *3.9751* | 0.0125 |

### 8.3.2.2   *Training Performance Metrics*

*Table 21: Training AUC of all models with a larger set of predictors for binary response cases with sample size N = 50k (the best value is bolded; the worst value is italicized)*

| Functional Form | Correlation | XGB | FFNN | RF |
|---|---|---|---|---|

|  |  | Oracle AUC | Mean | SD | Mean | SD | Mean | SD |
|---|---|---|---|---|---|---|---|---|
| **Linear with global interactions (MB2L)** | **r = 0** | 0.903 | *0.9238* | 0.0014 | 0.9246 | 0.0010 | **0.9505** | 0.0009 |
|  | **r = 0.25** | 0.909 | 0.9276 | 0.0013 | *0.9261* | 0.0009 | **0.9443** | 0.0009 |
|  | **r = 0.5** | 0.916 | 0.9310 | 0.0013 | *0.9256* | 0.0005 | **0.9383** | 0.0008 |
| **GAM with global interactions (MB5L)** | **r = 0** | 0.895 | 0.9123 | 0.0007 | *0.8845* | 0.0007 | **0.9418** | 0.0006 |
|  | **r = 0.25** | 0.902 | 0.9211 | 0.0010 | *0.8950* | 0.0011 | **0.9449** | 0.0009 |
|  | **r = 0.5** | 0.910 | 0.9283 | 0.0012 | *0.9051* | 0.0014 | **0.9486** | 0.0007 |
| **Jumpy GAM with local interactions (MB6L)** | **r = 0** | 0.891 | 0.8826 | 0.0015 | *0.8441* | 0.0018 | **0.9527** | 0.0010 |
|  | **r = 0.25** | 0.884 | 0.8869 | 0.0012 | *0.8701* | 0.0022 | **0.9502** | 0.0006 |
|  | **r = 0.5** | 0.891 | 0.8906 | 0.0020 | *0.8566* | 0.0024 | **0.9476** | 0.0012 |
| **Complex (MB8L)** | **r = 0** | 0.859 | 0.8722 | 0.0013 | *0.8433* | 0.0026 | **0.9045** | 0.0030 |
|  | **r = 0.25** | 0.864 | 0.8751 | 0.0021 | *0.8429* | 0.0023 | **0.9114** | 0.0010 |
|  | **r = 0.5** | 0.855 | 0.8798 | 0.0013 | *0.8381* | 0.0032 | **0.9104** | 0.0014 |

Table 22: Training AUC of all models with a larger set of predictors for binary response cases with sample size N = 500k (the largest value is bolded; the smallest value is italicized)

| Functional Form | Correlation | Oracle AUC | XGB | | FFNN | | RF | |
|---|---|---|---|---|---|---|---|---|
|  |  |  | Mean | SD | Mean | SD | Mean | SD |
| **Linear with global interactions (MB2L)** | **r = 0** | 0.903 | 0.9114 | 0.0006 | *0.9063* | 0.0005 | **0.9608** | 0.0002 |
|  | **r = 0.25** | 0.909 | 0.9150 | 0.0004 | *0.9097* | 0.0003 | **0.9557** | 0.0003 |
|  | **r = 0.5** | 0.916 | 0.9202 | 0.0003 | *0.9180* | 0.0003 | **0.9503** | 0.0003 |
| **GAM with global interactions (MB5L)** | **r = 0** | 0.895 | 0.8986 | 0.0005 | *0.8938* | 0.0005 | **0.9286** | 0.0005 |
|  | **r = 0.25** | 0.902 | 0.9077 | 0.0004 | *0.9028* | 0.0005 | **0.9332** | 0.0005 |
|  | **r = 0.5** | 0.910 | 0.9154 | 0.0048 | *0.9119* | 0.0004 | **0.9393** | 0.0002 |
| **Jumpy GAM with local interactions (MB6L)** | **r = 0** | 0.891 | 0.8876 | 0.0005 | *0.8647* | 0.0011 | **0.9300** | 0.0006 |
|  | **r = 0.25** | 0.884 | 0.8924 | 0.0004 | *0.8785* | 0.0007 | **0.9274** | 0.0004 |
|  | **r = 0.5** | 0.891 | 0.8952 | 0.0002 | *0.8818* | 0.0005 | **0.9265** | 0.0005 |
| **Complex (MB8L)** | **r = 0** | 0.859 | 0.8522 | 0.0004 | *0.8415* | 0.0015 | **0.8684** | 0.0009 |
|  | **r = 0.25** | 0.864 | 0.8547 | 0.0006 | *0.8399* | 0.0023 | **0.8772** | 0.0012 |

| | r = 0.5 | 0.855 | 0.8591 | 0.0006 | *0.8429* | 0.0017 | **0.8775** | 0.0009 |

Table 23: Training MSE of all models with a larger set of predictors for continuous response cases with sample size N = 50k (the smallest value is bolded; the largest value is italicized)

| Functional Form | Correlation | Oracle MSE | XGB | | FFNN | | RF | |
|---|---|---|---|---|---|---|---|---|
| | | | Mean | SD | Mean | SD | Mean | SD |
| Linear with global interactions (MC2L) | r = 0 | 0.984 | 1.0752 | 0.0144 | **0.7053** | 0.0074 | *3.2683* | 0.0334 |
| | r = 0.25 | 0.984 | 1.0731 | 0.0169 | **1.0441** | 0.0093 | *3.2288* | 0.0341 |
| | r = 0.5 | 0.984 | **1.0285** | 0.0070 | 1.0552 | 0.0136 | *3.0755* | 0.0320 |
| GAM with global interactions (MC5L) | r = 0 | 0.984 | **1.0294** | 0.0090 | 1.3422 | 0.0106 | *2.1323* | 0.0176 |
| | r = 0.25 | 0.984 | **1.0294** | 0.0082 | 1.3194 | 0.0830 | *2.3457* | 0.0173 |
| | r = 0.5 | 0.984 | **1.0342** | 0.0133 | 1.1339 | 0.0146 | *2.4588* | 0.0354 |
| Jumpy GAM with local interactions (MC6L) | r = 0 | 0.984 | **0.8573** | 0.0077 | 1.3431 | 0.0201 | *2.2657* | 0.0261 |
| | r = 0.25 | 0.984 | **0.8524** | 0.0067 | 1.3551 | 0.0339 | *2.0597* | 0.0138 |
| | r = 0.5 | 0.984 | **0.8536** | 0.0085 | 1.3749 | 0.0200 | *1.9716* | 0.0142 |
| Complex (MC8L) | r = 0 | 0.984 | **1.1128** | 0.0128 | 1.4767 | 0.0341 | *3.1597* | 0.0359 |
| | r = 0.25 | 0.984 | **1.4659** | 0.0233 | 1.4932 | 0.0497 | *3.1158* | 0.0333 |
| | r = 0.5 | 0.984 | **1.4090** | 0.0186 | 1.5243 | 0.0790 | *3.0231* | 0.0351 |

Table 24: Training MSE of all models with a larger set of predictors for continuous response cases with sample size N = 50k (the smallest value is bolded; the largest value is italicized)

| Functional Form | Correlation | Oracle MSE | XGB | | FFNN | | RF | |
|---|---|---|---|---|---|---|---|---|
| | | | Mean | SD | Mean | SD | Mean | SD |
| Linear with global interactions (MC2L) | r = 0 | 0.984 | 1.2113 | 0.0117 | **0.9413** | 0.0022 | *3.0142* | 0.0080 |
| | r = 0.25 | 0.984 | 1.2208 | 0.0144 | **0.9430** | 0.0019 | *2.9888* | 0.0072 |
| | r = 0.5 | 0.984 | 1.2076 | 0.0092 | **0.9467** | 0.0023 | *2.8436* | 0.0048 |
| GAM with global interactions (MC5L) | r = 0 | 0.984 | 1.1882 | 0.0053 | **1.1377** | 0.0095 | *1.9235* | 0.0100 |
| | r = 0.25 | 0.984 | 1.1794 | 0.0059 | **1.1607** | 0.0019 | *2.0815* | 0.0106 |
| | r = 0.5 | 0.984 | 1.1703 | 0.0033 | **1.1519** | 0.0066 | *2.1336* | 0.0140 |
| Jumpy GAM with local interactions (MC6L) | r = 0 | 0.984 | **1.0404** | 0.0046 | 1.2085 | 0.0152 | *1.9684* | 0.0030 |
| | r = 0.25 | 0.984 | **1.0422** | 0.0045 | 1.2114 | 0.0112 | *1.9759* | 0.0073 |

|  | r = 0.5 | 0.984 | **1.0412** | 0.0026 | 1.2170 | 0.0153 | *1.8689* | 0.0071 |
|---|---|---|---|---|---|---|---|---|
|  | r = 0 | 0.984 | 1.7294 | 0.0231 | **1.2875** | 0.0109 | *3.4157* | 0.0098 |
| **Complex (MC8L)** | r = 0.25 | 0.984 | 1.4309 | 0.0098 | **1.2925** | 0.0079 | *3.3096* | 0.0131 |
|  | r = 0.5 | 0.984 | **1.2949** | 0.0045 | 1.3128 | 0.0149 | *3.1566* | 0.0083 |

Pouriyeh, Seyedamin, Sara Vahid, Hamid Reza Arabnia, and Giovanna Sannino. 2017. "A comprehensive investigation and comparison of machine learning techniques in the domain of heart disease." *IEEE Symposium on Computers and Communication.* IEEE.

Tan, Aik Choon, and David Gilbert. 2003. "An empirical comparison of supervised machine learning techniques in bioinformatics." *First Asia Pacific Bioinformatics Conference.*

Wright, Marvin N., Andreas Ziegler, and Inke R. König. 2016. "Do little interactions get lost in dark random forests?" *BMC Bioinformatics* 17 (145).

Xiao, Jing, Ruifeng Ding, Xiulin Xu, Haochen Guan, Xinhui Feng, Tao Sun, Sibo Zhu, and Zhibin Ye. 2019. "Comparison and development of machine learning tools in the prediction of chronic kidney disease progression." *Journal of Translational Medicine* 17.